\documentclass{article}
\usepackage{enumitem}

\usepackage{titlesec}
\usepackage[dvipsnames]{xcolor} 
\usepackage{ucs}
\usepackage{longtable}
\usepackage[colorlinks=true,
  urlcolor=ForestGreen,
  linkcolor=Blue,
  citecolor=Violet,
  pagebackref=true,
  pdftitle={Titel},
  pdfsubject={},
  pdfauthor={},
  pdfkeywords={keywords}
]{hyperref}
\usepackage{graphicx}
\usepackage{subfigure}
\usepackage[subfigure]{tocloft}
\usepackage[font=footnotesize]{caption}
\usepackage{tabularx}
\usepackage{nameref}
\usepackage{tikz-cd}

\usepackage{amsmath} 	  
\usepackage{amssymb}	  
\usepackage{amsthm}	    
\usepackage{mathtools}  
\usepackage[normalem]{ulem} 
\usepackage{graphicx,wrapfig,lipsum}
\usepackage{float}
\usepackage{floatflt}
\usepackage[capitalize,nameinlink]{cleveref}

\newtheorem{lemma}{Lemma}

\usepackage[most]{tcolorbox}
\tcbuselibrary{breakable} 
\usepackage{wrapfig}
\usepackage[toc,page]{appendix}

\usepackage{pictexwd,dcpic}
\selectfont

\usepackage{geometry}
\def\wider{%
   \advance\leftskip -\parindent
   \advance\rightskip -\parindent}

\newcommand{\bt}[1][normal]{\begin{tikzcd}[ampersand replacement = \&, column sep=#1,row sep=#1]}
\newcommand{\et}{\end{tikzcd}}

  \tikzset{
    symbol/.style={%
        draw=none,
        every to/.append style={%
            edge node={node [sloped, allow upside down, auto=false]{$#1$}}}
    }
}

\makeatletter
\newcommand{\tpitchfork}{%
    \raise-0.1ex
    \vbox{
    \baselineskip\z@skip
    \lineskip-.52ex
    \lineskiplimit\maxdimen
    \m@th
    \ialign{##\crcr\hidewidth\smash{$-$}\hidewidth\crcr$\pitchfork$\crcr}
  }%
}
\makeatother

\newcommand{\Hom}{\mathrm{Hom}}
\newcommand{\Cat}[1]{\mathbf{#1}}

\usepackage{tocloft}

\usepackage{newtxtext}
\usepackage{lipsum} 

\makeatletter
\def\moverlay{\mathpalette\mov@rlay}
\def\mov@rlay#1#2{\leavevmode\vtop{%
   \baselineskip\z@skip \lineskiplimit-\maxdimen
   \ialign{\hfil$\m@th#1##$\hfil\cr#2\crcr}}}
\newcommand{\charfusion}[3][\mathord]{
    #1{\ifx#1\mathop\vphantom{#2}\fi
        \mathpalette\mov@rlay{#2\cr#3}
      }
    \ifx#1\mathop\expandafter\displaylimits\fi}
\makeatother

\usepackage{mathabx}
\usepackage{calc}
\usepackage{amscd}
\usepackage{pictexwd,dcpic}
\usepackage{makeidx}
\usepackage{tikz}
\usetikzlibrary{calc,intersections,through,backgrounds,angles,quotes}
\usepackage{pgfplots}
\pgfplotsset{compat=1.18}

\usepackage{array}
\usepackage{latexsym}
\usepackage{epsfig}
\newcommand{\R}{\mathbb{R}}

\newcommand{\bel}[1]{\begin{equation}\label{#1}}

\newcommand{\be}{\begin{equation}}
\newcommand{\ee}{\end{equation}}
\newcommand{\ba}{\begin{eqnarray}}
\newcommand{\ea}{\end{eqnarray}}
\newcommand{\rf}[1]{(\ref{#1})}

\newcommand{\qe}{\end{equation}}

\numberwithin{equation}{section}   
\numberwithin{figure}{section}

\newtheorem{theo}{Theorem}[section]
\newtheorem{satz}{Proposition}[section]

\theoremstyle{definition}
\newtheorem{defi}{Definition}[section]
\theoremstyle{remark}

\theoremstyle{remark}

\theoremstyle{remark}

\newcommand{\diam}{\mathrm{diam}}
\title{Merging Hazy Sets with m-Schemes: A Geometric Approach to Data Visualization}
\author{Lukas Silvester Barth\textsuperscript{\rm 2}, Hannaneh Fahimi\textsuperscript{\rm 1,\rm 2},
Parvaneh Joharinad\textsuperscript{\rm 1,\rm 2}, J\"urgen Jost\textsuperscript{\rm 2}
               Janis Keck\textsuperscript{\rm 2, \rm 3}\\
    \textsuperscript{\rm 1} Center for Scalable Data Analytics and Artificial Intelligence (ScaDS,AI) Dresden/Leipzig\\
    \textsuperscript{\rm 2} Max Planck Institute for Mathematics in the Sciences\\
    \textsuperscript{\rm 3} Max Planck Institute for Human Cognitive and Brain Sciences,\\
    lukas.barth@mis.mpg.de, fatemeh.fahimi@mis.mpg.de, parvaneh.joharinad@mis.mpg.de,\\ 
   jjost@mis.mpg.de, janis.keck@maxplanckschools.de. 
}
\begin{document}

\maketitle
\begin{abstract}
  Many machine learning algorithms try to visualize high dimensional metric data in 2D in such a way that the essential geometric and topological features of the data are highlighted.  In this paper, we introduce a framework for aggregating dissimilarity functions that arise from locally adjusting a metric through density-aware normalization, as employed in the IsUMap method. We formalize these approaches as m-schemes, a class of methods closely related to t-norms and t-conorms in probabilistic metrics, as well as to composition laws in information theory. These m-schemes provide a flexible and theoretically grounded approach to refining distance-based embeddings.
\end{abstract}
\section{Introduction}
\subsection{Data}
The data that are collected and that we are confronted with are often not only big, but also high dimensional. A data point may represent a large number of feature values of an underlying entity. When these feature values are scalar, we simply get a point in some high dimensional Euclidean space where each axis corresponds to some feature. However, intrinsically, the data may not be so high dimensional. There typically exist strong correlations between the feature values. This leads to a much lower intrinsic dimension. Therefore, it is natural to capture that intrinsic dimension as a first step. And for that purpose, it is natural to assume that the data lie on some lower dimensional manifold sitting in the original high dimensional Euclidean space. Extrinsically, it may stretch into many different directions, like a curve winding in all directions in an ambient space. And it may be highly curved, and therefore geometrically quite different from a flat Euclidean space. But to understand the data, we are interested in the intrinsic dimension and the geometric properties of that manifold. As a submanifold of Euclidean space, it naturally acquires a Riemannian structure. Thus, we seek to construct some Riemannian manifold representing the data. That we propose a manifold means that we exclude lower dimensional singularities which may be difficult to detect from the data. A manifold is good for local interpolation between data points, but still offers enough freedom for a global structure that can be very different from a Euclidean one. But even if, for instance, we have reduced an ambient 300-dimensional space to a 10-dimensional Riemannian manifold, this still is not good enough for visualizing the data. Our usual visualization abilities extend only to 2 or at most 3 dimensions. But simply projecting a higher dimensional Riemannian manifold to 2 dimensions will not do. It will just loose all the non-trivial medium and large range properties of the data. At the very least, we should preserve, and for purposes of visualization perhaps even enhance,  clusters in the data.\\
From the preceding, we may already conclude that, if the data lie on a non-linear Riemannian manifold with  certain global structures, and we want to preserve and perhaps even highlight these structures in a 2D representation, linear methods may not suffice. And the preceding should suggest that we should rather utilize concepts and tools from Riemannian geometry. And these concepts and tools should be combined with abstract mathematical considerations. First, the data themselves are discrete and not continuous, and so, the Riemannian manifold is a construct to approximate the data and interpolate between them. Second, we need to reconstruct global structures from local patches of the Riemannian manifold. In fact, a Riemannian manifold can be well approximated linearly at a local scale, but in general not globally. \\
So, how can geometric theory be converted into a machine learning scheme to construct efficient visualizations of high dimensional data that bring out their key local and global properties? In this contribution, we shall describe a modification and an improvement, IsUMap (an acronym to be explained below),  of a state-of-the art such machine learning method, UMAP.  UMAP is quite successful and widely employed in data analysis. It still has certain deficiencies, stemming partly from a somewhat imprecise theoretical foundation. We can, however, inspect the theoretical foundations and analyze where we can not only improve these foundations, but the algorithm itself. We shall demonstrate the improved capacity of IsUMap at various artificial and empirical data sets. We shall also embed this into the wider context of metric geometry.
For this purpose, we introduce aggregation operations, referred to as m-schemes, which allow us to merge sets equipped with dissimilarity functions, ultimately yielding an uber-metric space. This framework covers a broad range of merging options, including the canonical approach for gluing metric spaces in metric geometry (cf. \cite{Burago01}). Moreover, this aggregation produces a continuous triangulated space, which facilitates interpolation. In practice, computing the entire space is highly expensive and, in many cases, infeasible, requiring restriction to the original sample points. But from a theoretical point of view, this approach provides a rigorous and well-structured framework. A special application of this aggregation process, which, however, avoids most of the computational complexity, is the IsUMap scheme for data representation , c.f.  \cite{Barth24a,Barth25}, where the canonical merging operator is employed.

While IsUMap does not use triangles systematically, for a deeper understanding of the geometry, the metric relations in triangles are important. For  a corresponding generalization of the fundamental concept of Riemannian geometry, curvature, to the context of discrete metric spaces, we refer to \cite{Joharinad19}. In fact, while the data are discrete, Riemannian manifolds naturally occur as spaces into which metric data can be embedded and which then naturally yield a concise representation that allows for interpolations. \\
Since this is an expository article, we need to refer for some details to \cite{Barth24a,Barth25,Joharinad23}.

\subsection{Existing methods}
In order to construct a Riemannian manifold representing some metric data set, one first constructs a graph whose vertices are the data points, and sufficiently close such points are connected by edges. \emph{Sufficiently close} can either mean that we connect each point with its $k$ nearest neighbors, for some fixed $k$, creating a $k$-neighborhood graph. Or one selects a scale $\varepsilon >0$ and connects any two points of distance at most $\varepsilon$, to produce a $\varepsilon$-neighborhood graph. The first option encounters the problem that this neighborhood relation need not be symmetric. That is, if $p$ is among the $k$ nearest neighbors of $q$, $q$ need not be among the $k$ nearest neighbors of $p$. One can, of course, symmetrize the graph, that is, connect $p$ and $q$ if at least one of them is among the $k$ nearest neighbors of the other. -- The second option does not account well for non-uniformity of the data distribution. That is, in some region, many data points may be quite close to each other while other regions may be much more sparsely populated. \\
In any case, the graph also carries a metric $d$, by assigning to each edge a length given by the distance between its end points. We also note that the distances between the data points are those in the ambient (Euclidean) space, since the Riemannian metric is not yet known, but has to be constructed. It is a basic property of a smooth submanifold of Euclidean space, or more generally, of any Riemannian submanifold of some ambient Riemannian manifold, that when  points are sufficiently close to each other, the extrinsic distance in the ambient space is also close to the intrinsic distance on the submanifold. (Note that the extrinsic distance is always less or equal to the intrinsic one, with equality only when we are on a flat piece of the submanifold.)\\
Whichever option one chooses, one obtains a metric graph that is assumed to approximate the underlying Riemannian manifold from which the data are supposed to be sampled. In a foundational paper for manifold learning \cite{Belkin03}, Belkin and Niyogi show and exploit that the eigenvalues and eigenfunctions of the graph Laplacian converge to those of the underlying Riemannian manifold when the sampling gets finer and finer. The first eigenfunctions can then be used to construct an approximate embedding of that Riemannian manifold into some Euclidean space. For mathematical details, see for instance \cite{Joharinad23}. \\
A basic scheme for embedding a metric graph $(\Gamma,d)$  such as those just described is multi-dimensional scaling (MDS, \cite{Borg05}). In the metric version, one simply constructs a planar graph $\Gamma_0$ with the same vertex set as $\Gamma$ so that the average of the Euclidean distances between vertices of $\Gamma_0$ is as close as possible to their distance in $\Gamma$. And the Isomap scheme introduced in \cite{Tenenbaum00} precisely uses classical MDS for a metric graph constructed as above from a data set. Thus, the above step where a Riemannian manifold is constructed from the metric data is bypassed.\\
  Another early scheme, stochastic neighborhood embedding
method (SNE) \cite{Hinton02}, first uses 
 distances  to define conditional probabilities (in fact, with Gaussian weights from the distances) for 
a  random walk on the neighborhood graph and then a lower
dimensional representation whose corresponding
random walk is  similar to that on the original graph. An improved version, t-SNE \cite{vanderMaaten08}, uses the Student-t instead of a Gaussian distribution to achieve superior results.   \\
But such schemes overlook local properties of the data for the sake of constructing a global approximation. UMAP \cite{McInnes18} therefore first carefully chooses representations of the local neighborhoods of data points and then glues them together to reconstruct a global representation that can then be approximated by a 2D graph (see \cite{Damrich21}). While this often leads to considerable improvements of the visualization, in the sense that structural properties of the data set, like clusters, become visible in the 2D representations, it still does not account properly for non-uniformities in the data. While UMAP does uniformize the data, it can have  a problem with tearing the structures apart, especially the low-dimensional ones, at places where no separation should take place. This is due to their graph-embedding and negative sampling procedure.
IsUMap, while building upon UMAP, can handle such situations well. It overcomes the deficiencies introduced by negative sampling by carrying out Dijkstra's routine.\\
Among the categorical tools used by UMAP and also by IsUMap as presented in \cite{Barth24a,Barth25} are the fuzzy simplicial sets introduced by David Spivak \cite{Spivak09}. For that, one has to transform distances or weights into probabilities, and then back.\footnote{Such concepts of statistical metrics were first introduced by Karl Menger, see e.g. \cite{Menger42}, and \cite{Seising07} for the general history of concepts of fuzziness.}  For that, a monotonically decreasing surjective function from $[0,\infty]$ to $[0,1]$ has to be used. UMAP uses $f(x):=\exp{(-x)}$, but this seems a bit arbitrary. In \cite{Barth25}, therefore general transformation functions have been considered and their performance compared. The scheme presented here avoids the choice of a transfer function. Of course, the choice of an arbitrary function may offer the advantage of flexibility.  In the present contribution, we put the flexibility in a different  place which is  conceptually  more natural. We use the flexibility for rescaling local metrics, to bring out essential local features better. In Riemannian geometry, one can use conformal changes of a metric for that purpose. This does not work for arbitrary metrics, because multiplying a metric by a positive function may destroy the triangle inequality. We therefore work with a more general concept, that of a dissimiliarity. A dissimilarity need not satisfy the triangle inequality and therefore can be rescaled without leaving the corresponding category. Of course, at the end of the construction, we may want to have a metric that satisfies the triangle inequality. We shall therefore carefully explain in our scheme how to restore this inequality if that is our desire.

\section{The mathematics behind UMAP and IsUMap}
The scheme essentially consists of three steps:
  \begin{enumerate}
  \item Connect each data point with its $k$ nearest neighbors, to get a   multigraph (a graph that may have multiple edges between vertices), or a collection of star graphs
    
  \item Develop a canonical scheme for merging these multigraphs, with tools from category theory
    
    \item Project the resulting space onto 2D with minimal distortion.
    
    \end{enumerate}
To carry out and understand these steps, some mathematics is needed that we shall now describe. 

    \subsection{Some background in Riemannian geometry}\label{Riemann}
 Bernhard Riemann (1826-1866) in 1854 \cite{Riemann16}
  introduced the abstract concept of a  manifold of arbitrary dimension $n$ equipped with a metric tensor $(g_{ij})_{i,j=1,\dots ,n}$ with which one can compute a scalar product between tangent vectors at a point.
  
    \begin{equation}
      \label{1}
      \langle V,W\rangle =\sum_{i,j=1,\dots ,n} g_{ij}(x)v^i w^j
    \end{equation}
  for $V=\sum_i v^i \frac{\partial}{\partial x^i}, W=\sum_i w^j \frac{\partial}{\partial x^j}$ at a point $x$,
    
  Such a Riemannian manifold need not be realized in any ambient Euclidean space, but is intrinsically determined by its metric tensor. However, any smooth submanifold of a Euclidean space automatically inherits a Riemannian metric from the ambient Euclidean metric. \\
  
  Riemann solved the problem of determining a complete  set of local invariants for such a metric tensor. These come from  the \emph{curvature tensor}
  \begin{equation}
    \label{2}
    	R^k_{\ell ij} = \frac{\partial \Gamma^k_{j\ell}}{\partial x^i}
		- \frac{\partial \Gamma^k_{i\ell}}{\partial x^j}
		+ \Gamma^k_{im} \Gamma^m_{j\ell} - \Gamma^k_{jm} \Gamma^m_{i\ell}
              \end{equation}
              with $\Gamma^i_{jk}=\frac{1}{2}g^{i\ell}(\frac{\partial}{\partial x^k}g_{j\ell}+\frac{\partial}{\partial x^k}g_{k\ell}-\frac{\partial}{\partial x^\ell}g_{jk})$  (summation signs over double indices omitted; $g^{i\ell}$ is the inverse of the metric tensor). From the curvature tensor, Riemann derived the \emph{sectional curvatures} that are assigned to 2D-subspaces of the tangent space of the manifold at each point. These constitute the invariants. For details, we refer to \cite{Jost17a}.  \\

              Somewhat surprisingly, for the moment, we do not need curvatures, only the concept of a Riemannian metric.   In fact, Riemann introduced another important device, that of \textit{normal coordinates}.   Essentially, given a point $p$ in a Riemannian manifold $M$, for other points $q$ in the vicinity of $p$, normal coordinates record the distance $d(p,q)$ and the angles that the shortest geodesic from $p$ to $q$ makes with some reference direction at $p$. Like Euclidean polar coordinates.\\
              This does not work globally, because in general the shortest geodesic from $p$ to a more distant point in $M$ need not be unique.\\
              
              In practice, one needs to construct a model from a sample on the manifold, using information from an initial embedding that approximately preserves the configuration of sample points on the underlying manifold. To improve the representation, one then performs modifications to the model, but one needs a theoretical understanding of the effect of such modifications on the data manifold.\\
 For instance, suppose we use a $k$-neighborhood graph to represent the local connectivity of the manifold, weighting the edges according to radial distances. It is then natural  to scale these weights, to enhance local features, but we need to keep the geometric counterpart of this scaling on the manifold in mind. \\
 From the structure of normal coordinates discussed above, the radial distance corresponds to the first coordinate in this system. Consequently, one may investigate how imposing specific changes to this coordinate translates into a corresponding modification of the manifold itself.  \\
 
  Thus, for such a scheme, in a discrete metric space, for a point $x$, we take its $k$ nearest neighbors $x_1,\dots, x_k$, ranked according to the distance from $p$ and consider the star graph $S(x)$ with $x$ as its central vertex and edges of length $d(x,x_j)$ to the $x_j$ and then modify that star graph  to enhance local features.   \\
  This leads to a modification of the  metric, and it is expedient to turn to categories that are somewhat more general than metric spaces to gain additional flexibility.

\subsection{Dissimilarities}\label{dissim}
\begin{defi}
A \emph{dissimilarity set} $(X,\delta)$ is a set $X$ equipped with  a map $\delta:X\times X\to \mathbb{R}_{\ge 0}\cup \{\infty\}$ with $\delta(x,x)=0$. 
The category of dissimilarity sets $\mathbf{DM}$ has as
objects dissimilarity sets and as morphisms  non-expansive maps that is, maps that do not increase dissimilarities. The category of finite dissimilarity sets is denoted by $\mathbf{FinDM}$.\\
A dissimilarity $\delta$ is \emph{symmetric} if
\begin{equation}
  \label{dis2}
  \delta(x,y)=\delta(y,x) \quad \text{ for all }x,y. 
\end{equation}
We then have corresponding categories $\mathbf{symDM}$ and $\mathbf{FinsymDM}$.\\
The \emph{diameter} of a  dissimilarity set $(X,\delta)$ is
\begin{equation}
  \label{dis3}
  \diam(X):= \sup_{x,y \in X}\delta(x,y)\ .
\end{equation}
The dual of a similarity set $(X,\delta)$ is $(X,\delta^\ast)$
with
\begin{equation}
  \label{dis4}
  \delta^\ast(x,y):=\delta(y,x) \ .
\end{equation}
\end{defi}
We choose dissimilarities rather than similarities, because subsequently we want to connect the concept with that of a (somewhat generalized) metric, and metrics  quantify differences.\\
Points with $\delta(x,y)=\delta(y,x)=0$ cannot be distinguished by dissimilarity.  And points with $\delta(x,y)=\delta(y,x)=\infty$ are incompatible. In  pseudo-metric spaces,  $d(x,y)=0$ defines an equivalence relation. For dissimilarities, however,  being indistinguishable in the sense of $\delta$ is not an equivalence relation, because, since we do not require $\delta$ to be symmetric or to satisfy the triangle inequality,   $x$ could be indistinguishable from $y$ and $y$ from $z$ while $x$ is not indistinguishable from $z$.  \\
Not requiring the triangle inequality gives us the flexibility of rescaling. Whenever $(X,\delta)$ is a dissimilarity set and $f:[0,\infty]\to [0,\infty]$ is a monotonically increasing function with $f(0)=0$, then $(X,f\circ \delta)$ is again a dissimilarity set. \\
We can replace the triangle inequality with a more flexible relation. For instance, we can substitute the summation in the triangle inequality with a binary operator $T$ on $[0,\infty]$, establishing a comparison between the aggregated dissimilarities  $T(\delta(x,y),\delta(y,z))$ and $\delta(x,z) $. To ensure a coherent and meaningful structure, we must impose specific conditions on the aggregation operator $T$. \\
  For instance, the statistic metric introduced in \cite{Menger42}, which assigns to each pair $(x,y)$ a distribution on the positive part of the real line instead of a single scalar value, incorporates a \emph{triangular norm} (\emph{t-norm} for short)  as a replacement for the triangle inequality. \cite{Menger42} then specifies the properties of a t-norm to ensure a coherent and meaningful structure. For example, along with symmetry and monotonicity, such a t-norm $T$ must satisfy a condition to guarantee that metric spaces are special cases of statistical metric spaces.  \cite{Schweizer03} provides a more detailed discussion on the development of Menger's notion of statistical metrics.

\subsection{Uber-metric spaces}\label{uber}
For (symmetric) dissimilarity sets, we do not require the triangle inequality. We therefore refine the concept correspondingly. 
\begin{defi}
An \emph{uber-metric space} $(X,d)$ is a set $X$ equipped with  a map $d:X\times X\to \mathbb{R}_{\ge 0}\cup \{\infty\}$ such that
\begin{enumerate}
\item  $d(x,x)=0$,
  \item $d(x,y)=d(y,x)$, and 
\item $d(x,z)\le d(x,y)+d(y,z)$.
\end{enumerate}
The category of uber-metric spaces $\mathbf{UM}$ has as
objects uber-metric spaces and as morphisms  non-expansive maps that is, maps that do not increase distances. The category of finite uber-metric spaces\index{finite uber-metric space} is denoted by $\mathbf{FinUM}$.
\end{defi}
The second property in the definition of an uber-metric space implies that  if $d(x,z)=\infty$, then for any $y$ either $d(x,y)=\infty$ or $d(z,y)=\infty$.\\
From an uber-metric we obtain a topology in the same way as from an ordinary metric space. Therefore, we can talk about uber-metric \emph{spaces}.\\
\medskip

An uber-metric is the type of generalized metric that we want to  equip our star graphs with. 
  \begin{eqnarray}
  \label{eq:nn6UM}
  d_x(x,x_{j})&=&\frac{d(x,x_{j})-\rho_x}{\sigma_x}\quad
                           \text{ for }j=1,\dots k\\
  \nonumber
  d_x(y,y)&=& 0\quad \text{ for all }y\\
  \label{eq:nn7UM}
  d_x(x_{j},x_{\ell})&=&a(d_x(x_{j},x)+d_x(x,x_{\ell})) \quad \text{ for  neighbors }x_j,x_\ell \text{ of }x\\
  \label{eq:nn8UM}
  d_x(y,z)&=&\infty \quad \text{ in all other cases}\\
              \nonumber
   \rho_x&=& d(x,x_1) \\
              \nonumber
   \sigma_x&=& d(x,x_k)                     
  \end{eqnarray}
      The factor $a$ in  \eqref{eq:nn7UM} should be $\le 1$ to preserve the triangle inequality. We may simply take $a=1$, but for instance also $\frac{1}{\sqrt{2}}$ as suggested by Michael Freedman. That latter choice would correspond to the expected Euclidean distance of two points randomly chosen with the same distance from a center in a high-dimensional space. Of course, we could also directly use the (suitably rescaled) Euclidean distances between $x_j$ and $x_\ell$. In UMAP, one puts the distance $=\infty$ between different neighbors of the center $x$. This then violates the triangle inequality for such incompatible pairs, and consequently one has to work with an  extended pseudo-metric rather than an ubermetric. 
            
            The choice of $\rho_x$ eases the curse of dimensionality because randomly picked points in a high dimensional ball tend to concentrate near the outer boundary. For many data sets, however, already $\rho_x=0$ works well. \\
            
            $\sigma_x$ then just achieves some normalization.  This is important if the data points are non-uniformly distributed on the manifold.\\

   Now, what to do with those local star graphs? How to recombine them into a global structure?

   \section{Hazy sets}\label{hazy}
   \begin{defi}
     We equip $[0,\infty]$ with the topology whose nontrivial open sets are the intervals $(s,\infty]$ for $s\ge 0$. Often, we indicate such a set simply by $s$. The intervals $(s,\infty]$ are ordered by inclusion, and we write $i_{ts}:t\to s$ for the inclusion $(t,\infty] \subset (s,\infty]$ for $s\le t$. This gives us the category $\Cat{H} $.\\
      A \emph{hazy set} $S$ is a sheaf on $\Cat{H}$ for which all restriction maps $S(i_{ts}:t\to s):S(s)\to S(t)$ are injections.
      Their category is denoted by $\Cat{Haz}$.\\
      We call $s$ the \emph{haziness} of $S(s)$ and consider $S(s)$ as the set of haziness at most $s$. \\
        We can also put 
   \begin{equation}
     \label{h1}
     X:= S(\infty)
   \end{equation}
   which, as it happens, is the colimit of the $S(s)$ in the category of sets. Then all $S(s)$ become subsets of $X$. We can also look at
   \begin{equation}
     \label{h2}
     S(=t):=S(t)\backslash \bigcup_{s>t}S(s),
   \end{equation}
   the set of haziness precisely $t$.\\
   \medskip
   
      Analogously to Spivak's fuzzy simplicial sets \cite{Spivak09}, we can also define a \emph{hazy simplicial set} as a functor $\Cat{\Delta^{op}}\to \Cat{Haz}$ where $\Delta$ is the simplicial indexing category. This gives us the functor category $\Cat{sHaz}$ of hazy simplicial sets. 
   \end{defi}
 
   Using some standard category theory, we can also understand a hazy simplicial set
   as a functor $S:(\Delta\times \Cat{H})^{\text{op}}\to \Cat{Set}$. We write $S^n_s:=S([n],(s,\infty])$ where $n$ indicates $n$-simplices. These are the $n$-simplices in the image of $S$ of haziness at most $s$. \\
   One can then show, as in \cite[Prop.4.3.5]{Barth25}, 
   \begin{satz}
     The haziness of a simplex is at least equal to the maximum of the hazinesses of its
     faces, or  simpler: A simplex is at least as hazy as its faces.\\
     And all degeneracies of a simplex are as hazy as the simplex.
   \end{satz}

\subsection{Hazyness becomes metric}
From the Yoneda embedding $(\Delta \times \Cat{H}) \to \Cat{Set}^{\Delta\times \Cat{H}^{op}}$ we obtain  functors $ \Delta_s^n$ with 
\begin{equation}
  \label{h3}
  \Delta_s^n(m,t) = \Hom_{\Delta\times \Cat{H}}((m,t), (n,s)).
 \end{equation}
Such morphisms exist only for $s\le t$. And therefore, morphisms $\Delta^n_s\to \Delta^\ell_r$ (obtained by composing morphisms $((m,t)\to (n,s) \to (\ell,r)$) can exist only for $r\le s$. \\
We consider  $\Delta^n_s$ as  the $n$-simplices of haziness at most $s$. We then have the \emph{smear functor} $\Cat{Sm}$ that associates to $\Delta^n_s$ a geometric $n$-simplex of diameter $s$,
\begin{equation}
  \label{h4}
  \Cat{Sm}_\Delta(\Delta^n_{s}):= \left\{ x \in\mathbb{R}^{n+1}~\bigg|x^i \ge 0 \text{ and }~\sum_{i=1}^{n+1} x^i = s \right\}.
\end{equation}
On morphisms, it operates by rescaling by $\frac{s}{t}$ when $s\le t$.\\
By using Kan extensions as in \cite[Sect.4.2.4]{Barth25}, we can also consider $\Cat{Sm}$ as a functor from $\Cat{sHaz}$ to $\Cat{UM}$, the category of uber-metric spaces.  The image of a hazy simplicial set is indeed an uber-metric space. Let us explain this (a more detailed argument follows along the lines of \cite[Section 4.3.3]{Barth25}). Every simplex is equipped with a metric by \eqref{h4}. This is compatible with the glueing of simplices along shared faces, as described by the colimit of the Kan extension. According to the properties of a hazy simplicial set, the faces of a hazy simplex cannot be more hazy than the simplex itself. Therefore, under \eqref{h4}, their diameter cannot become larger. Hence, a shortest path within such a face cannot be made shorter by going into an ambient higher-dimensional simplex. (On the other hand, a path in the interior a simplex could possibly be shortened by going into a face. proceeding their along a shortest path and returning to the original simplex to reach the end point of the path.)\\
\medskip

We also want to go into the opposite direction, from an uber-metric space to a simplicial hazy set. 
\begin{defi}The \emph{tightening functor} is 
  \begin{equation}
      \begin{split}
        \Cat{Ti}:\Cat{UM}&\to\Cat{sHaz}\\
        Y &\mapsto \Cat{Ti}(Y): (\Delta\times \Cat{H})^{\text{op}}\to\Cat{Set}\\
        &\qquad\qquad\qquad (n,s) \mapsto
        \Hom_{\Cat{UM}}(\Cat{Sm}(\Delta^n_{s}),Y)\ .
      \end{split}
    \end{equation}
\end{defi}
We recall that a morphism in $\Cat{UM}$ has to be distance non-increasing. Therefore, the larger $s$, the more morphisms there are into a given uber-metric space $Y$, because $\Cat{Sm}(\Delta^n_{s})$ is a simplex with diameter $s$.
\begin{theo}\label{adj}
  $\Cat{Ti}$ is right adjoint to $\Cat{Sm}$, i.e.
    \begin{equation}
        \Hom_{\Cat{UM}}(\Cat{Sm}(S),Y)
        \simeq
          \Hom_{\Cat{sHaz}}(S,\Cat{Ti}(Y))\ .
    \end{equation}
  \end{theo}
   Here, we recall                
                \begin{defi}
                   Given two functors $F:\Cat{A}\to\Cat{B}$ and $G:\Cat{B}\to\Cat{A}$, 
          $F$ is \textit{left adjoint}\index{left adjoint}\index{adjoint} to $G$, and that $G$ is \textit{right adjoint}\index{right adjoint} to $F$, and we write
          $F \dashv G$, iff there is an isomorphism
          \begin{equation}
            \begin{split}
              \Hom_{\Cat{A}}(A,G(B))\simeq \text{Hom}_{\Cat{B}}(F(A),B)
            \end{split}
          \end{equation}
          which is natural in both $A \in \Cat{A}$ and $B \in \Cat{B}$. One also refers to
          $F$ and $G$ together as an \textit{adjunction}.
                \end{defi}

The \emph{proof} of Thm. \ref{adj} is analogous to that of \cite[Prop.4.3.10]{Barth25}. The essential point is that $\Cat{Sm}$ makes a hazy set $S(n,s)$ as large as needed to accommodate all distances up to $s$, while $\Cat{Ti}$ makes an uber-metric space as little hazy as possible under the constraint of  accommodating all maps from simplices of size $s$. \\
In some sense, this can be seen as a refinement of
the facts that the forgetful functor $F:\Cat{UM}\to \Cat{Sets}$ that
      maps a uber-metric space $(Y,d)$ to its underlying set $Y$ has a left
      adjoint
      \begin{eqnarray*}
        L:&\Cat{Sets}&\to \Cat{UM}\\
        &Y&\mapsto (Y,d_\infty)
      \end{eqnarray*}
      where  $d_\infty(x,y)=\infty$ for all
      $x\neq y$, and that  it has a right adjoint
      \begin{eqnarray*}
         R:&\Cat{Sets}&\to \Cat{UM}\\
        &Y&\mapsto (Y,d_0)
      \end{eqnarray*}
      where  $d_0(x,y)=0$ for all $x,y$. The left adjoint makes the metric as large as possible, while the right adjoint makes it as small as possible. Thus, for the left adjoint, any map between sets is turned into a non-increasing map from an uber-metric space, and for the right adjoint, it becomes a non-increasing map into such a space. Likewise, $\Cat{Sm}$ facilitates maps into an uber-metric space whereas $\Cat{Ti}$ facilitates morphisms into a hazy set.

      \subsection{m-schemes}\label{haz}
      As described, in our IsUMap scheme, in a first step, we produce star graphs representing the local geometries around data points.   In the next step, these star graphs then have to be merged into a global graph. For that purpose, we now develop a general framework. The star graphs carry uber-metrics, but since one and the same edge may carry several distances, depending on the perspective of the data point from where we look, it is natural to work with dissimilarities. Also another reason for working with dissimilarities is that  rescaling of local metrics may destroy triangle inequalities (although the triangle inequality can be restored in a canonical way, as we shall see below). And since merging is most natural in probabilistic settings, it is natural to turn to the setting of hazy sets, as hazyness can be turned into the probability of presence. The smaller the hazyness of a set, say an edge in one of our star graphs, the more likely it is to be present. \\

      In order to proceed in some generality, we start with
      \begin{defi}
        An \emph{m-scheme} is a function
$$M:\left[0,\infty \right] \times \left[ 0,\infty \right] \to \left[ 0,\infty \right]$$ that satisfies the following properties :
\begin{itemize}
\item Symmetry : $M(s,t)=M(t,s),$
\item Monotonicity : $M(s,t)\leq M(v,w)$ if $s\leq v$ and $t\leq w$,
\item Associativity : $M(r,M(s,t))=M(M(r,s),t),$
\item  Boundary condition $M(s,\infty)=s$.
\end{itemize}
      \end{defi}
Such a notion was defined in \cite{Kampe67}, in their axiomatic development of information theory, where it was called  a \emph{composition law}. We call it an \emph{m-scheme} as we want to utilize it for \emph{merging} (generalized) \emph{metric} spaces.\\
      One might also want to impose the condition
      \begin{equation}
        \label{me1}
        M(s,s)=s 
      \end{equation}
      so that when we merge the same dissimilarity or haziness twice, we get that same value. But this is quite restrictive, and among the m-schemes that we might naturally work with, it seems that only  the canonical one satisfies this condition. Therefore, for the sake of generality,  we do not impose \eqref{me1} in the sequel. \\
     
      Any m-scheme satisfies
      \begin{equation}
        \label{me2}
        M(s,0)=M(0,s)=0\quad \text{ for all }s\ .
      \end{equation}
      This follows from $M(0,\infty)=0$, monotonicity and symmetry.\\
      We have two extremal  examples of m-schemes:
      \begin{itemize}
\item  $M_{\min}(s,t)=\min (s,t)$
\item 
$       
M_{\mathrm{ext}}(s,t)=
\begin{cases}
t\quad   \text{ if } s=\infty\\
s\quad   \text{ if }  t=\infty ,\\
0\quad   \text{ else}.
\end{cases}
$
\end{itemize}
In order to embed them in a family and to interpolate between them, let $V$ be any subset of $\R^{\ge 0}$ with $0\in V$  and consider
\begin{equation}
  \label{me4}
 M_{V}(s,t)=
\begin{cases}
t\quad   \text{ if } s=\infty\\
s\quad   \text{ if }  t=\infty ,\\
\sup_{v\in V}\{~v~|~v\le s \text{ and }v\le t~\} \quad   \text{ else}.
\end{cases} 
\end{equation}
For $V=\R^{\ge 0}$, we get $M_{\min}$, while for $V=\{0\}$, $M_{\mathrm{ext}}$ results. Also, if $U\subset V$, then $M_U\le M_V$. Thus, we obtain a natural family of m-schemes if we take $V=[0,a]$. Among all the $M_V$, only for $V=\R^{\ge 0}$, that is, for $M_{\min}$, we get \eqref{me1}.  \\
m-schemes are analogous to t-(co)norms that are schemes for merging probabilities. We recall that  a \emph{t-conorm}\index{t-conorm} is a binary operation
$$T^{\text{co}}:\left[0,1 \right] \times \left[ 0,1 \right] \rightarrow \left[ 0,1 \right]$$
which satisfies, for all $a,b\in  \left[ 0,1 \right]$, symmetry, associativity, monotonicity and $T^{\text{co}}(a,0)=a$.
For a proper function $f:[0,\infty]\to [0,1]$, nondecreasing with $f(\infty)=0$ and $f(0)=1$ the combination of $f$ with any t-conorm  defines an m-scheme. Moreover, with any such a function $f$ the combination of $f^{-1}$ and $M_{\min}$ is the 
the maximum t-conorm $T^{\text{co}}_{M}(a,b)=\max (a,b)$, and similarly, $M_{\mathrm{ext}}$ is the analogue of the drastic sum 
$       
T^{\text{co}}_{D}(a,b)=
\begin{cases}
b\quad   \text{ if } a=0,\\
a\quad   \text{ if }  b=0,\\
1\quad   \text{ else}.
\end{cases}
$. \\
\medskip
There is a  connection to  Acz\'el's basic representation theorem and its application in synthesizing judgments. In fact, if the t-norm $T$ (where a t-norm is obtained from a t-conorm via $T(a,b)= 1-T^{\text{co}}(1-a,1-b)$) is continuous and $0<T(x,x)<1 \text{ for } 0<x<1$, then $T$ admits the representation 
\[
T(x,y)=f^{(-1)}(f(x)+f(y)),
\]
where $f:[0,1]\to [0,\infty]$ is a continuous and strictly decreasing function with $f(1)=0$ and 
\begin{equation}
f^{(-1)}(x)=
\begin{cases}
f^{-1}(x)\quad &\text{ if }  0\leq x \leq f(0)\\
0 \quad &\text{ if } x\geq f(0).
\end{cases}
\end{equation} 
The function $f$ is an additive generator of the t-norm $T$, c.f. \cite{Alsina82}. \\
\medskip
And one can switch between m-schemes (composition laws) and t-norms by  the following lemma of  \cite{Schweizer11}.
\begin{lemma}
  Let $T$ be a t-norm. Then
  \[
M_T(r,s)=\psi\left(T(\psi^{-1}(r),\psi^{-1}(s))\right)
\]
is an m-scheme (composition law) if and only if $\psi:[0,1]\to \R^+$ is continuous and strictly increasing, and conversely.
\end{lemma}
For example, for the t-norm $W(x,y):= \max(x+y-1,0)$, and $\psi(x):=\frac{1}{c}\log(1-x)$ for some  $c>0$ $M_W$ is the Wiener-Shannon law with parameter $c$, expressed by
\begin{equation}\label{WSlaw}
M_W(r,s)=\max\left(-\frac{1}{c}\log(e^{-cr}+e^{-cs}), 0\right).
\end{equation}
The t-norm $\Pi(x,y):=xy$, composed with the same function $\psi$, defines the composition law
\begin{equation}\label{productlaw}
M_{\Pi}(r,s)=-\frac{1}{c}\log\left(e^{-cr}+e^{-cs}-e^{-c(r+s)}\right),
\end{equation}
and combined with the function $\psi(x)=\frac{-1}{\log x}$, it yields the \emph{hyperbolic law}
\begin{equation}\label{hyplaw}
H(r,s)=\dfrac{rs}{r+s}.
\end{equation}
The canonical m-scheme we use in the context of IsUMap is called the  \emph{Inf-law} in this context.\\

\subsection{Merging dissimilarities and hazy sets}\label{merge}
We want to set up a correspondence between uber-metrics and 
 hazy simplicial sets. In order to avoid technical complications, we assume that all sets be finite, like the sample sets occurring in data analysis.\\
The scheme is the following. A hazy simplicial set derived from an uber-metric space is a  diagram (inverse system) of simplicial sets. From this, we construct a classical hazy simplicial set, which consists of a simplicial set (where $n-$dimensional simplices are identified) along with a haziness function. This hazy simplicial set is then realized as a metric space through a smearing process.
The specific properties of a hazy simplicial set that enable metric realization are
\begin{itemize}
\item[1-] The haziness of each simplex is at least equal to the maximum of the haziness of its faces. That is, the haziness function is non-increasing as we move toward lower-dimensional faces.
\item[2-] Each vertex is assigned a haziness value of zero. This property is particularly relevant in the discrete setting or when fixing the vertex set (as a sample from the uber-metric space). 
\item[3-] As the scale exceeds the diameter (in the case of finite diameter), the entire space (the sampled subset in the applications) is represented as a full simplex equipped with a haziness function.
\end{itemize} 
In the forward direction (uber-metric $\rightarrow$  hazy simplicial set), the triangle inequality is implicitly reflected in the haziness of edges. However, in general, a hazy simplicial set does not require edges to have haziness values that satisfy the triangle inequality. Despite this, in the backward direction (hazy simplicial set $\longrightarrow$ uber-metric) the reconstruction process ensures that the resulting binary function satisfies all the properties of an uber-metric.\\ 
By incorporating an m-scheme, one can merge two hazy simplicial sets. The symmetry property of the m-scheme ensures that this merging process is order-independent, while its monotonicity property guarantees that haziness remains non-decreasing when moving from a simplex to its faces after merging. Additionally, after merging, all vertices will retain a haziness value of zero.
One potential constraint arises from the condition  $M(s,\infty)=s$, as it requires careful construction to maintain monotonicity while ensuring consistency with the merging process.\\

We shall now use m-schemes to merge hazy sets, dissimilarities, or uber-metrics.
The general scheme for a set $X$ equipped with two dissimilarities $\delta$ and $\delta'$ and an m-scheme $M$, of course, 
yields a dissimilarity function $\delta_M:=M(\delta_1,\delta_2)$, which is defined by
\begin{equation}
  \label{me5}
\delta_M(x,y):=M(\delta_1(x,y),\delta_2(x,y)) \ .               
\end{equation}
We now apply this to the different structures that we have introduced.  Crucial properties are symmetry and the triangle inequality, and we want to see how they fare under merging. Of course, if we merge two symmetric structures, the result will be symmetric again. 
We can also use merging  to symmetrize a dissimilarity $\delta$; we put
\begin{equation}
  \label{me6}
  \delta_M:=M(\delta,\delta^\ast)
\end{equation}
where $\delta^\ast$ is the dual of $\delta$, see \eqref{dis4}. Thus, through a merging operation, we can symmetrize dissimilarities.\\
\medskip

When the two  structures are not defined on the same underlying set $X$, merging is still possible. Let they be defined on two sets $X_1,X_2$ with a non-empty overlap, $X_1\cap X_2\neq \emptyset$. We then  put  $X=X_1\cup X_2$. This comes up when we want to  merge two arbitrary hazy simplicial sets (which do not in general have a full simplex as the underlying simplicial set). Even if we ensure that both merging components have the same underlying vertex set, they need not have the same  sets of $n-$simplices. In the present framework, we can simply handle this by letting  the haziness of non-present simplices  be $\infty$. \\
\medskip

When  merging two dissimilarities, one cannot guarantee that the resulting function satisfies the triangle inequality, even if the individual components do. 
Here is a simple example.
$\delta_1(x,y)=1, \delta_1(x,z)=\delta_1(y,z)=3$ and $\delta_2(x,z)=1, \delta_2(x,y)=\delta_2(y,z)=3$. Then, using $M=M_{\min}$, 
$\delta_M(x,y)=1, \delta_M(x,z)=1, \delta_M(y,z)=3$, violating the triangle inequality. \\
There are two ways to get the triangle inequality back. One is to turn the uber-metric spaces to be merged  into  hazy simplicial sets by the functor $\Cat{Sm}$, merge those with an m-scheme and then turn the result into an uber-metric space by $\Cat{Ti}$.  
\\
The result of using  $\Cat{Sm}(M(\Cat{Ti}(d)))$ above with  the canonical min-merge operation can be described  more directly. From merging two uber-metric spaces, we obtain a dissimilarity that is symmetric, but need not satisfy the triangle inequality. But the above operation turns such a symmetric dissimilarity $\delta$ into an uber-metric $d$ via
\begin{equation}
  \label{me7}
  d(x,y):=\inf_{x_o=x, x_1, \dots ,x_\ell=y}\sum_{i=1}^\ell \delta(x_{i-1},x_i)\ .
\end{equation}
This satisfies the triangle inequality $d(x,y)\le d(x,z)+d(z,y)$ because the infimum over  arbitrary  sequences of intermediate points between $x$ and $y$ cannot be larger than that over sequences that are required to include the point $z$. 
\begin{enumerate}
\item[•] \textbf{Comparison of the resulting metrics:} A hazy simplicial set in the finite setting is the same as  a hazy simplicial complex, which is a simplicial complex equipped with a haziness function. According to the properties of an m-scheme and the smear functor $\Cat{Sm}$, the distance between a  pair of vertices is the length of the shortest path consisting of edges (1-dim simplices). Since the haziness of edges obtained via the $\Cat{Ti}$ functor is the dissimilarity between their end points, the two resulting uber-metrics are identical on the underlying space $X$.
\item[•] \textbf{Comparison of the resulting underlying space:} It is important to note that the underlying space of the metric resulting from the merging process remains the same as the original space when merging directly in the category $\mathbf{symDM}$. However, if we proceed through $\Cat{sHaz}$ the underlying space expands. In this latter case, we also obtain a continuous (metric) interpolation between subsets of the original set.  
\end{enumerate} 

\subsection{Merging star graphs to create a global representation of a metric data set}

We now apply this to the setting described above where we have turned a metric data set into a collection of star graphs. 
For merging the star graphs, constructions from category theory  were introduced in \cite{McInnes18}, but we found that we had to correct some claims there, provide some details and modify the construction somewhat \cite{Barth24,Barth25}. In the present contribution, we modify the construction by using hazy instead of fuzzy sets, which seems more natural and straightforward. \\
\medskip

  Here, we  use the concepts and constructions  just introduced.  We        convert the star graphs with their uber-metrics into hazy graphs that can be merged in a canonical way with the help of an m-scheme, which includes the more concrete construction just described in \eqref{me7}.\\
  
As described  at the end of the previous section, if the aim is to only merge the distances within the sample set $X$, the same result is obtained  through the combination $\Cat{Sm}\circ\Cat{Ti}$ of functors and directly in $\mathbf{symDM}$. In fact, the star graphs here represent a segment of the hazy simplicial sets where the haziness parameter is restricted to the interval $[0,b]$ for some positive number $b$ (this may change depending on how one estimates the distances between outer vertices of the star graph). This object can be completed by only adding the missing edges and assigning the haziness value of $\infty$ to  them. 
 \medskip

                At this point, you may ask: Why don't we simply construct a graph from the sample by connecting sufficiently close points and use the original distances? Or connect two points when at least one of them is among the $k$ nearest neighbors of the other? (These schemes were described in the introduction.)\\
  One of the earlier schemes, {\bf Isomap}, essentially does that, but the results become better when we use the apparently more complicated procedure suggested by {\bf UMAP} and refined by us in {\bf IsUMap}. This captures the local structure and the possible variation  of density better. \\
Moreover, this method allows for different merging processes by employing an m-scheme to aggregate dissimilarities. In fact, even after incorporating local variations in the dataset, one can still use a more general $k-$nearest neighbor graph by introducing multiple edges between pairs and then applying the shortest path distance (as in Isomap) to derive the resulting metric. However, in our framework, this would correspond only to the m-scheme $M_{min}$.   \\
As a preparation for this step, we had converted the star graphs into hazy simplicial sets. Thus, the edge weights have become hazy. \\
 \medskip           

                  All star graphs are defined on the same underlying set of sample points, and so we can identify points in different star graphs as $[x]$. When we use the minimum m-scheme and go back from hazy simplicial sets to weighted graphs, we get
    \begin{equation}
        \begin{split}
            d_{\sim} ([x], [x']) = \inf(d_X (p_1 , q_1 ) + \cdots + d_X (p_n , q_n )),
        \end{split}
        \label{eq:simMetric2}
    \end{equation}
    where the infimum is taken over all pairs of those sequences $(p_1 , \cdots , p_n ),~ (q_1 , \cdots , q_n )$ of
    elements of $X$ that satisfy
    \begin{equation}
        \begin{split}
            p_1 \sim x,\quad q_n \sim x',\quad \text{ and }\quad p_{i+1} \sim q_i ~\text{ for all } ~1 \le i \le n - 1,
        \end{split}
      \end{equation}
as proven in \cite[Prop. 6.3.1.]{Barth25}. And as there, we can also use more general m-schemes in \eqref{eq:simMetric2}.\\      
      Other m-schemes yield different results, and we can experiment which one works best for a given data sample.

      \section{IsUMap}
     
      We can now describe the general merging process. The IsUMap scheme \cite{Barth24a,Barth25} is covered by  this general approach. Although we apply the merging and representation algorithm to the local modifications \rf{eq:nn6UM}-\rf{eq:nn8UM} of the initial metric, this process can be applied to any set of partly overlapping sets equipped with dissimilarity functions. \\ 
      
      We can now describe the IsUMap scheme \cite{Barth24a,Barth25}. The name can be seen as a contraction of Isomap and UMAP, as it uses components of both these methods, but the highlighed \emph{UM} should also stand for \emph{uber-metric.} As indicated, we here modify the scheme by using hazy instead of fuzzy sets, as this is arguably more natural. 
      \medskip
      
                Given a sample $X:=\{x_1,\dots ,x_N\}$ from a metric data set
                \begin{enumerate}
                \item  Define metrics $d_i$ on $X$ that assign a distance of $\infty$ to distinct pair $(x,y)$ unless both are among $x_i$ and its $k$ nearest neighbors. For such neighboring points, the distance is given by the normalized values in \rf{eq:nn6UM} or \rf{eq:nn7UM}. 
                \item Merge these metrics (conceptually by converting them into hazy graphs and using an m-scheme for combining hazy distances).\\
    The hazy graph corresponding to the metric $d_i$ is a star graph centered at $x_i$, where the outer vertices (the neighbors of $x_i$) are also connected by edges. The haziness value on each edge is equal to the distance between its endpoints, while absent edges are assigned a haziness value of infinity.
One can extend this construction to higher-dimensional simplices when all their one-dimensional faces are present. The haziness of such a simplex is determined by the maximum haziness of its one-dimensional faces. In the context of IsUMap, triangles with finite haziness are also included.\\             
                \item The union of these hazy graphs forms a multigraph $\Gamma$ on the vertex set $X$, since there may exist more than a single edge between a pair, as it may appear in more than one  hazy graph. In general, one can simply assume that there are $N$ edges connecting each pair with haziness values, some of which may be $\infty$.  In the context of IsUMap, the Dijkstra algorithm \cite{Dijkstra59} is then applied to compute the distance function $d_{\Gamma}$ on the resulting global graph $\Gamma$.\\
               However, in the general case, the chosen m-scheme is first applied to aggregate the haziness values of each edge into a single value, transforming the multigraph into an ordinary hazy graph. Dijkstra's algorithm is then used to compute the graph distance. 
                  
                \item Approximate it by a 2D graph $(\Gamma_0, d_{\Gamma_0})$ with the same vertex set, using metric multidimensional scaling (initialized with classical multidimensional scaling), that is, minimizing a function like
                  \begin{equation}
                    \sum_{i,j=1,\dots ,N} (d_{\Gamma_0}(y_i,y_j)-d_\Gamma(x_i,x_j))^2
                  \end{equation}
where $y_i\in \Gamma_0$ corresponds to $x_i\in \Gamma$. 
                \end{enumerate}

\subsection{Illustrations}
In this section, we present illustrations that demonstrate the application of different m-schemes on some toy datasets, such as Swiss Roll with a hole, Torus and Two moons and M\"obius strip. 
In these experiments, we first apply the local modifications similar to IsUMap, as defined in Eq.s \rf{eq:nn6UM}, \rf{eq:nn7UM}, \rf{eq:nn8UM}, to the input distance function. However, since these datasets are initially embedded in a low-dimensional space, we omit the subtraction of the distance to the first nearest neighbor in the locally modified metrics. Next, we utilize each of the four distinct m-schmes , i.e. $M_V$, $M_{\Pi}$,  Wienner-Shannon composition law $M_{W}$ and hyperbolic law $H$, to merge the weighted graphs derived form the local distances using the $\Cat{Sm}$ functor. \\
The resulting weighted graph is then mapped to $2$-dimensional Euclidean space via metric-MDS, after computing the intrinsic distance (graph distance) using  Dijkstra's algorithm. The corresponding results are shown in  \cref{tab:SwRoll}, \cref{tab:Torus}, \cref{tab:2moons}, and \cref{tab:Mobius}. 
To fully integrate this approach, the algorithm of IsUMap was modified to directly incorporate merging within the hazy setting, introducing a parameter that provides this as an optional choice.\\
\medskip 
The parameter $c$ in $M_W$ and $M_\Pi$ adjusts the impact of merging haziness values in a similar manner by scaling them up or down. However, if this scaling causes the logarithm in \cref{WSlaw} to reach or exceed $\geq1$ (e.g. for very small $c$), the scheme returns a zero value. This effect is evident in our experiments, where a meaningful visualization is only achieved for $M_W$ when the parameter is set to $1$. To investigate this more, we apply the algorithm with larger values of parameter $c$ for $M_W$ and $M_\Pi$ to Swiss Roll with a hole and Two moons. The results are shown in \cref{tab:higher c}. \\
\medskip
The m-scheme $H$ does not have a tuning parameter. However, this can be considered analogous to a neutral setting in other schemes, where the parameter does not affect or adjust the haziness values. For consistency, we present the results for $H$ in the column corresponding to parameter $1$ in other schemes. \\
\begin{table}[H]
	\centering
	\tiny
	\begin{tabular}{>{\centering\arraybackslash}m{0.5cm}|*{5}{>{\centering\arraybackslash}m{2.5cm}}}
		& \textbf{$0.01$} & \textbf{$0.25$} & \textbf{$0.5$}  & \textbf{$0.75$} & \textbf{$1.0$} 
		\\
		& (a) & (b) & (c) & (d) & (e) \\
		\hline \\
		\textbf{(1) $M_V$} & \includegraphics[width=0.15\textwidth]{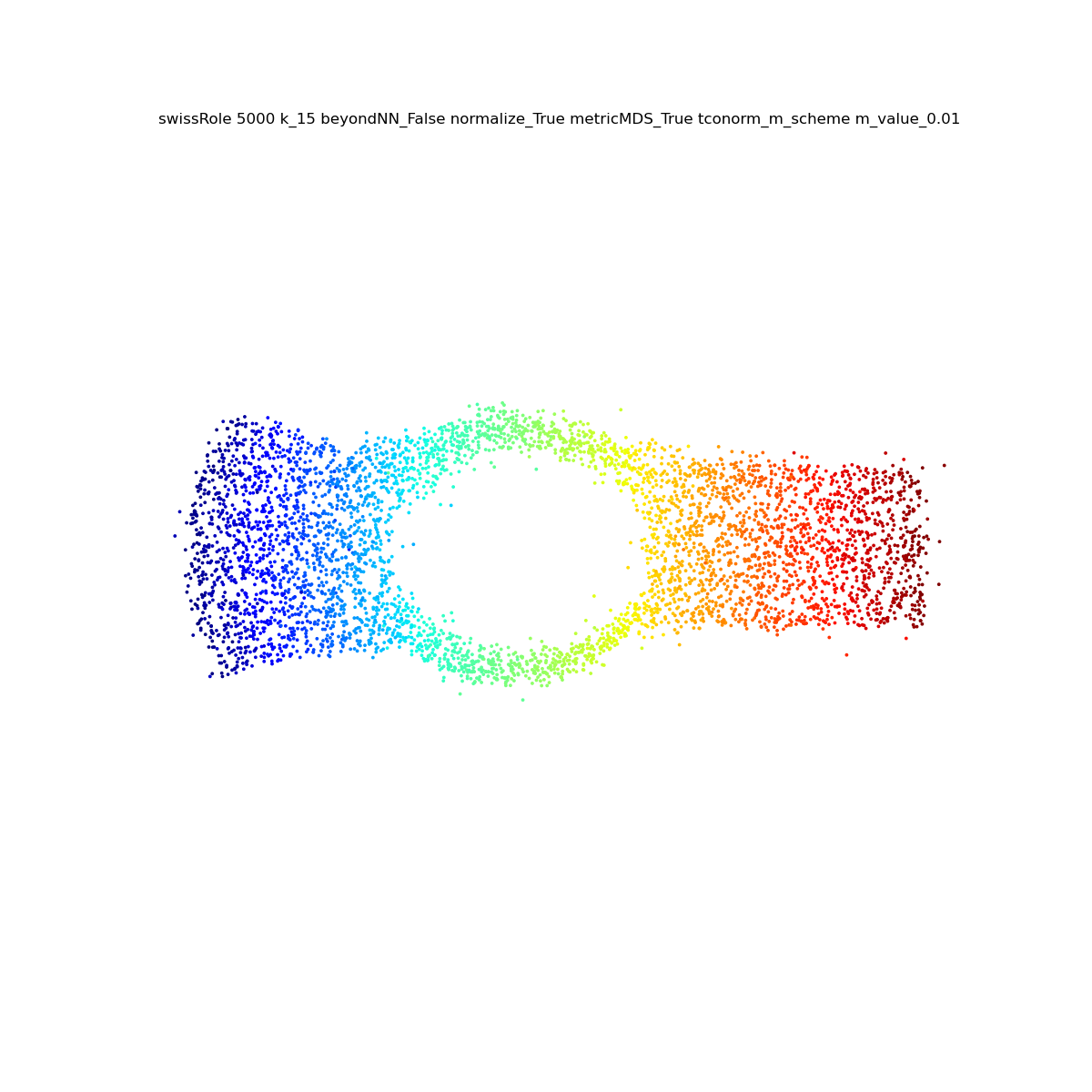} &  \includegraphics[width=0.15\textwidth]{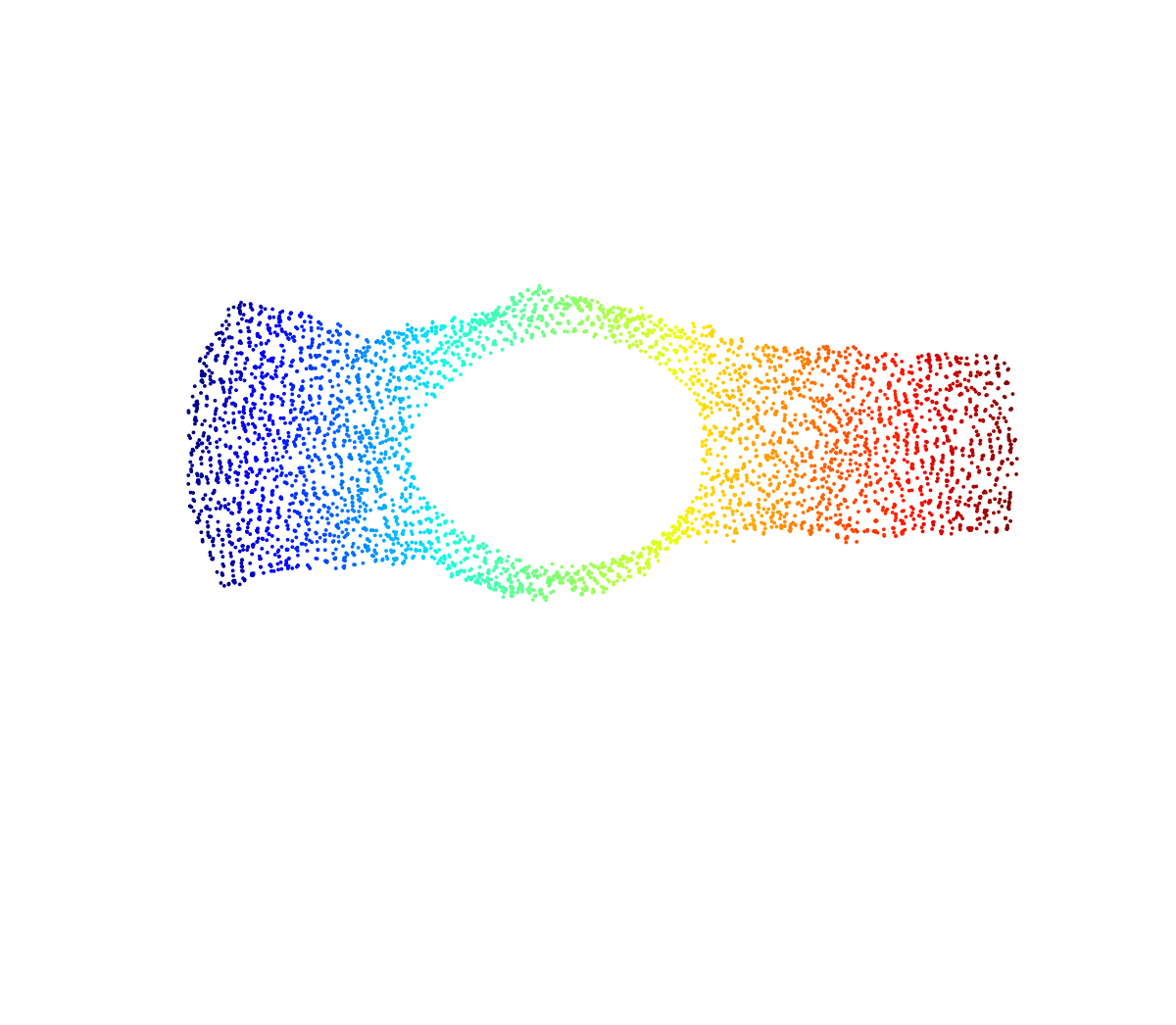} & \includegraphics[width=0.15\textwidth]{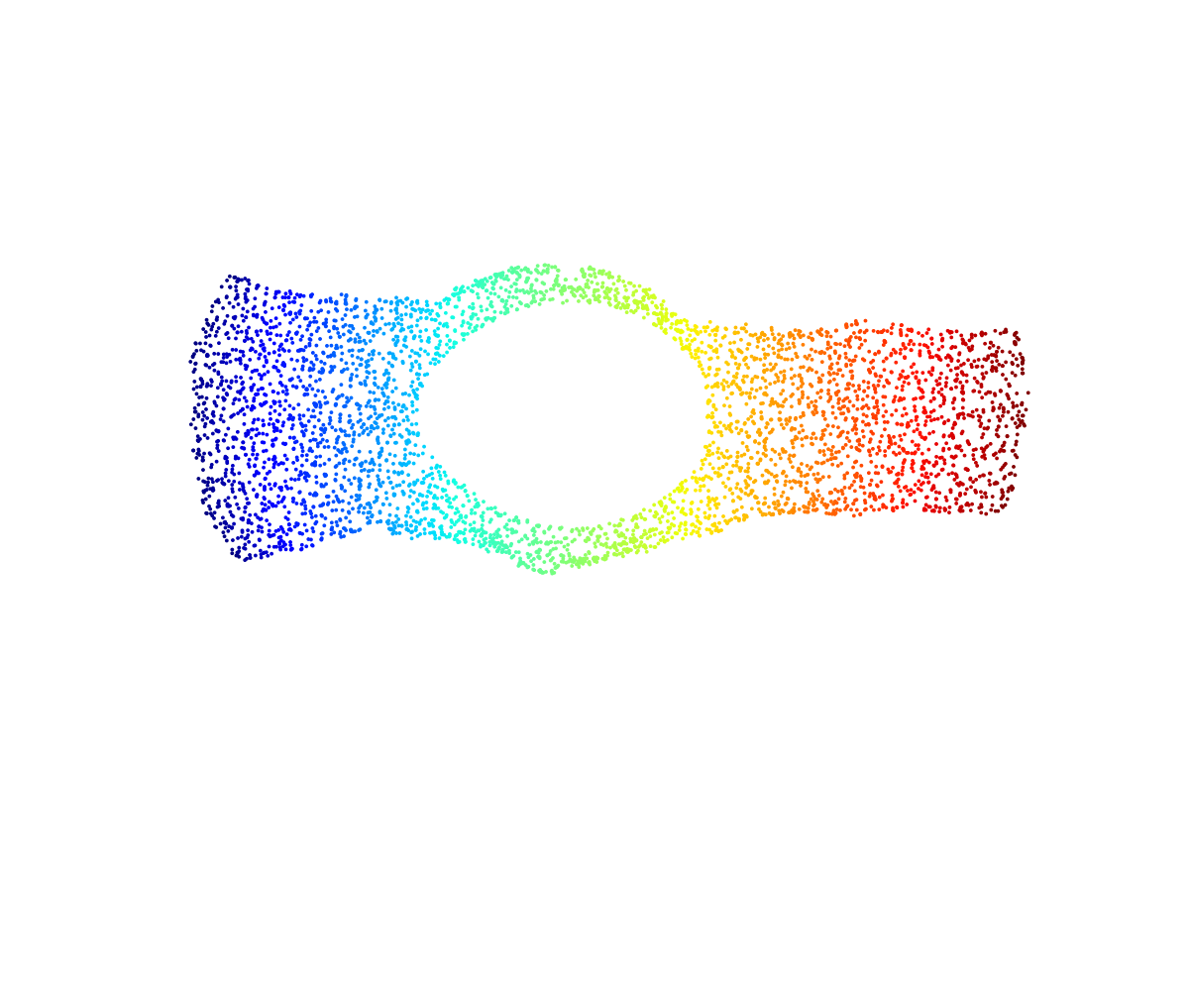}&
		\includegraphics[width=0.15\textwidth]{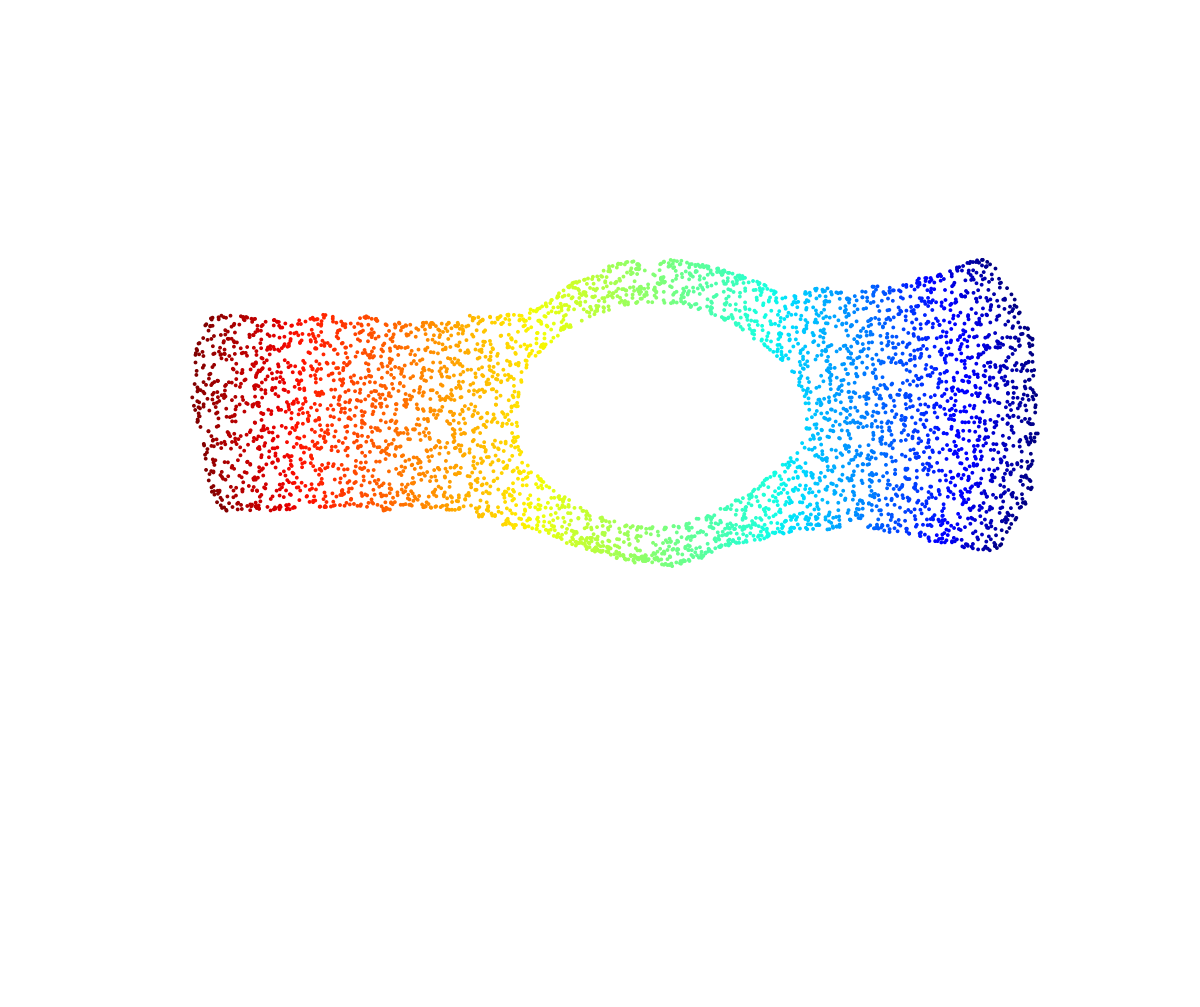} 
		&\includegraphics[width=0.15\textwidth]{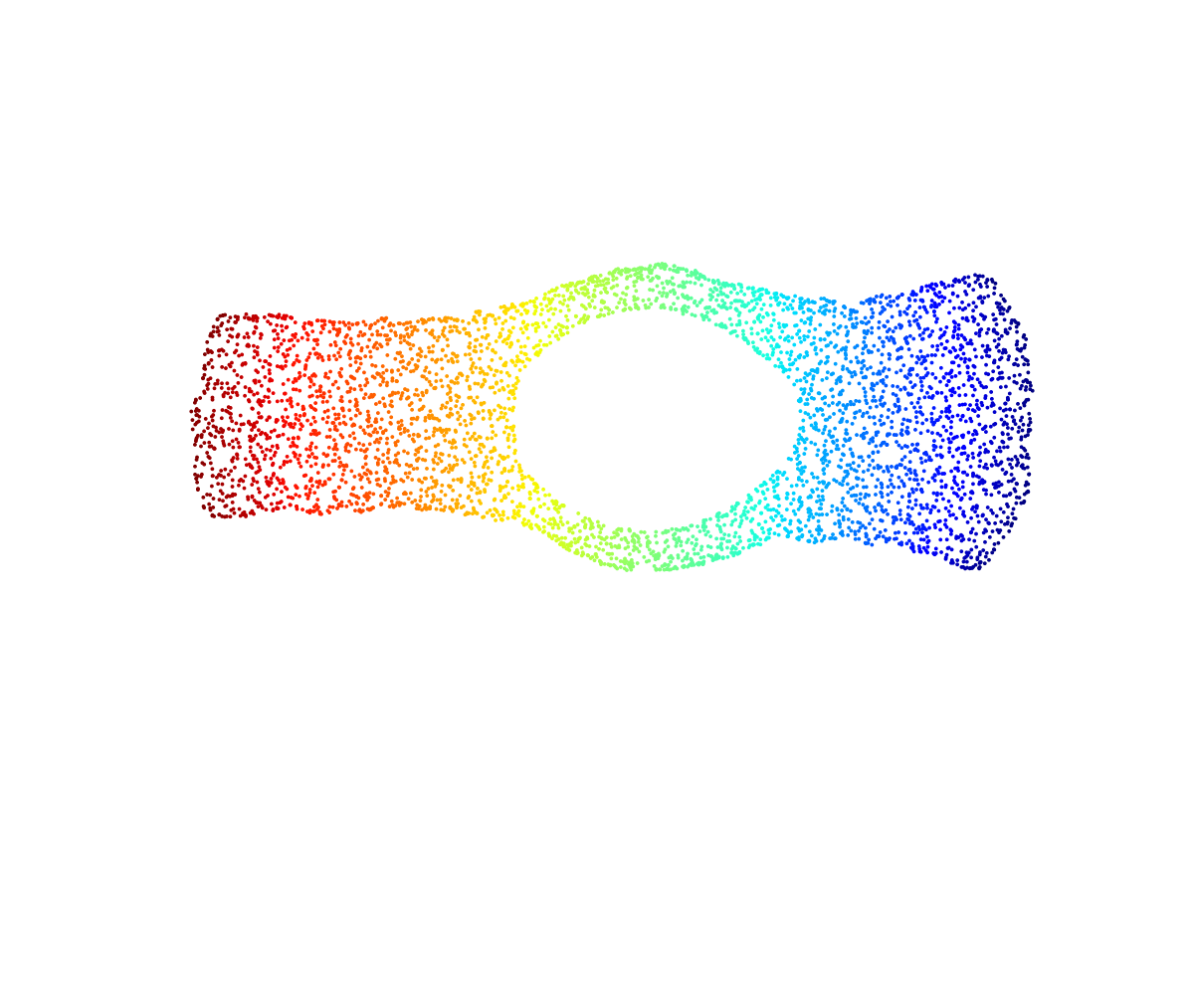} 
		\\
		\textbf{(2) $M_{\Pi}$} & \includegraphics[width=0.15\textwidth]{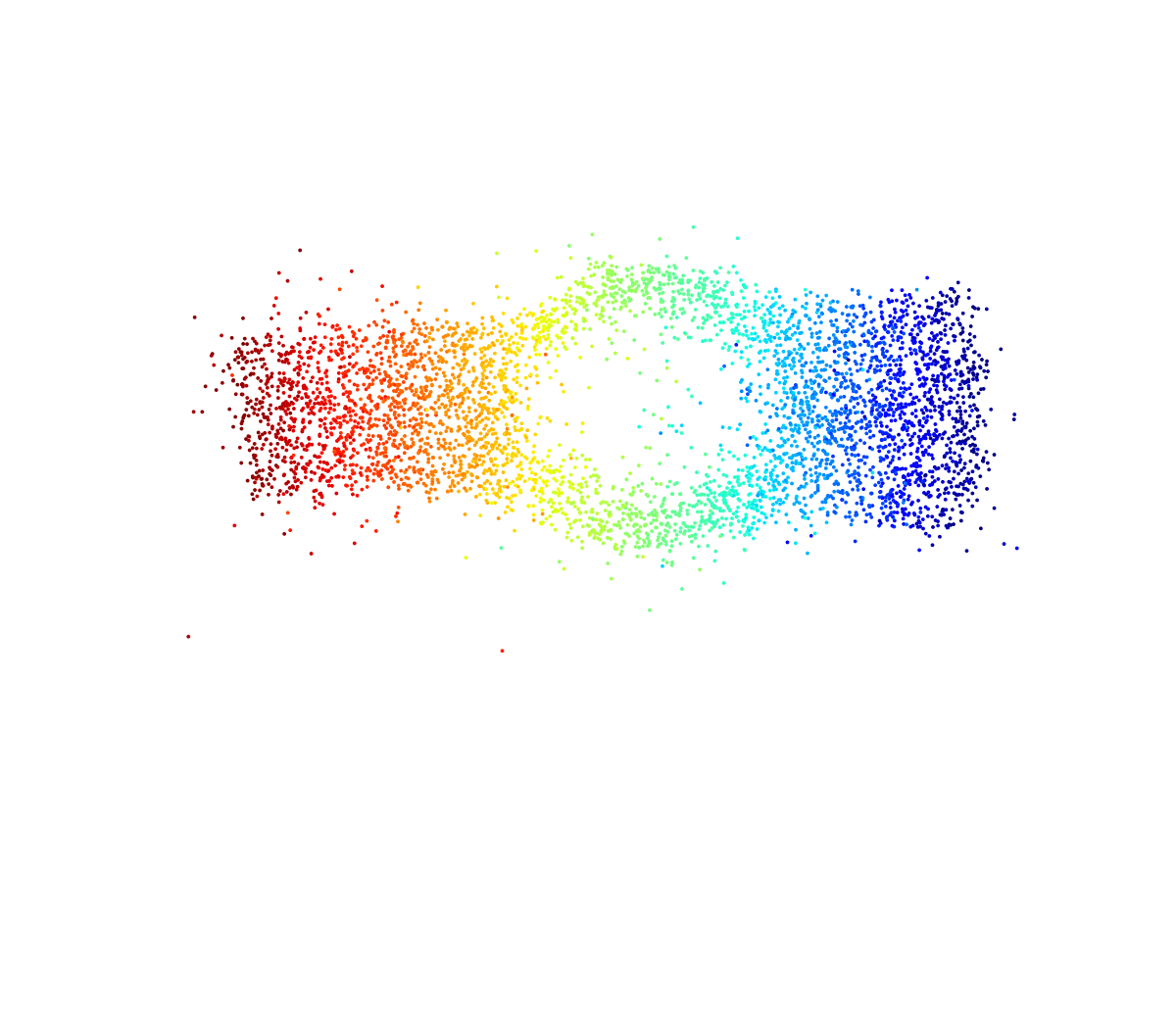} & \includegraphics[width=0.15\textwidth]{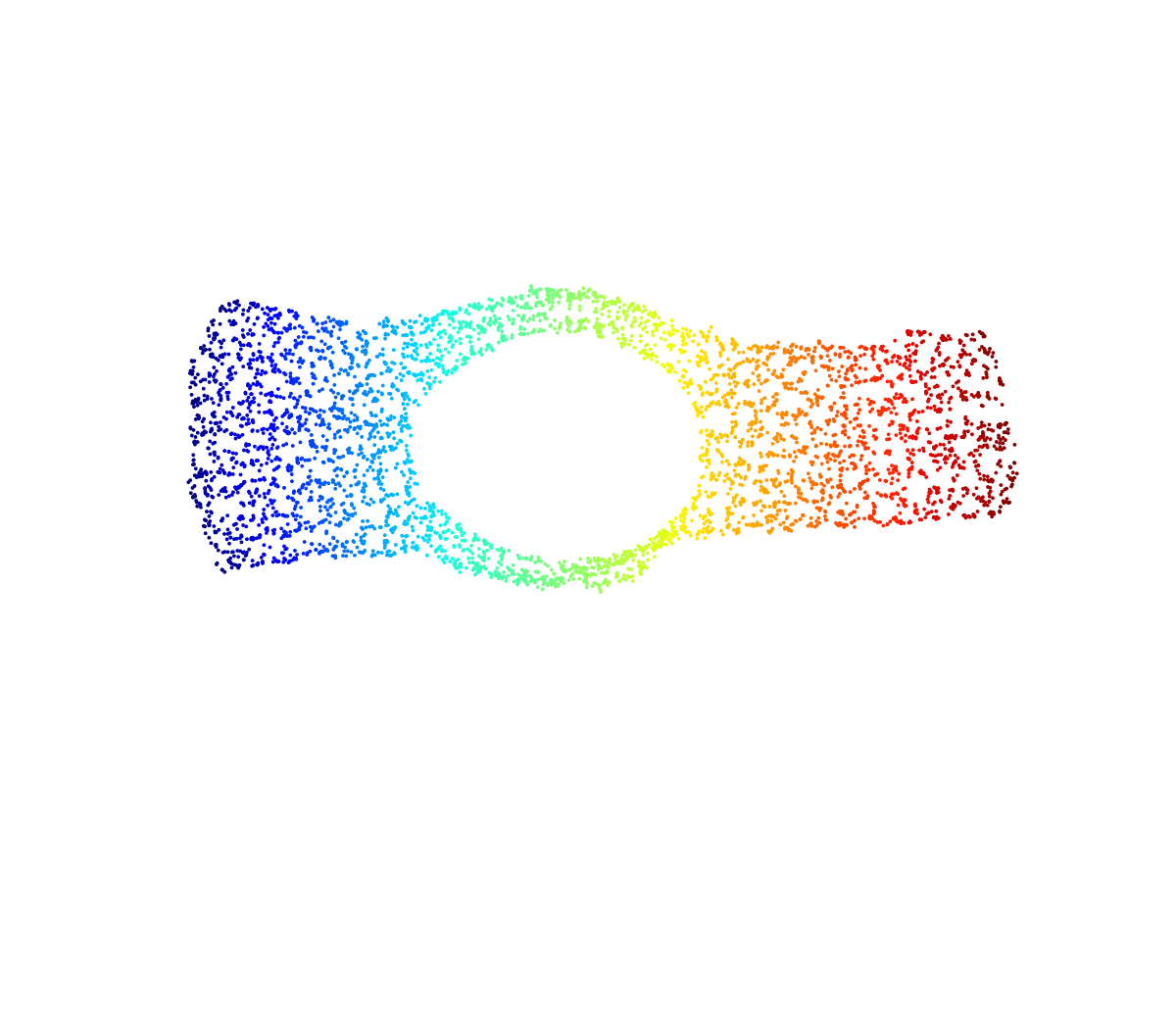} & \includegraphics[width=0.15\textwidth]{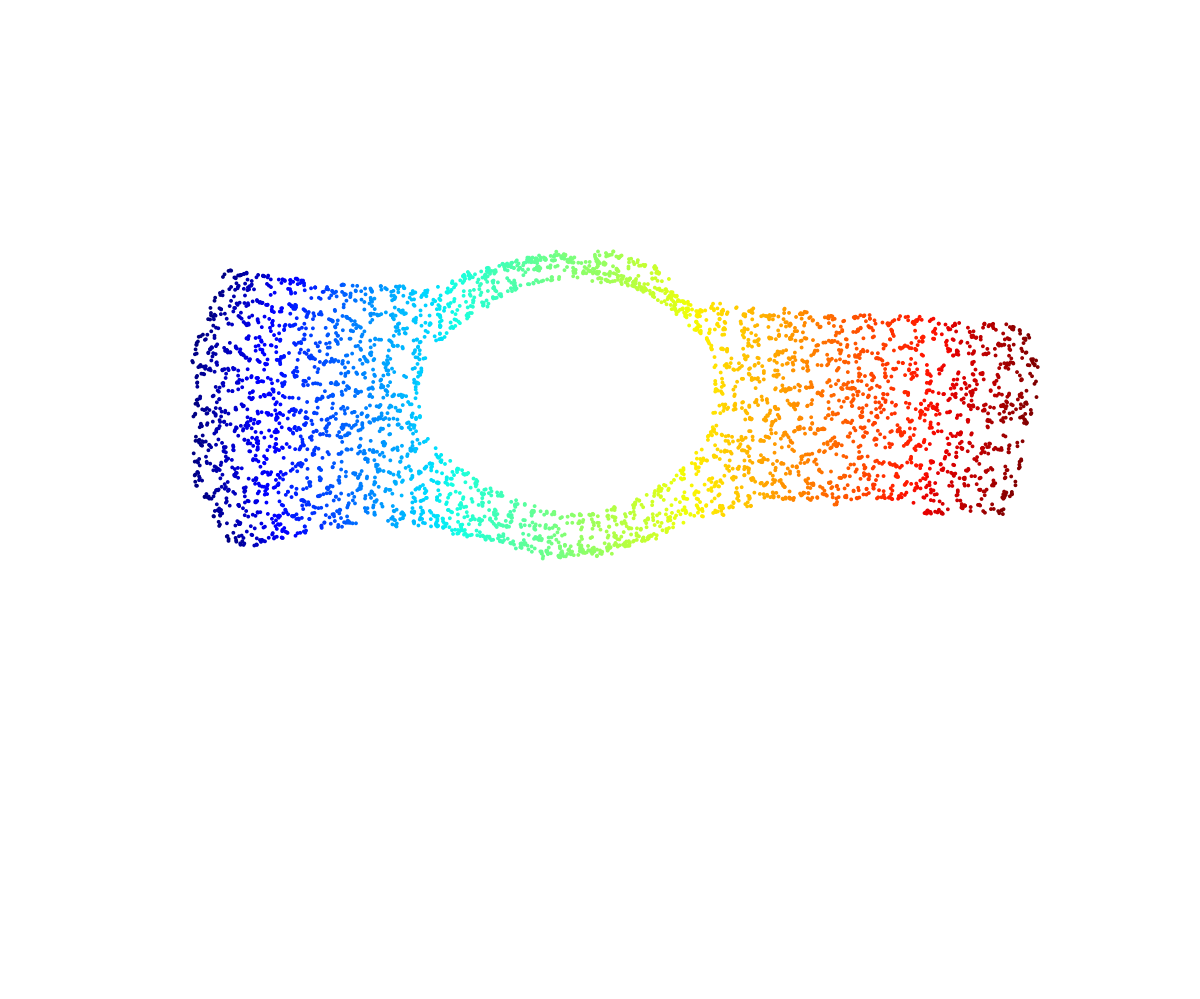} & \includegraphics[width=0.15\textwidth]{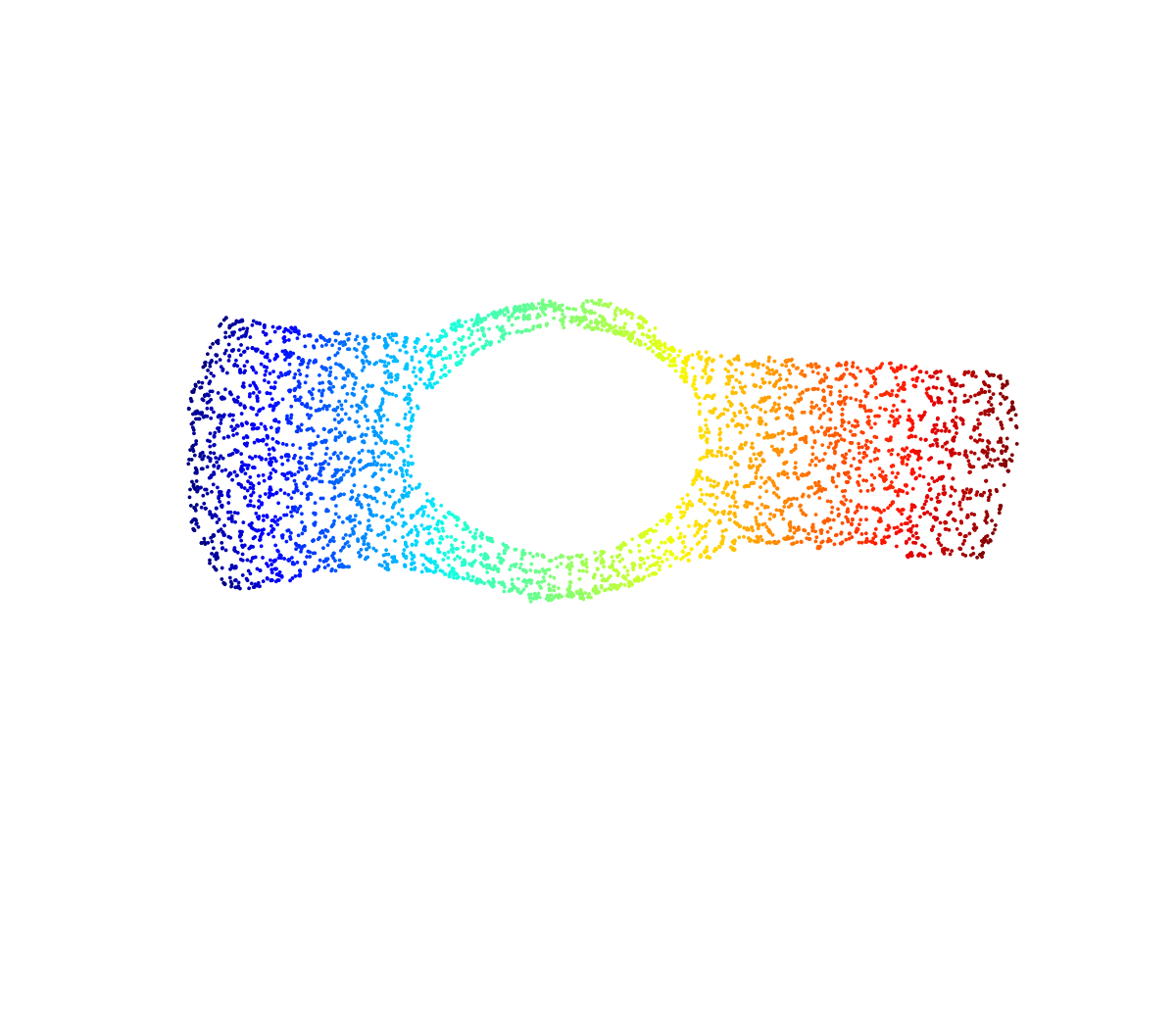} 
		& \includegraphics[width=0.15\textwidth]{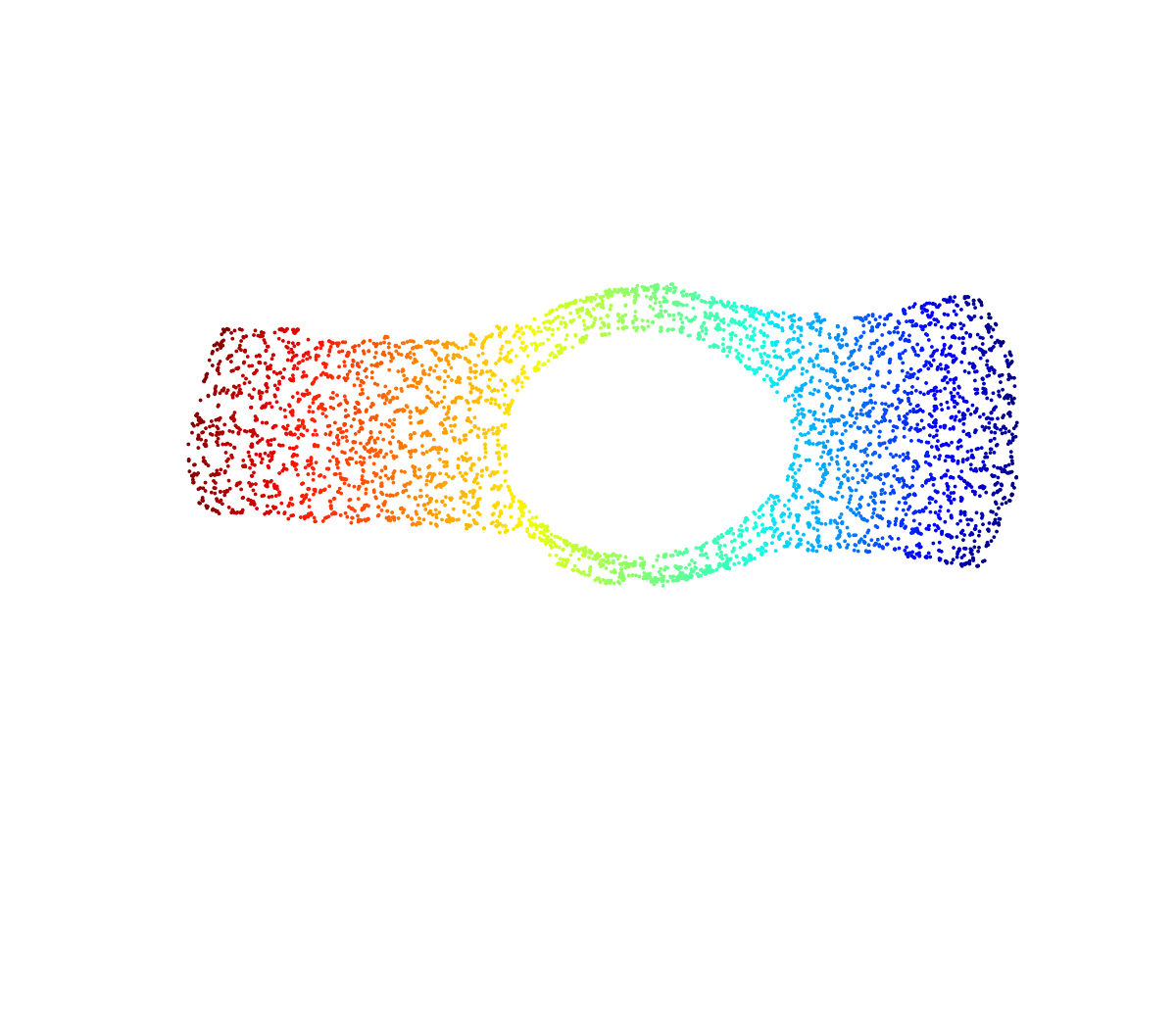} 
		\\
		\textbf{(3) $M_W$} & \includegraphics[width=0.15\textwidth]{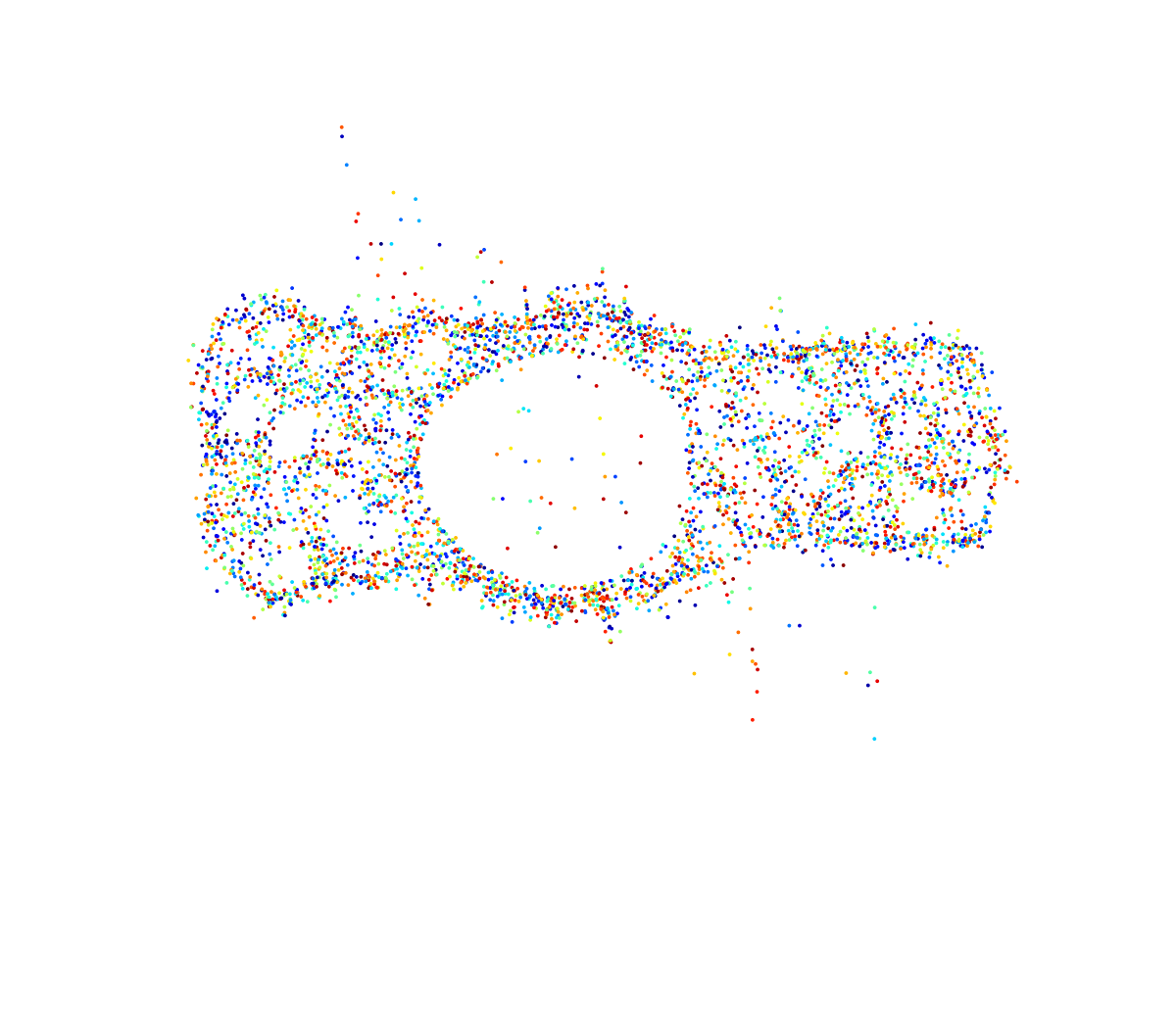} & \includegraphics[width=0.15\textwidth]{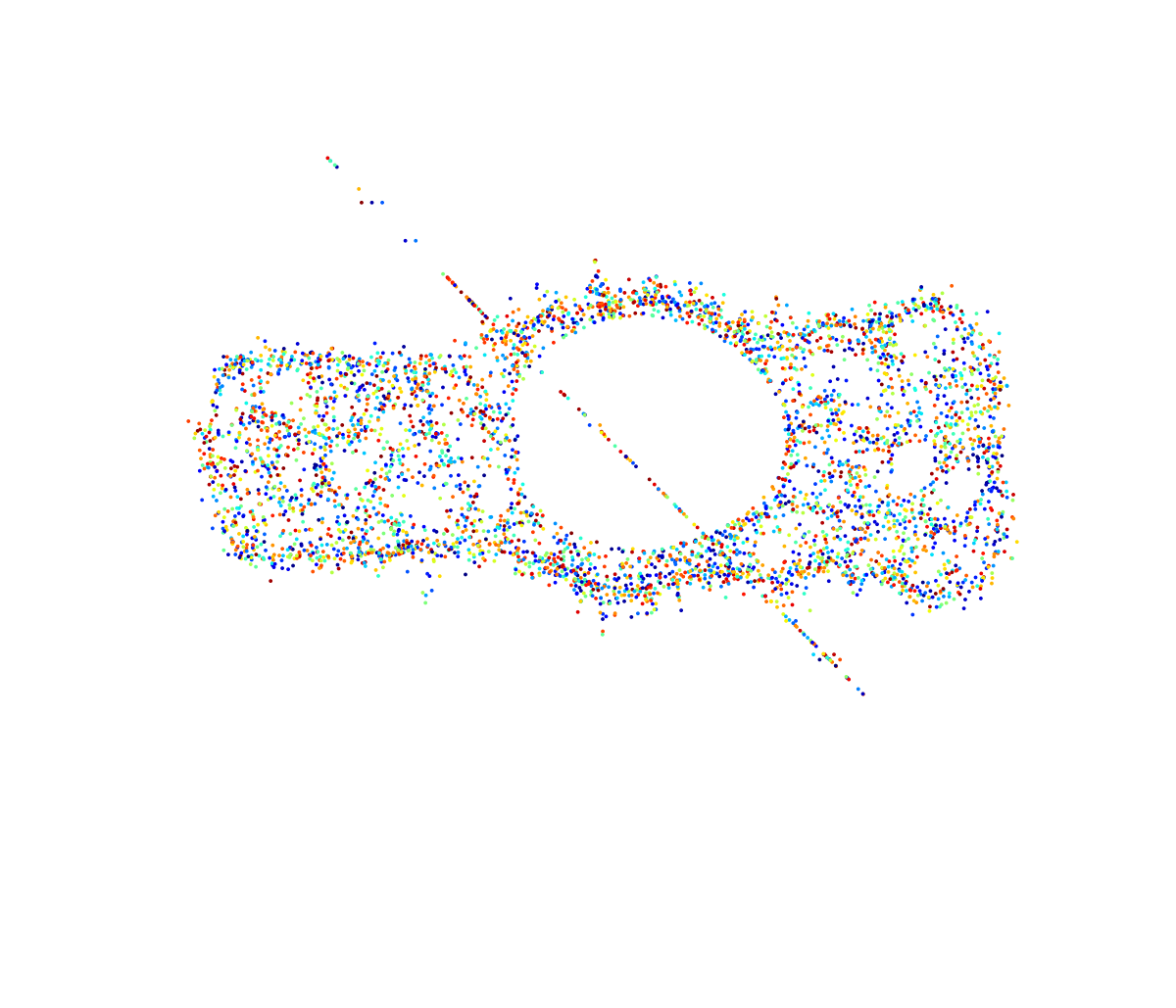} & \includegraphics[width=0.15\textwidth]{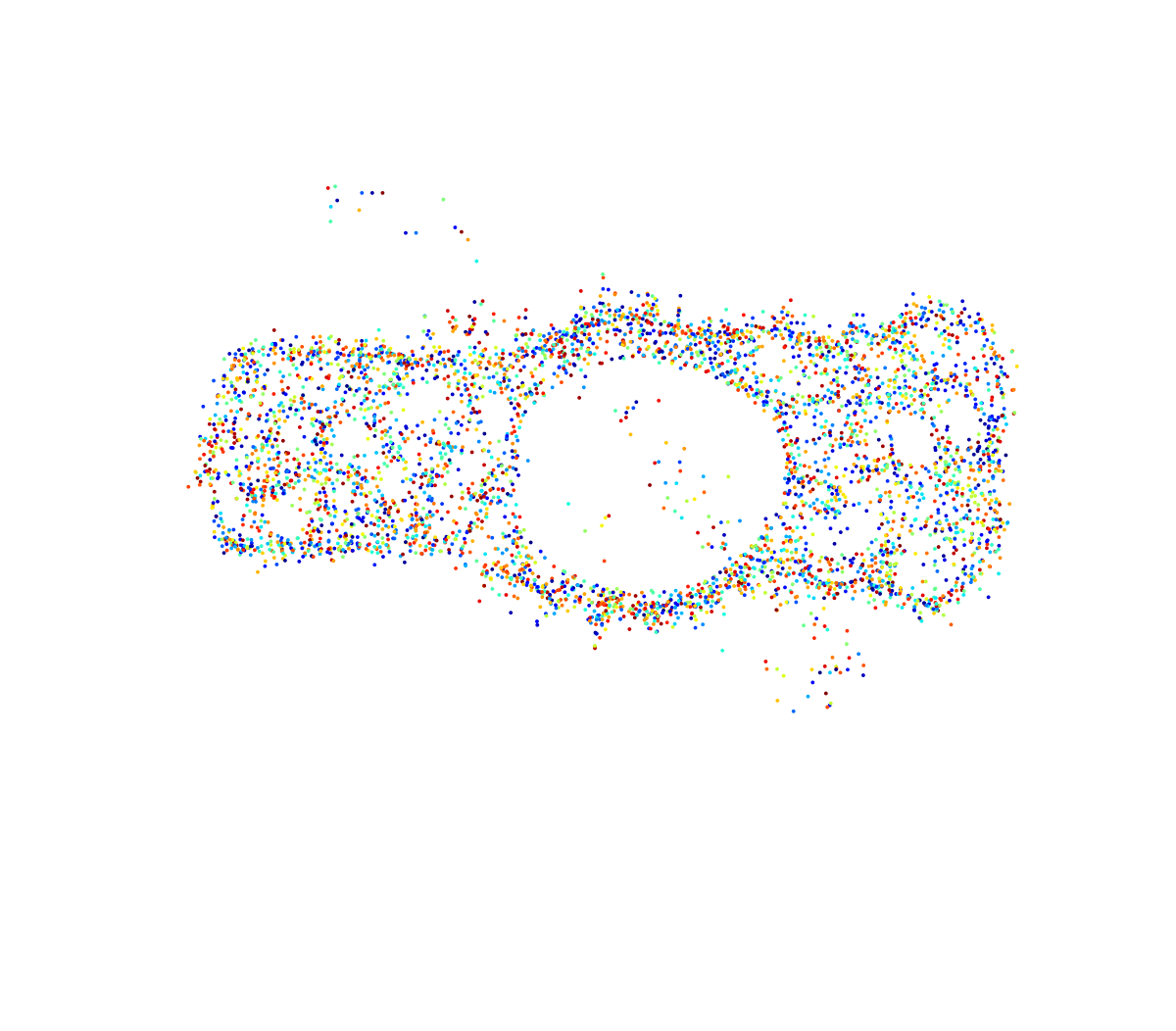} & \includegraphics[width=0.15\textwidth]{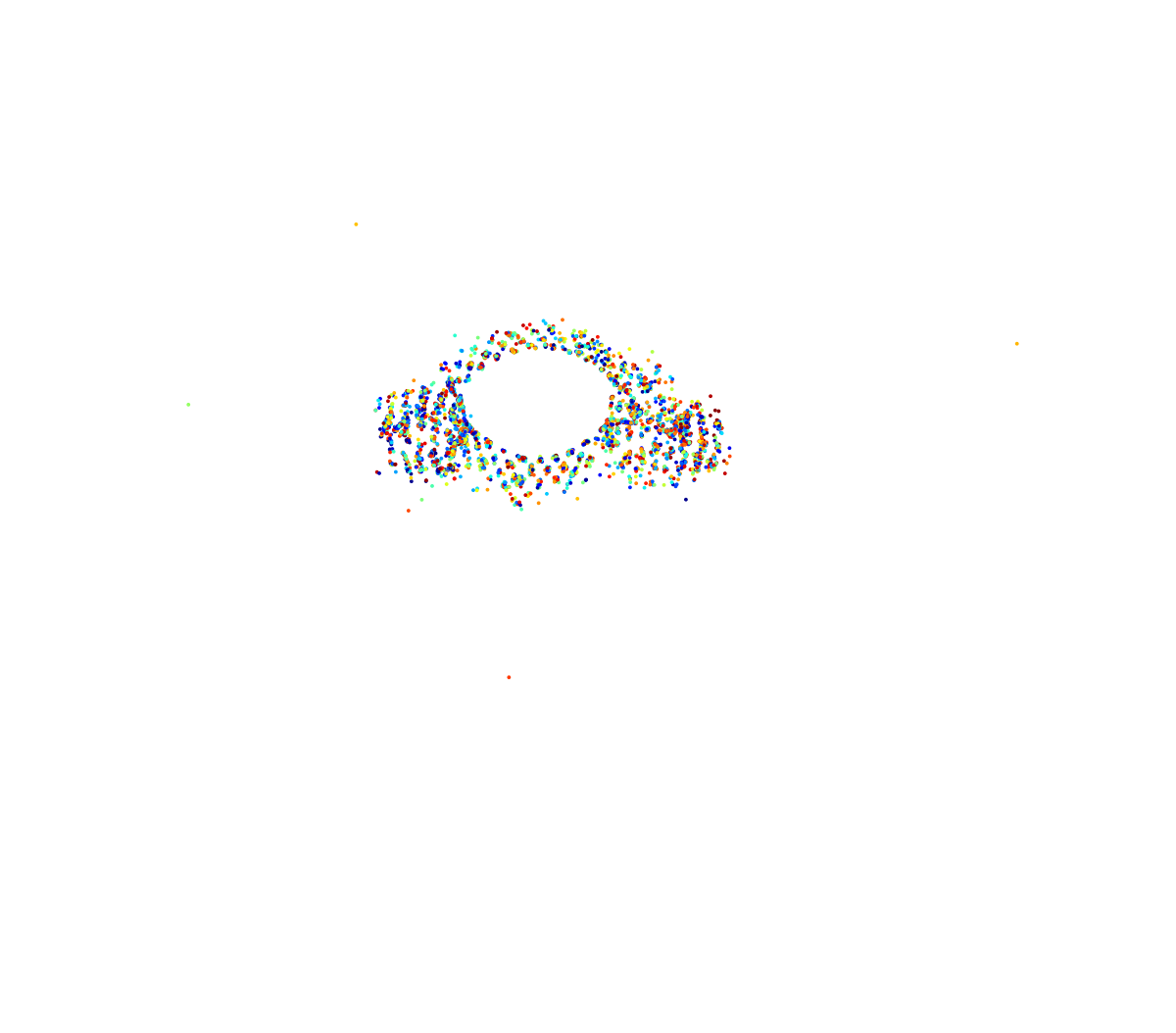} 
		& \includegraphics[width=0.15\textwidth]{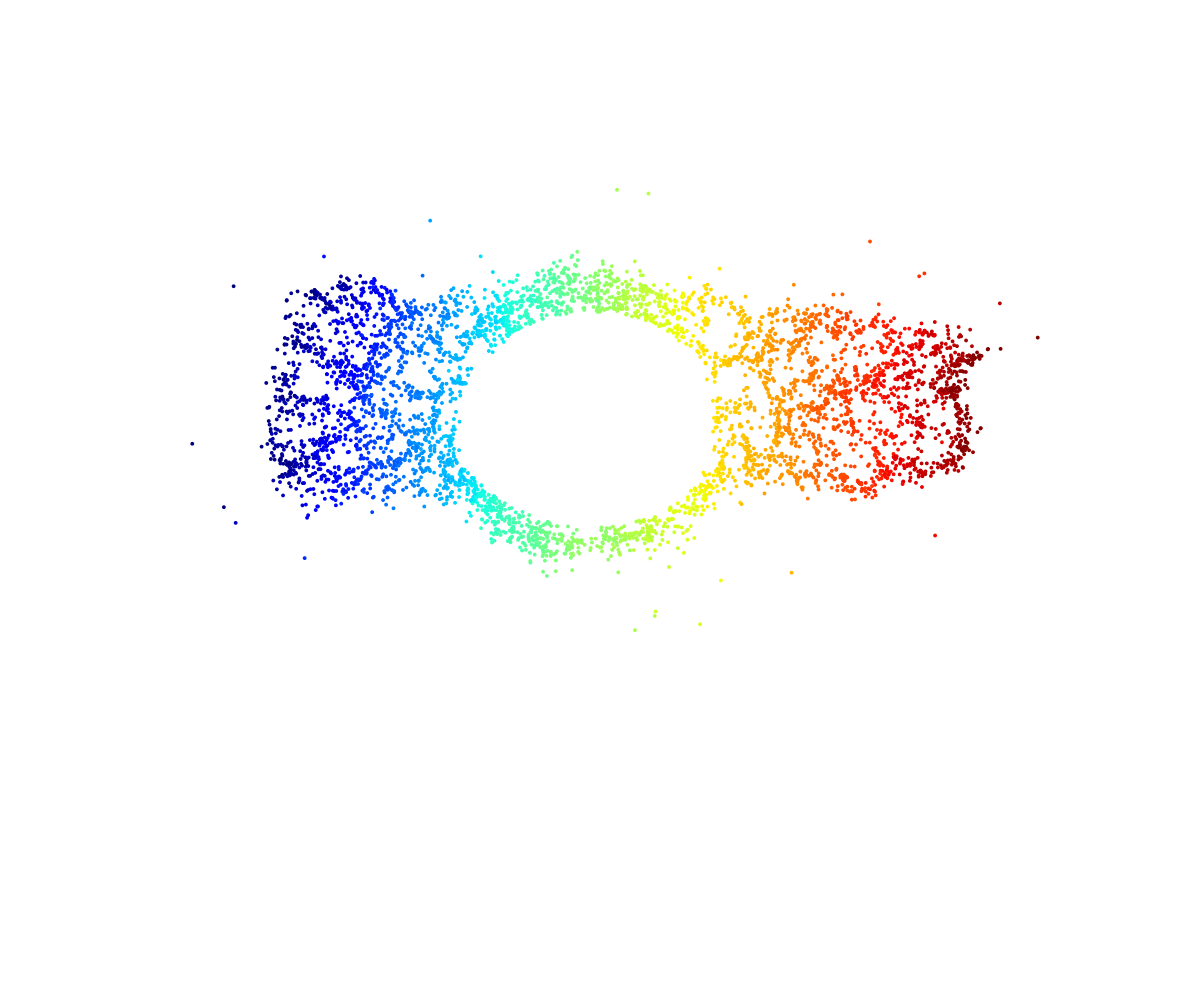}\\
\textbf{(4) $H$} & & & & &\includegraphics[width=0.15\textwidth]{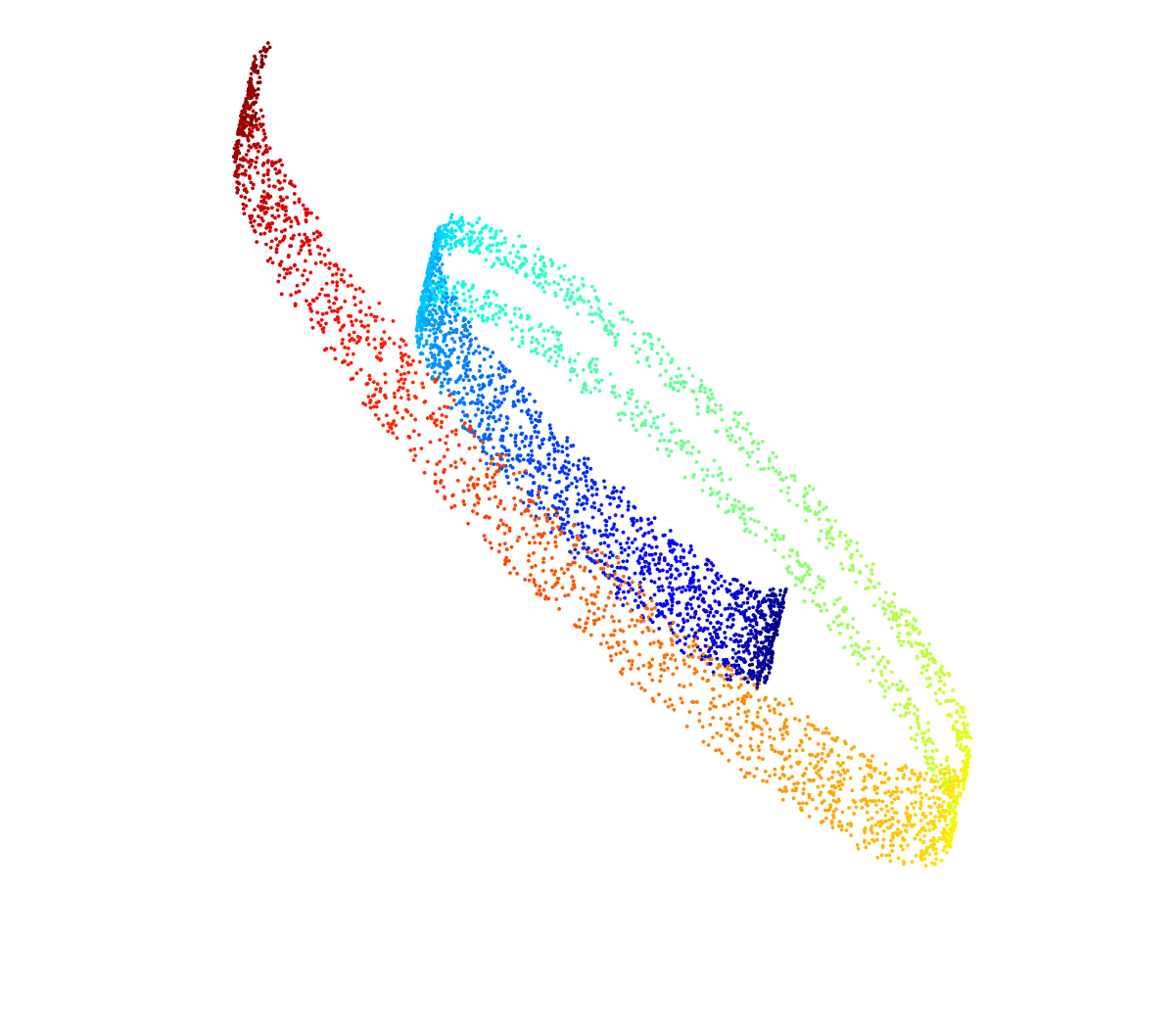} 	\\	
	\end{tabular}
	\caption{Sample 5000 dataset from Swiss Roll, mapped to $2$-d after local normalization of distances and applying $M_V$, $M_{\Pi}$, $M_W$  and $H$ (c.f. \ref{me4}, \ref{productlaw}, \ref{WSlaw}, and \ref{hyplaw} respectively ) with   various corresponding parameters and $k=15$ in neighborhood graph.}
	\label{tab:SwRoll}
\end{table}
\begin{table}[H]
	\centering
	\tiny
	\begin{tabular}{>{\centering\arraybackslash}m{0.5cm}|*{4}{>{\centering\arraybackslash}m{3cm}}}
		 & \textbf{$0.25$} & \textbf{$0.5$}  & \textbf{$0.75$} & \textbf{$1.0$} 
		\\
		& (a) & (b) & (c) & (d)  \\
		\hline \\
		\textbf{(1) $M_V$} & \includegraphics[width=0.17\textwidth]{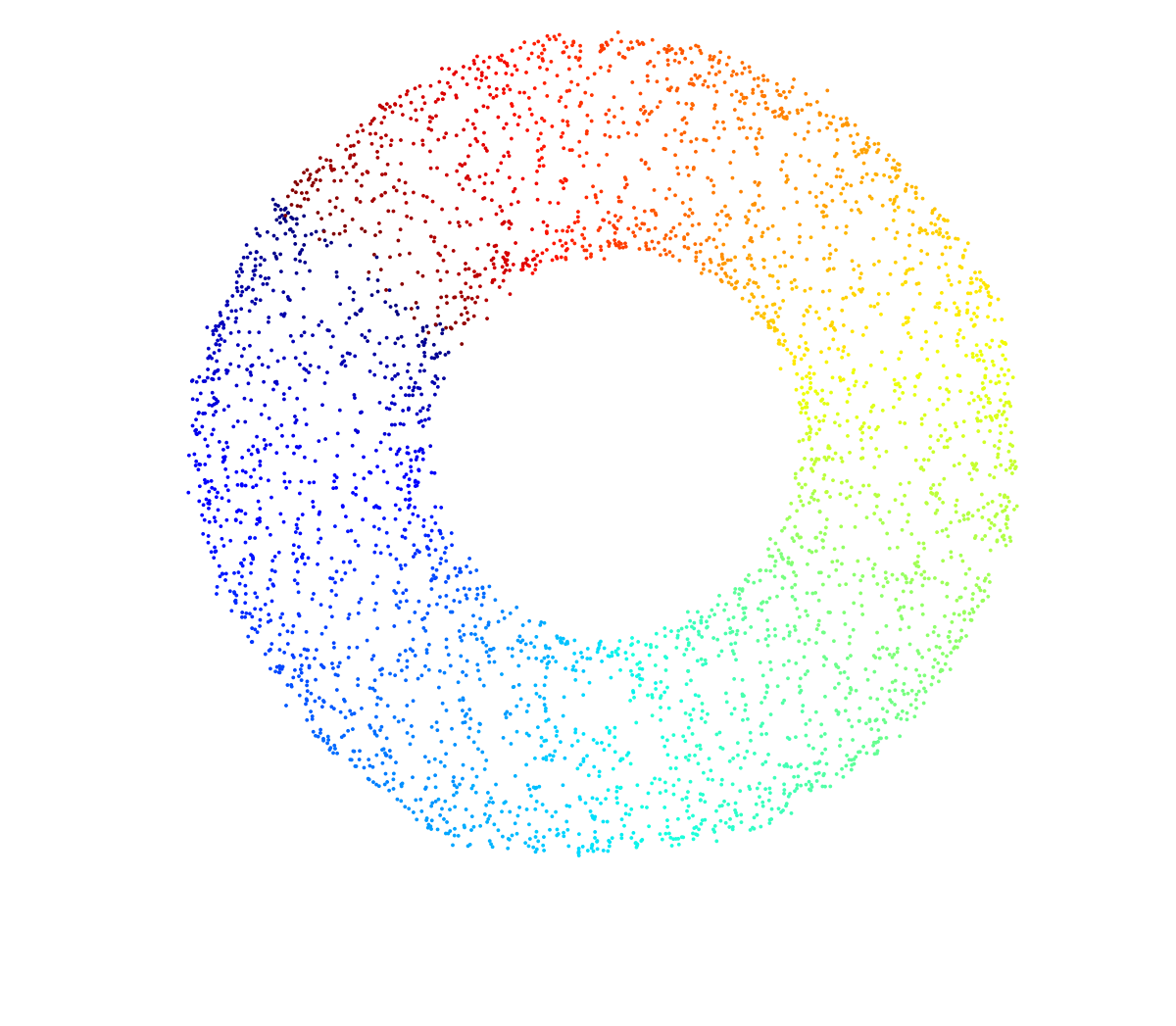} &  \includegraphics[width=0.17\textwidth]{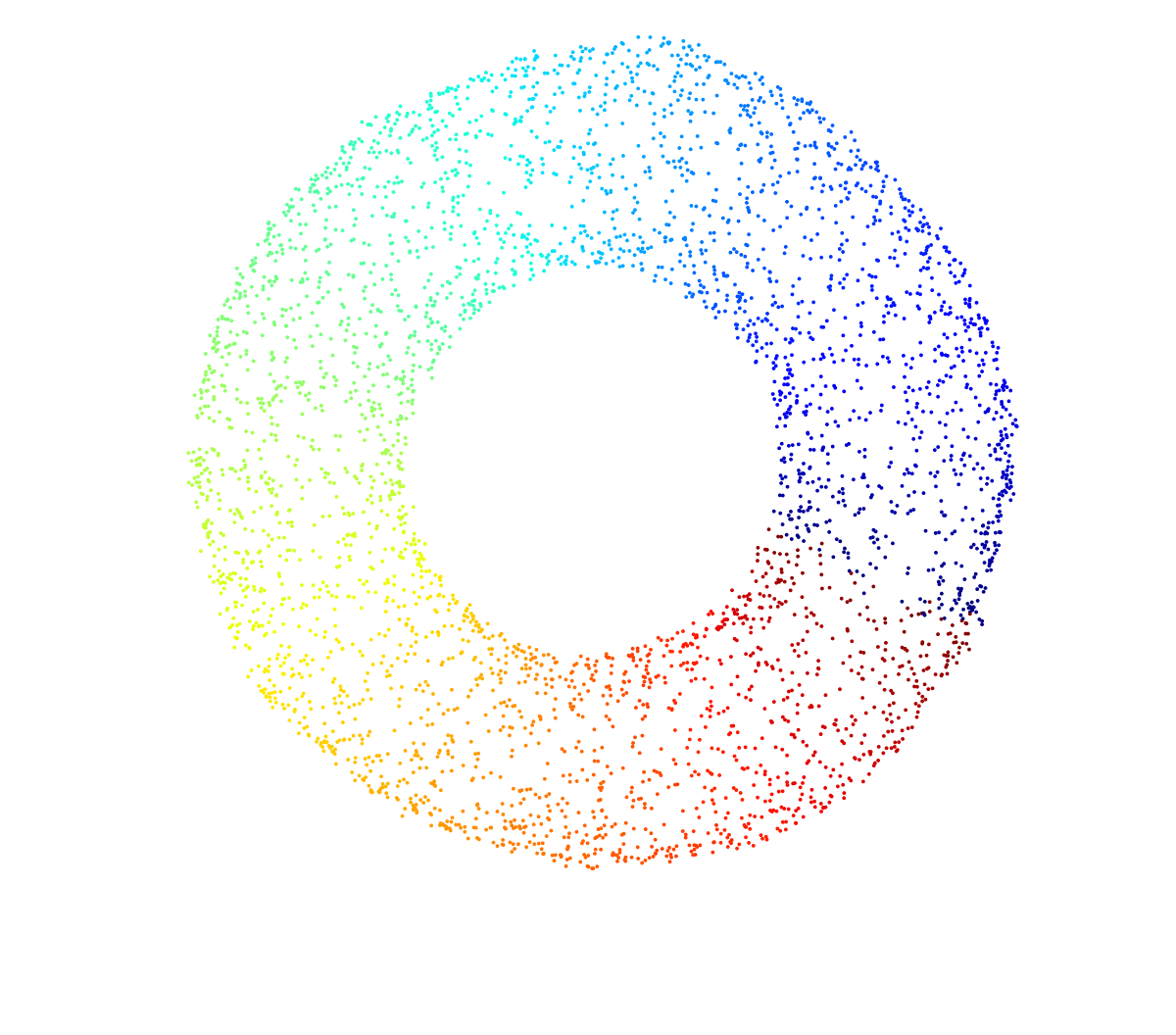} & \includegraphics[width=0.17\textwidth]{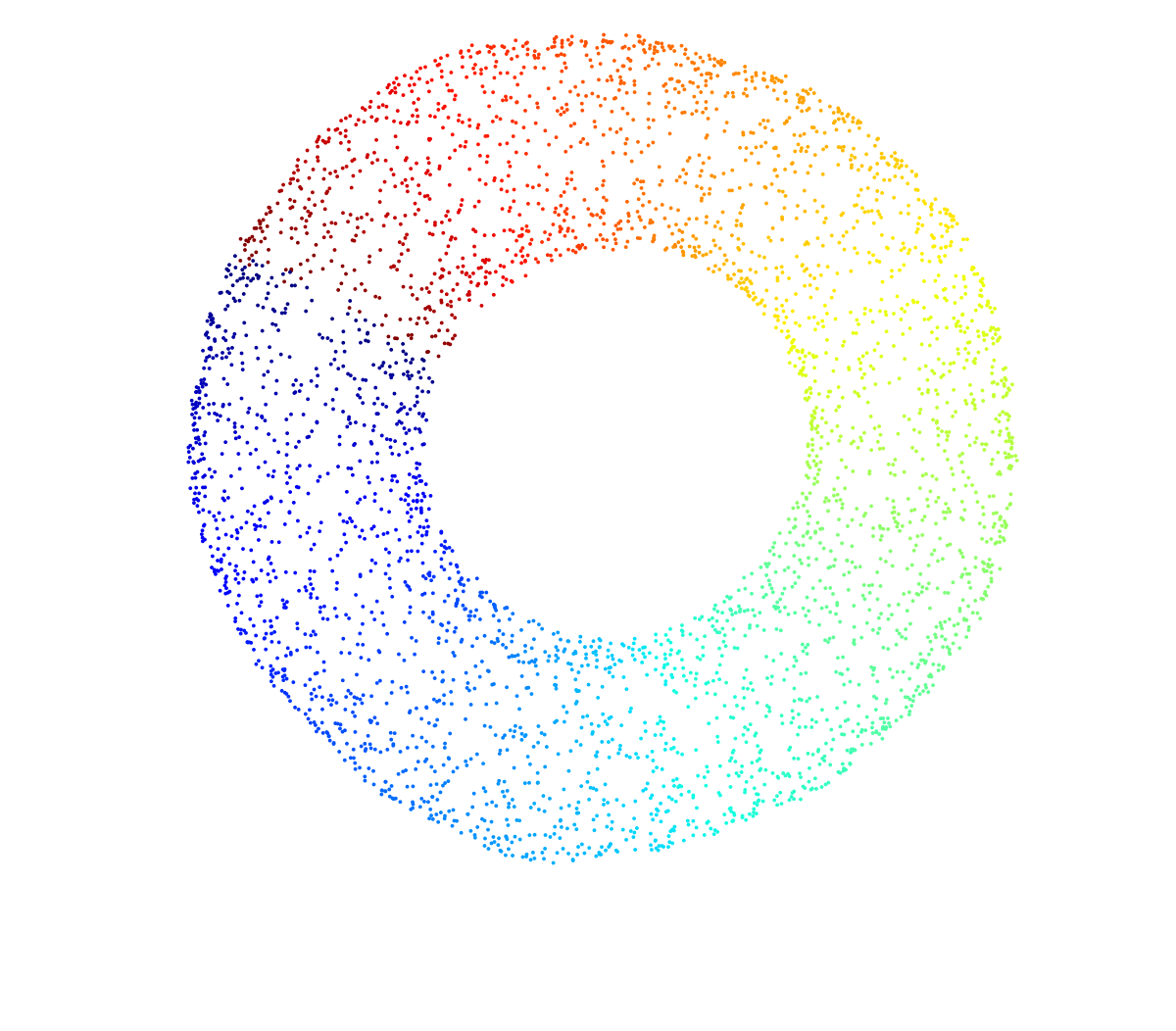}
				&\includegraphics[width=0.17\textwidth]{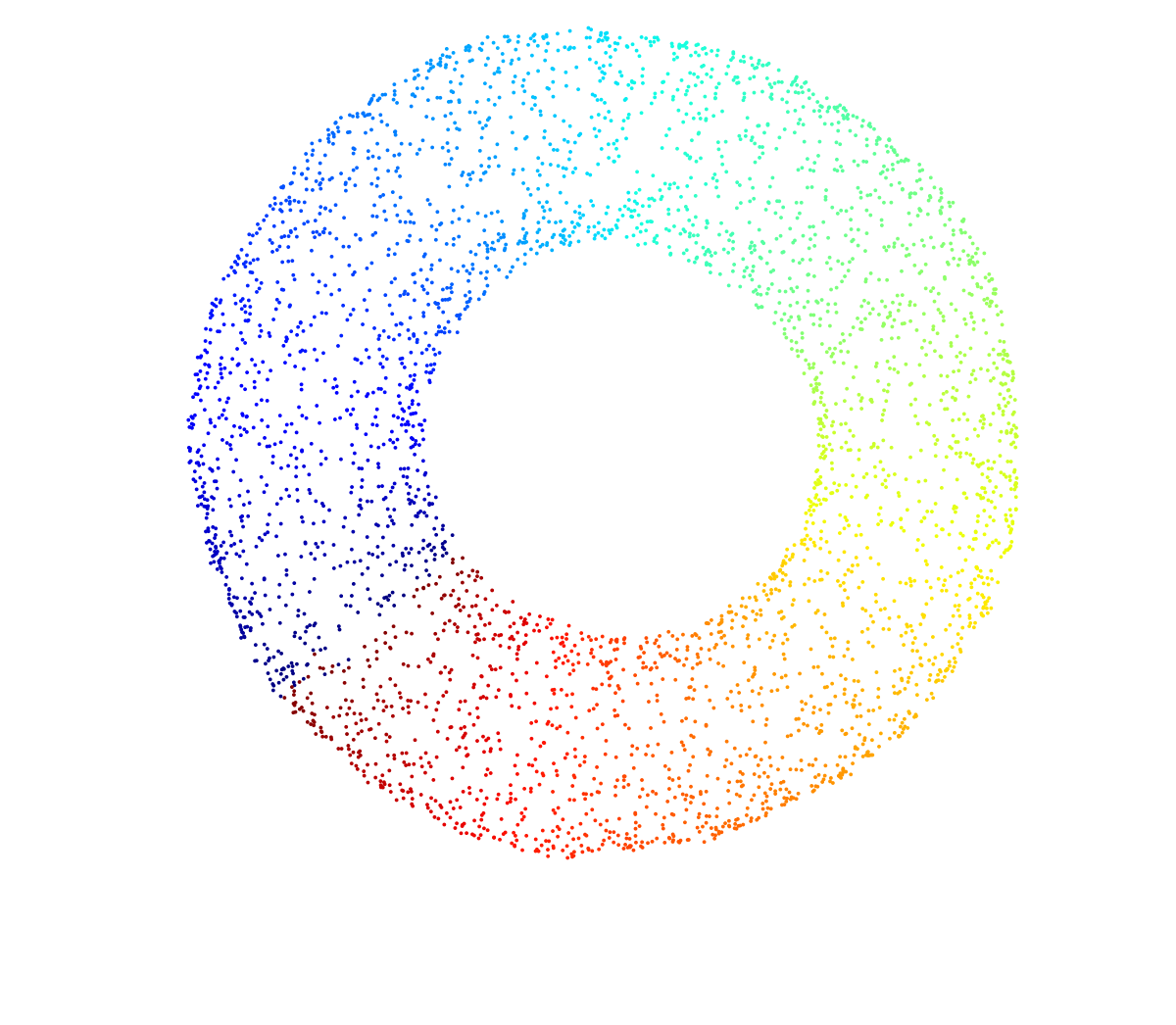} 
		\\
		\textbf{(2) $M_{\Pi}$} & \includegraphics[width=0.17\textwidth]{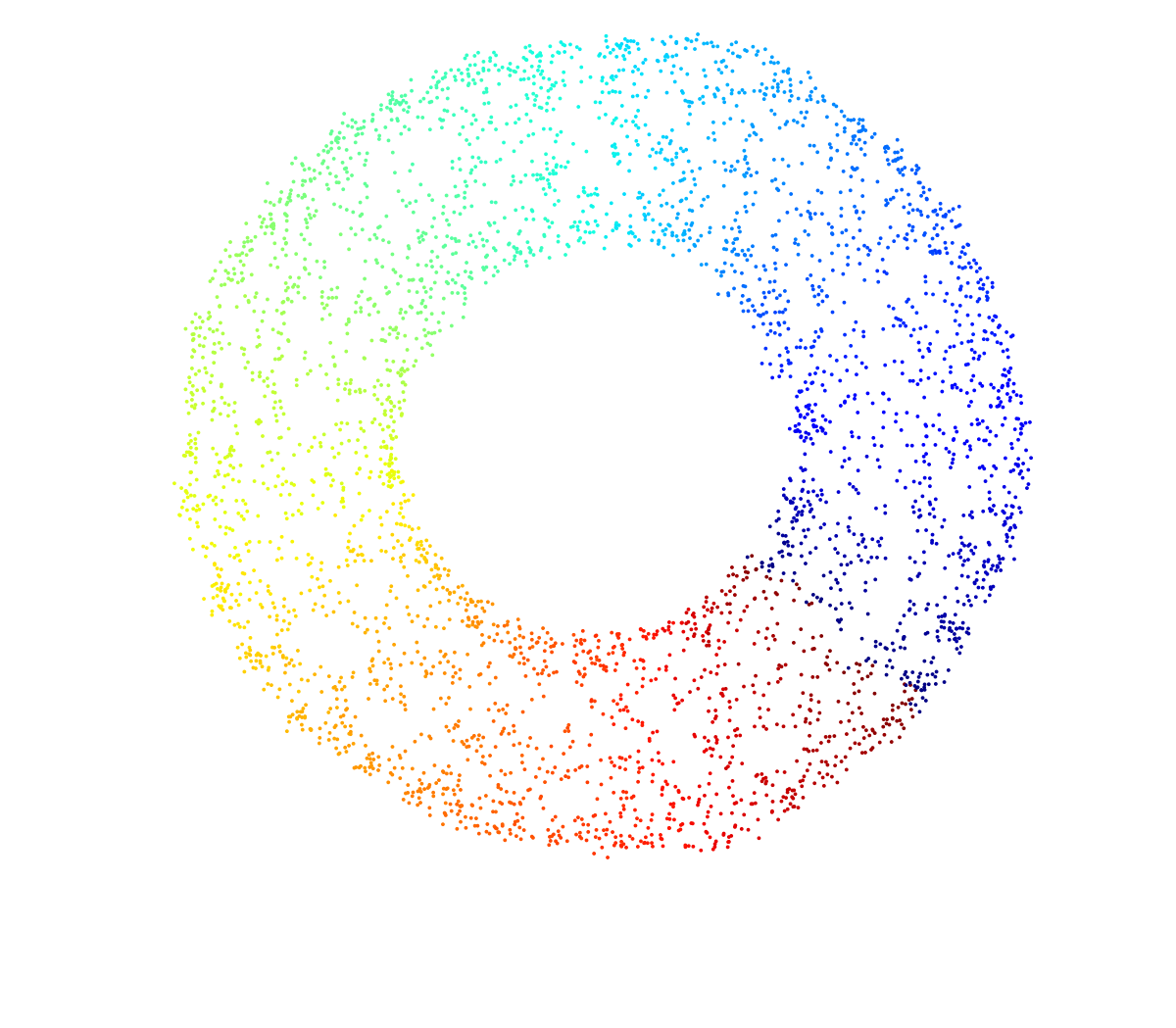} & \includegraphics[width=0.17\textwidth]{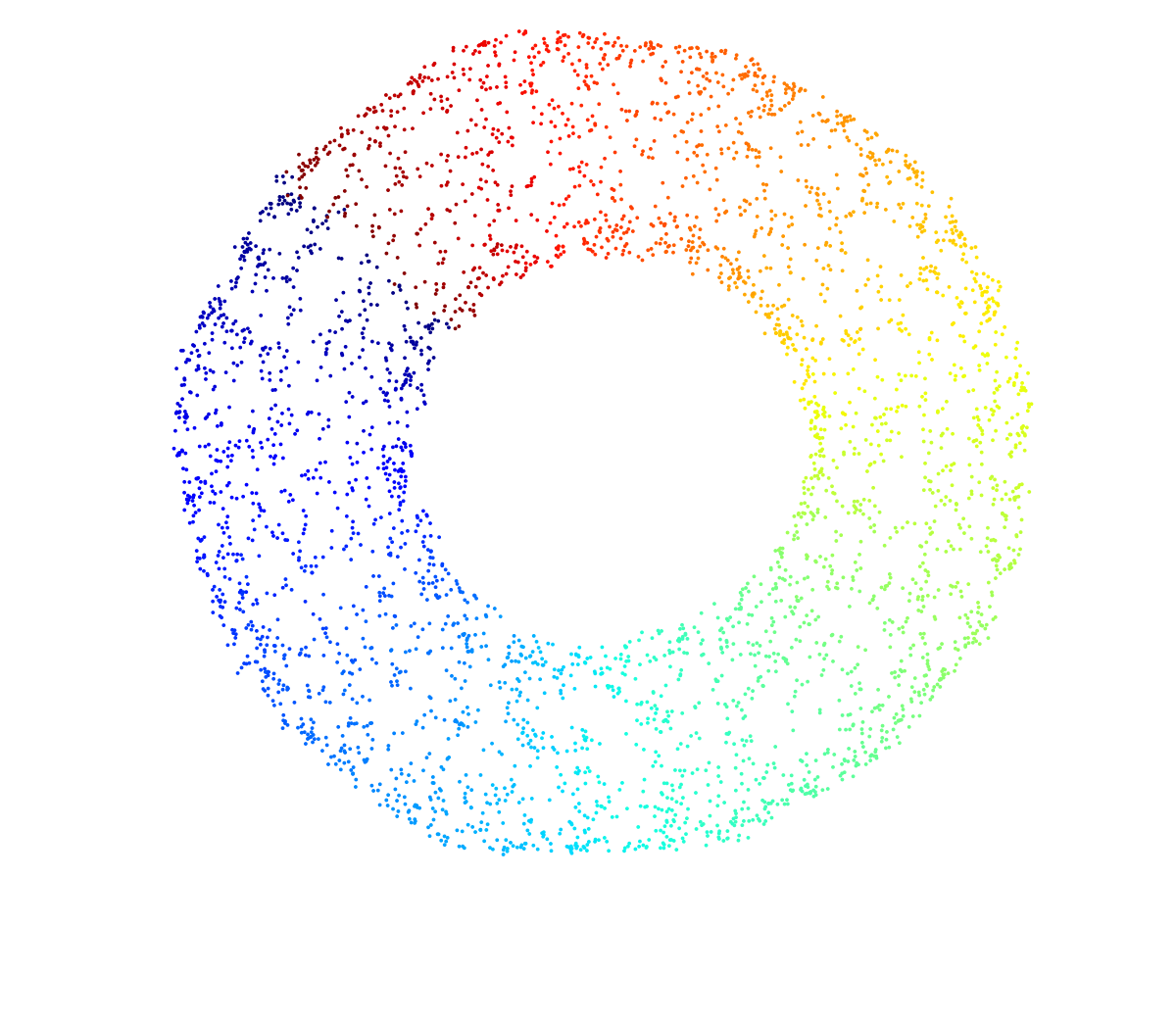} & \includegraphics[width=0.17\textwidth]{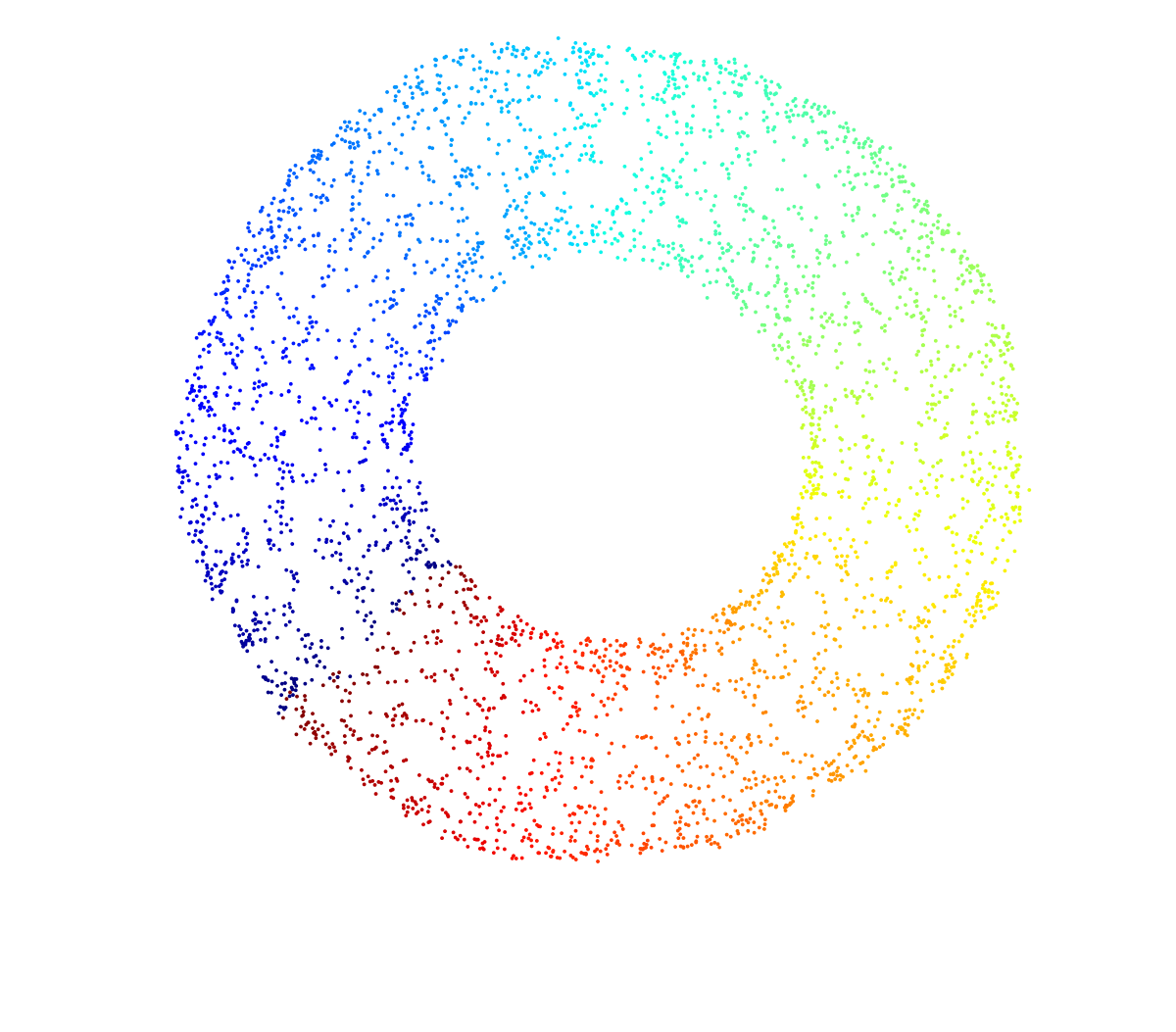} & \includegraphics[width=0.17\textwidth]{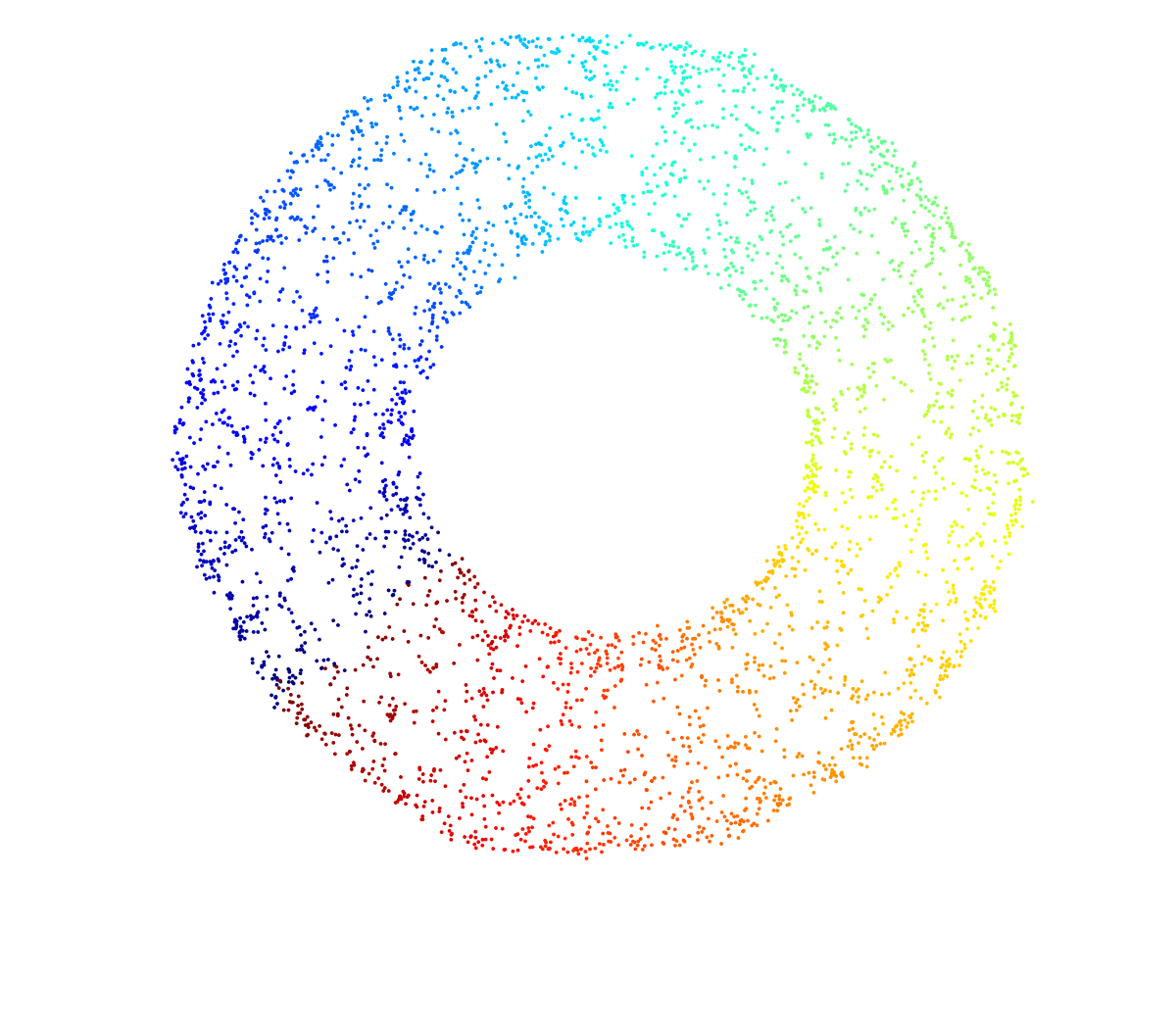} 
				\\
		\textbf{(3) $M_W$} & \includegraphics[width=0.17\textwidth]{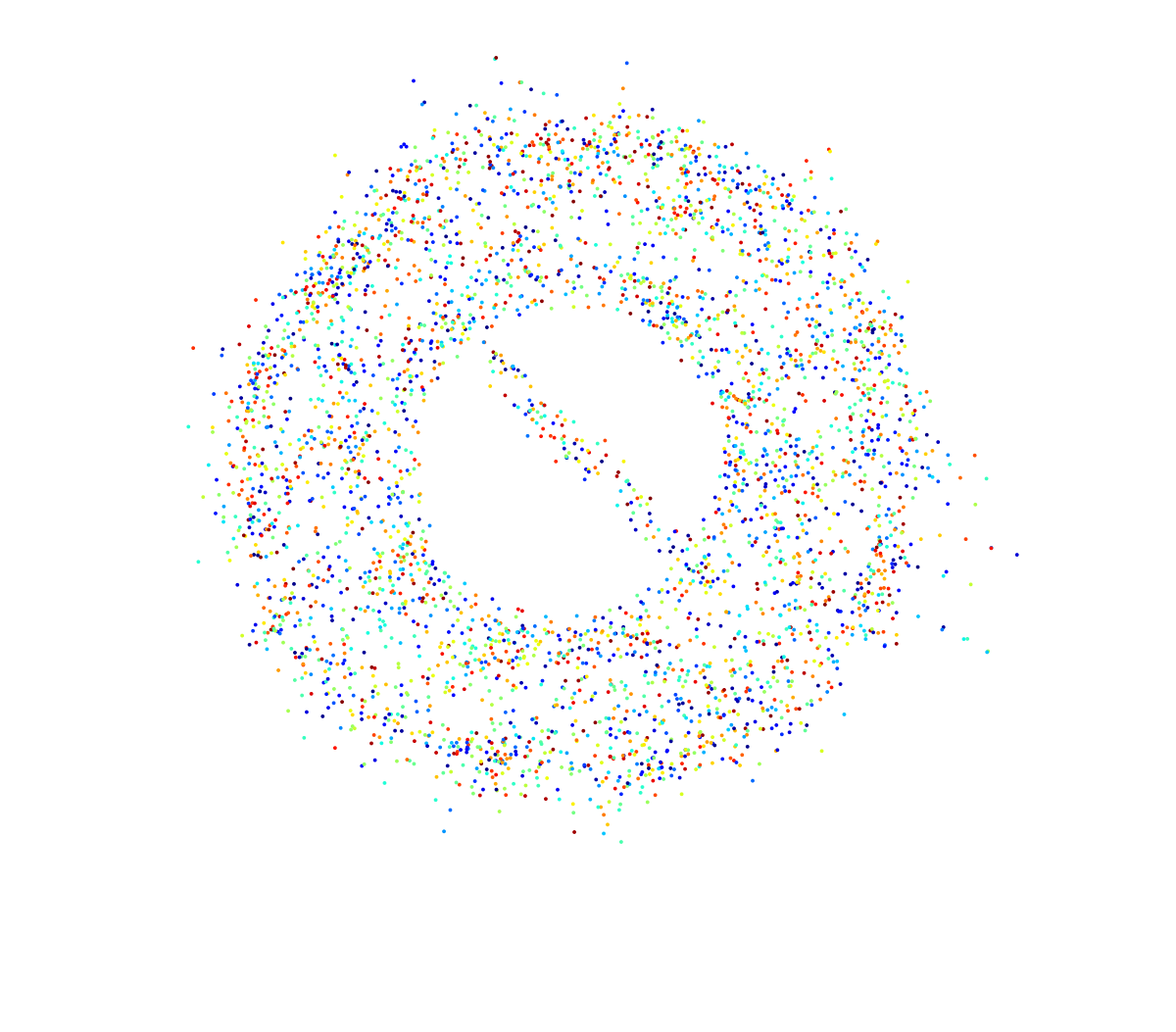} & \includegraphics[width=0.17\textwidth]{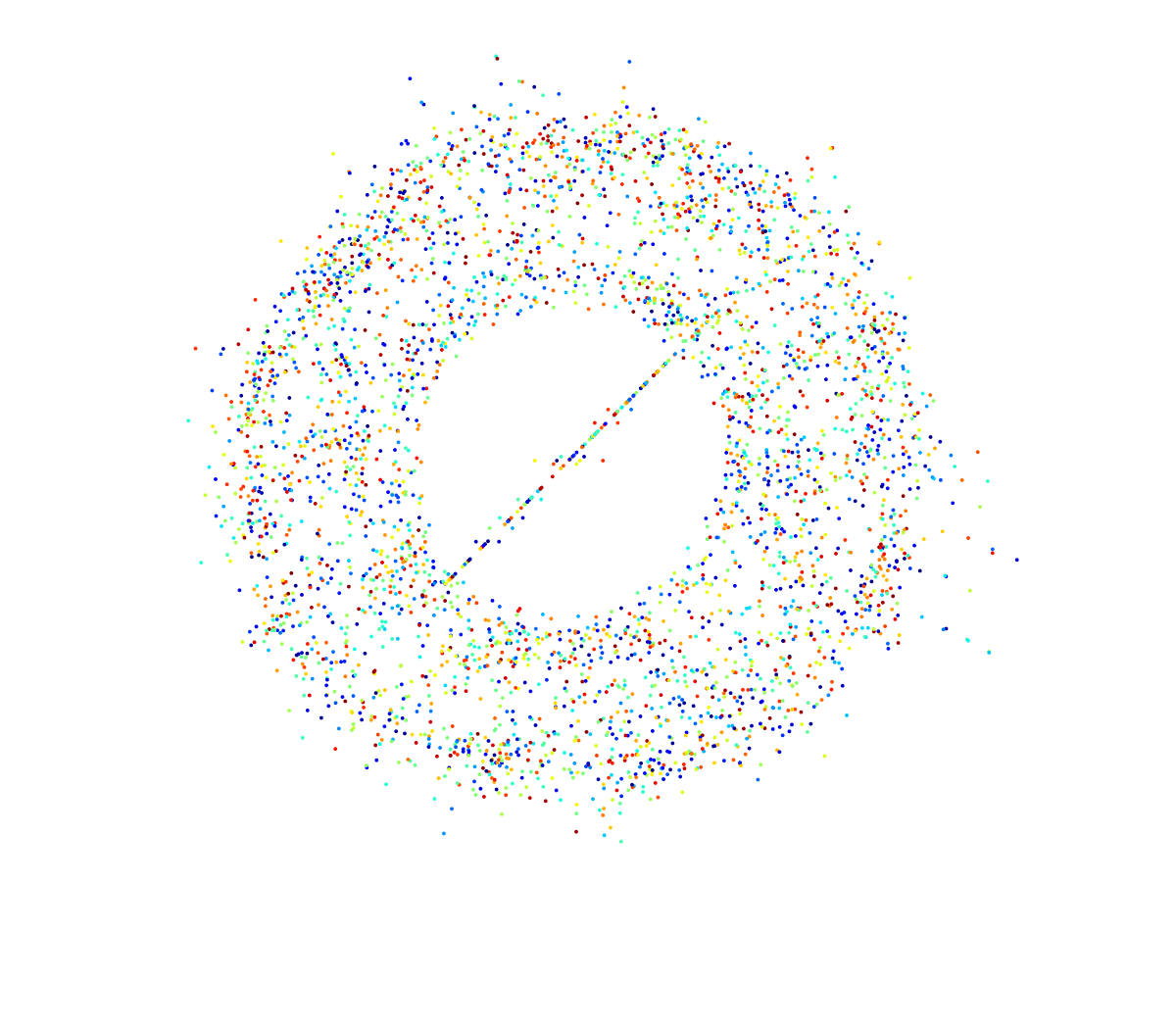} & \includegraphics[width=0.17\textwidth]{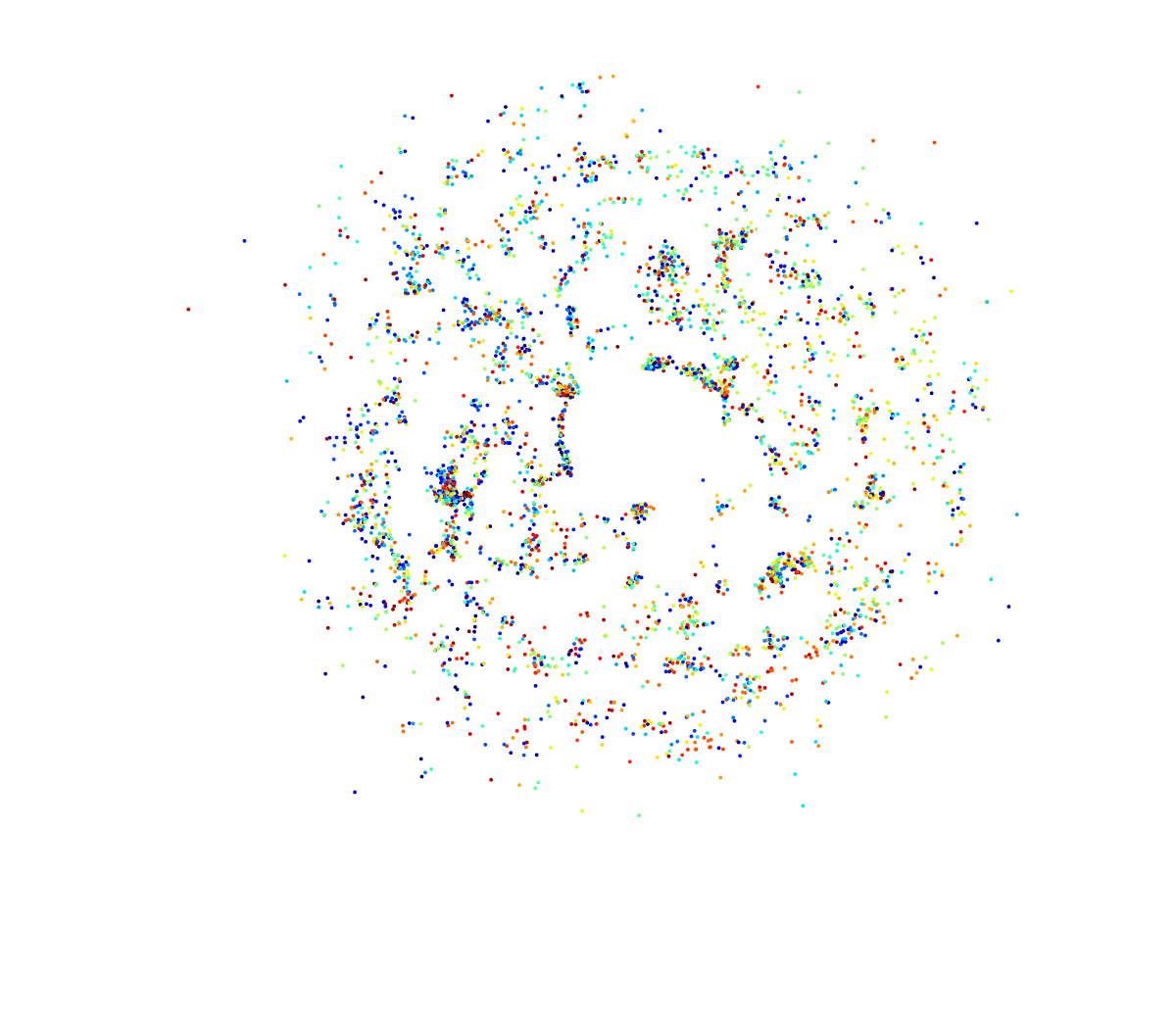} & \includegraphics[width=0.17\textwidth]{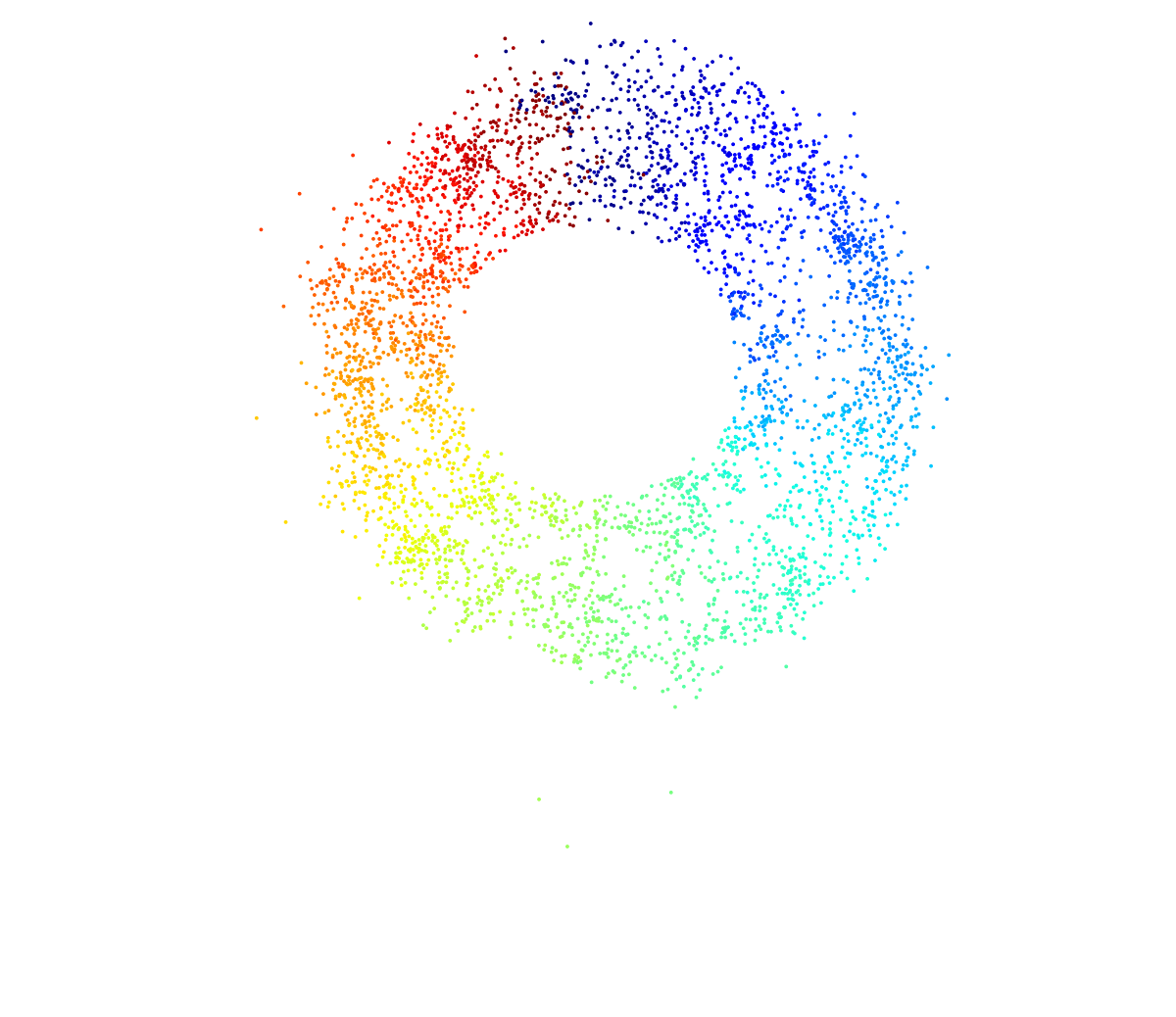}\\ 
		\textbf{(4) $H$} & & & &\includegraphics[width=0.17\textwidth]{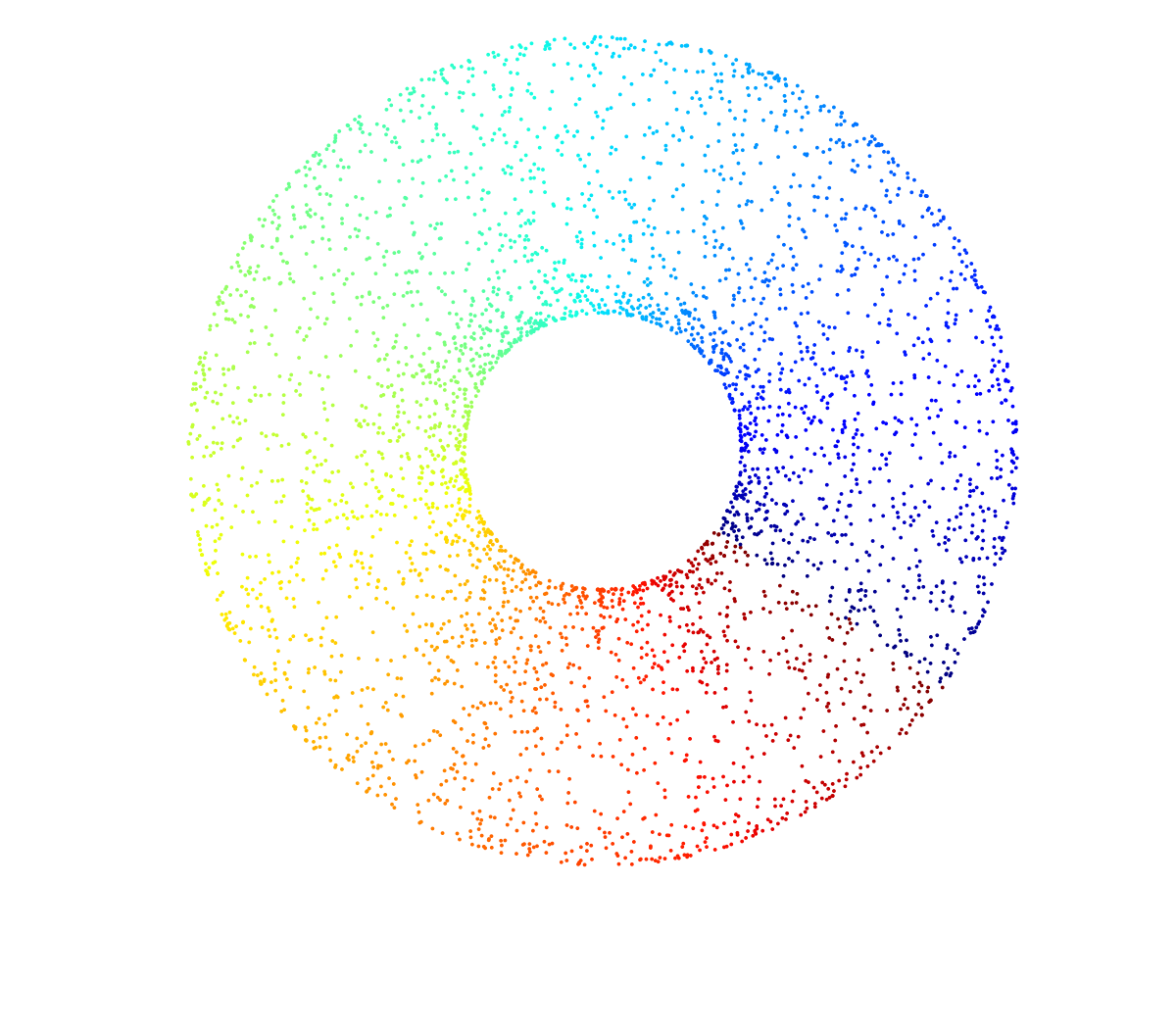}\\
			\end{tabular}
	\caption{Sample 4000 dataset from Torus, mapped to $2$-d after local normalization of distances and applying $M_V$, $M_{\Pi}$, $M_W$, and $H$ (c.f. \ref{me4}, \ref{productlaw}, \ref{WSlaw}, and \ref{hyplaw} respectively ) with   various corresponding parameters and $k=15$ in neighborhood graph.}
	\label{tab:Torus}
\end{table}
\begin{table}[H]
	\centering
	\tiny
	\begin{tabular}{>{\centering\arraybackslash}m{0.3cm}|*{5}{>{\centering\arraybackslash}m{2.5cm}}}
		& \textbf{$0.01$} & \textbf{$0.25$} & \textbf{$0.5$}  & \textbf{$0.75$} & \textbf{$1.0$} 
		\\
		& (a) & (b) & (c) & (d) & (e) \\
		\hline \\
		\textbf{(1) $M_V$} & \includegraphics[width=0.15\textwidth]{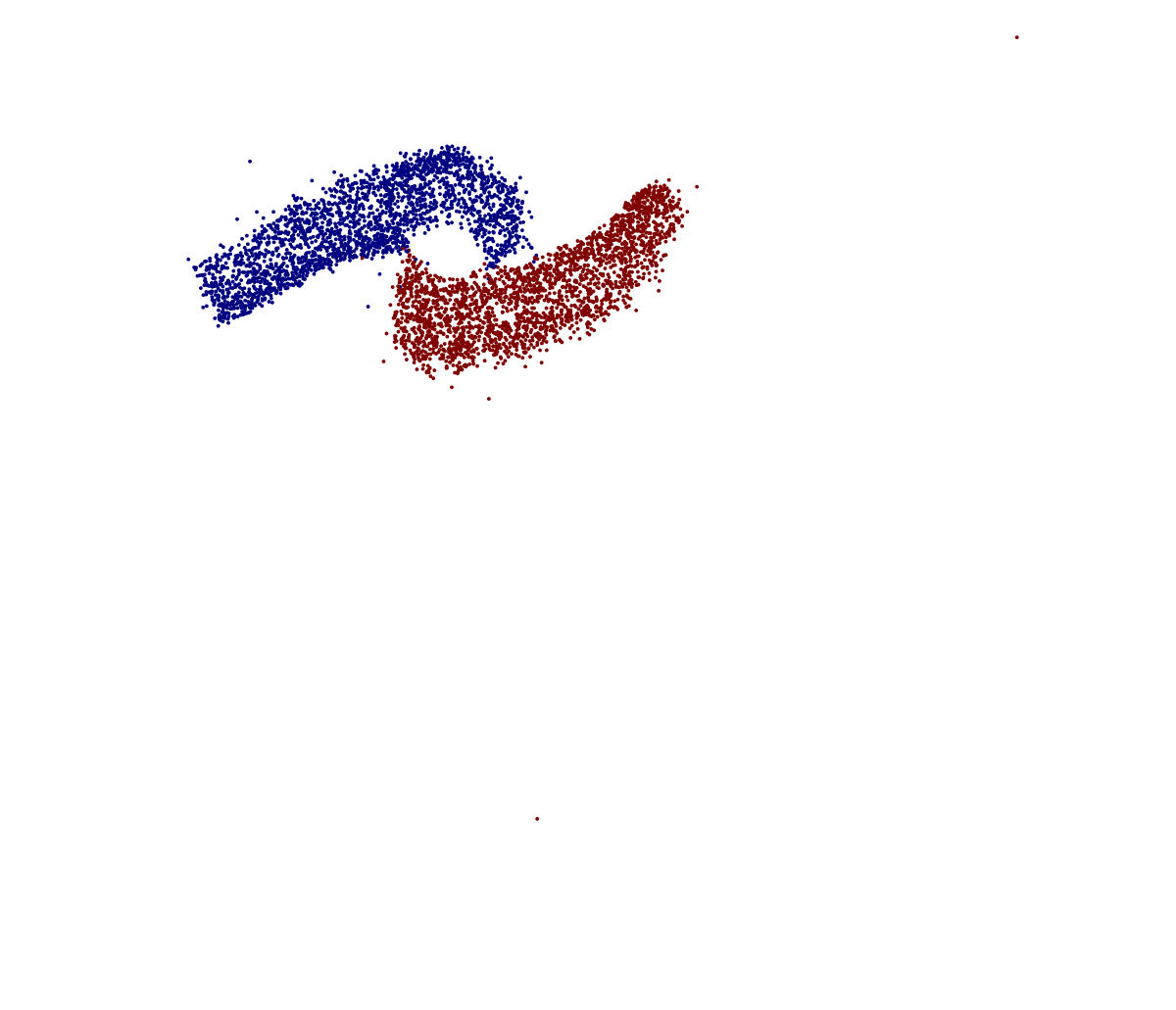} &  \includegraphics[width=0.15\textwidth]{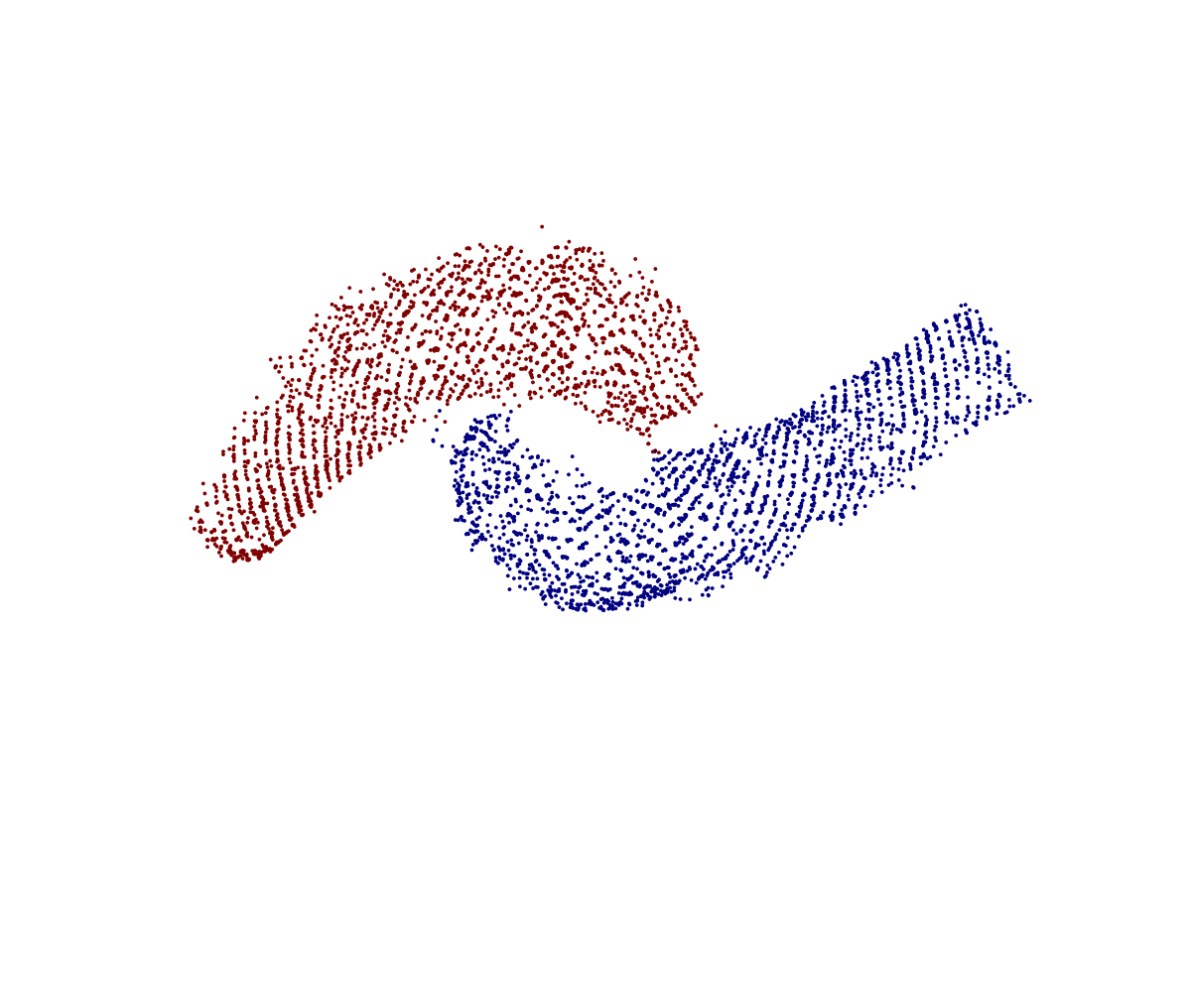} & \includegraphics[width=0.15\textwidth]{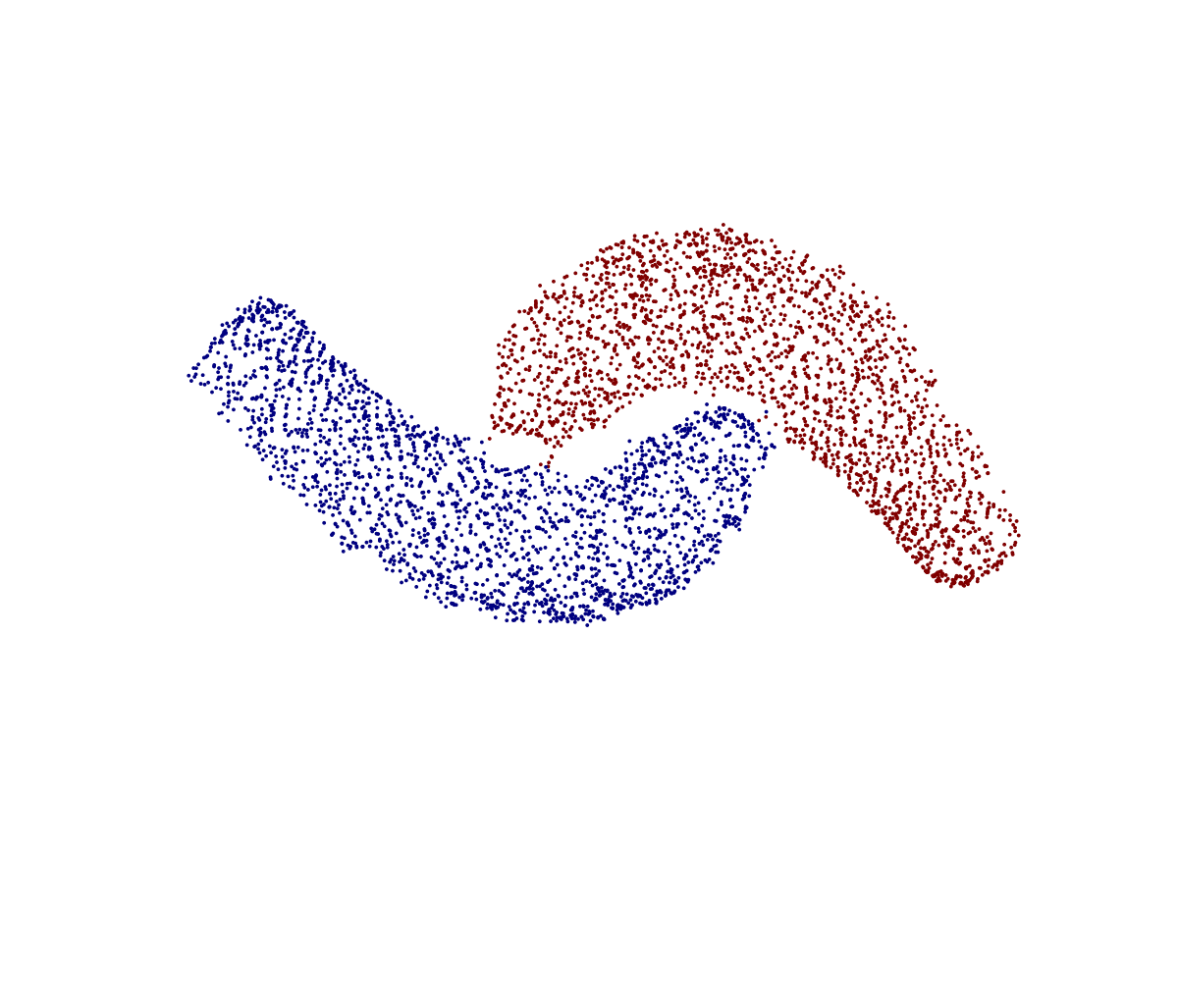}&
		\includegraphics[width=0.15\textwidth]{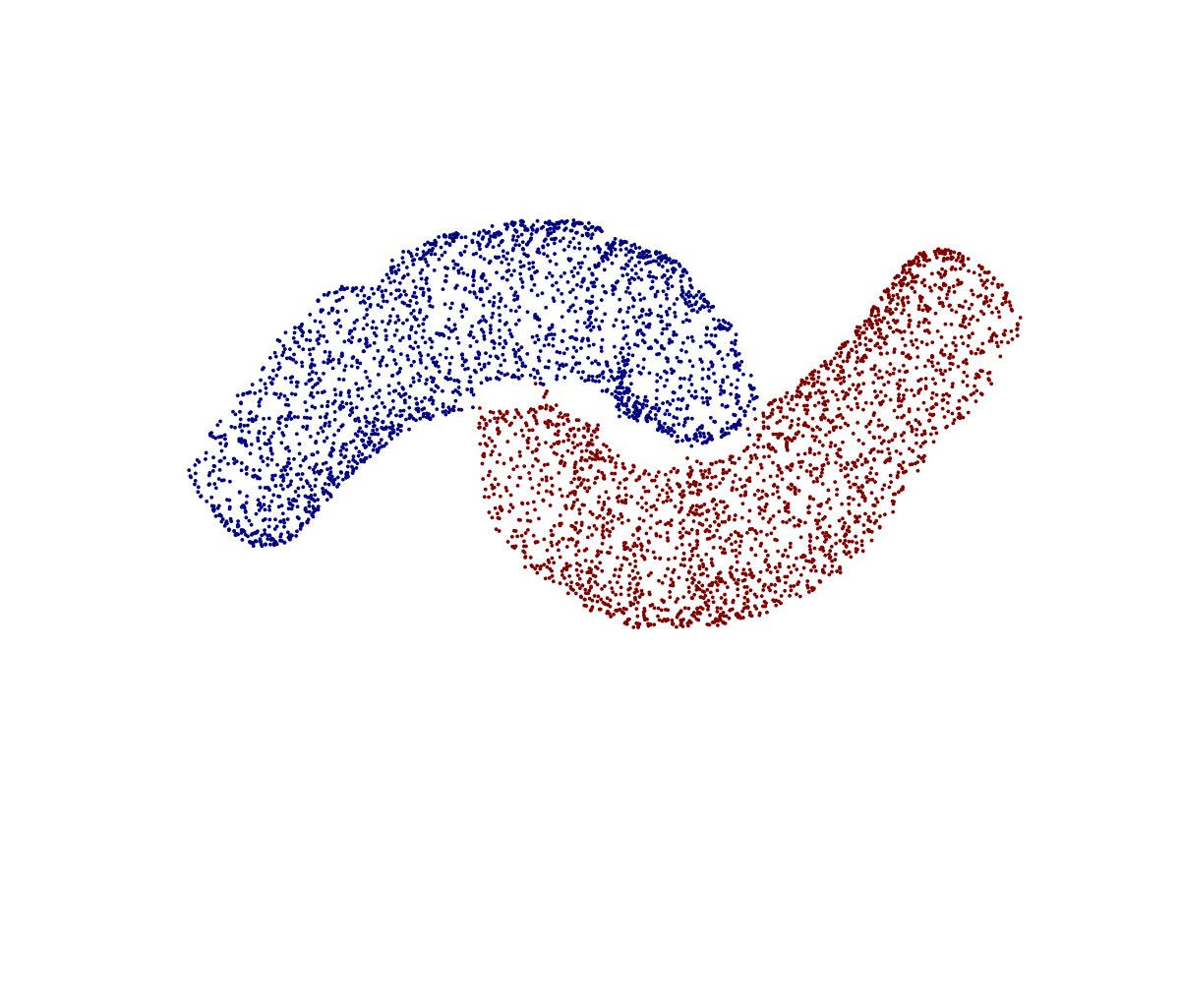} 
		&\includegraphics[width=0.15\textwidth]{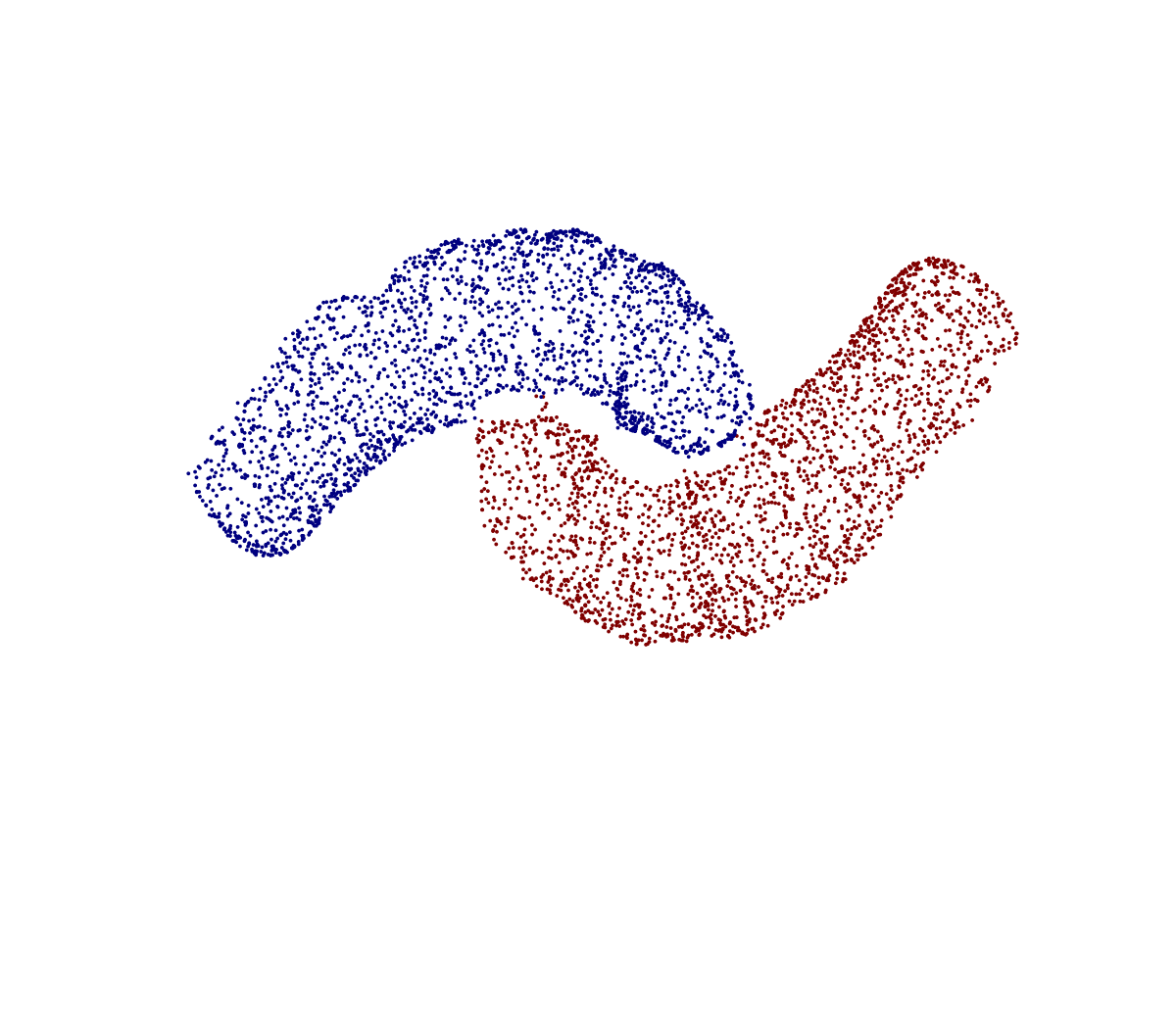} 
		\\
		\textbf{(2) $M_{\Pi}$} & \includegraphics[width=0.15\textwidth]{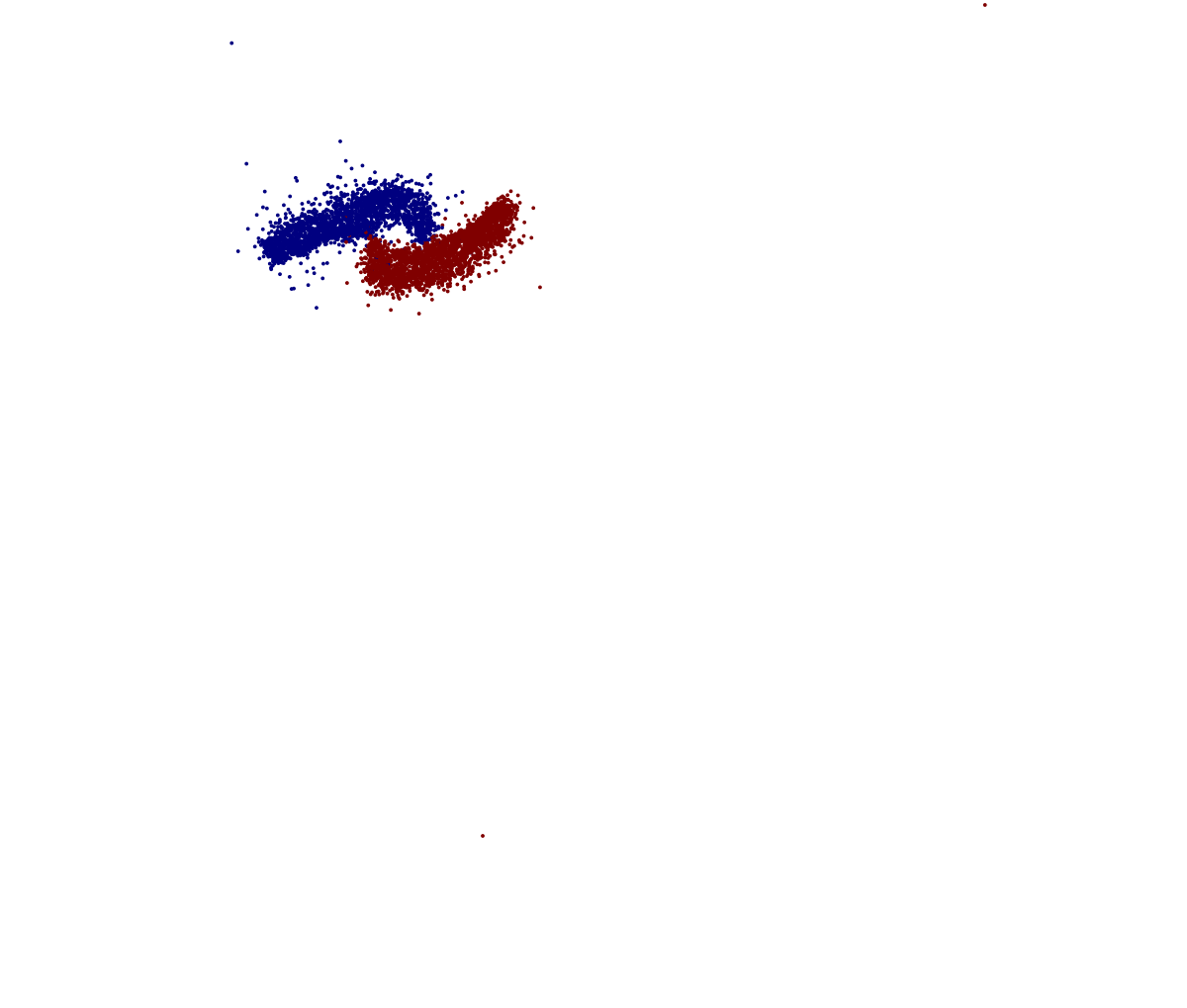} & \includegraphics[width=0.15\textwidth]{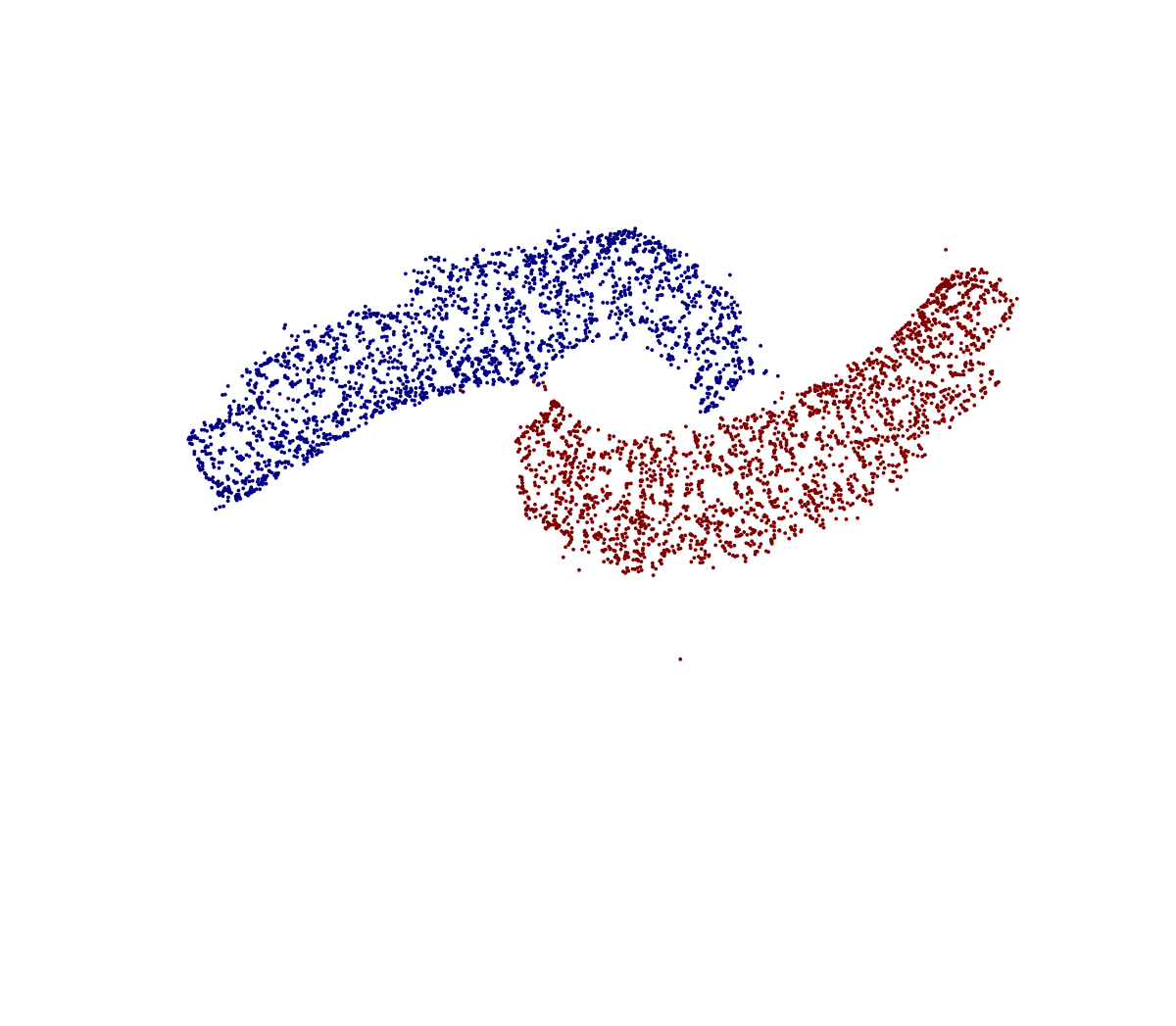} & \includegraphics[width=0.15\textwidth]{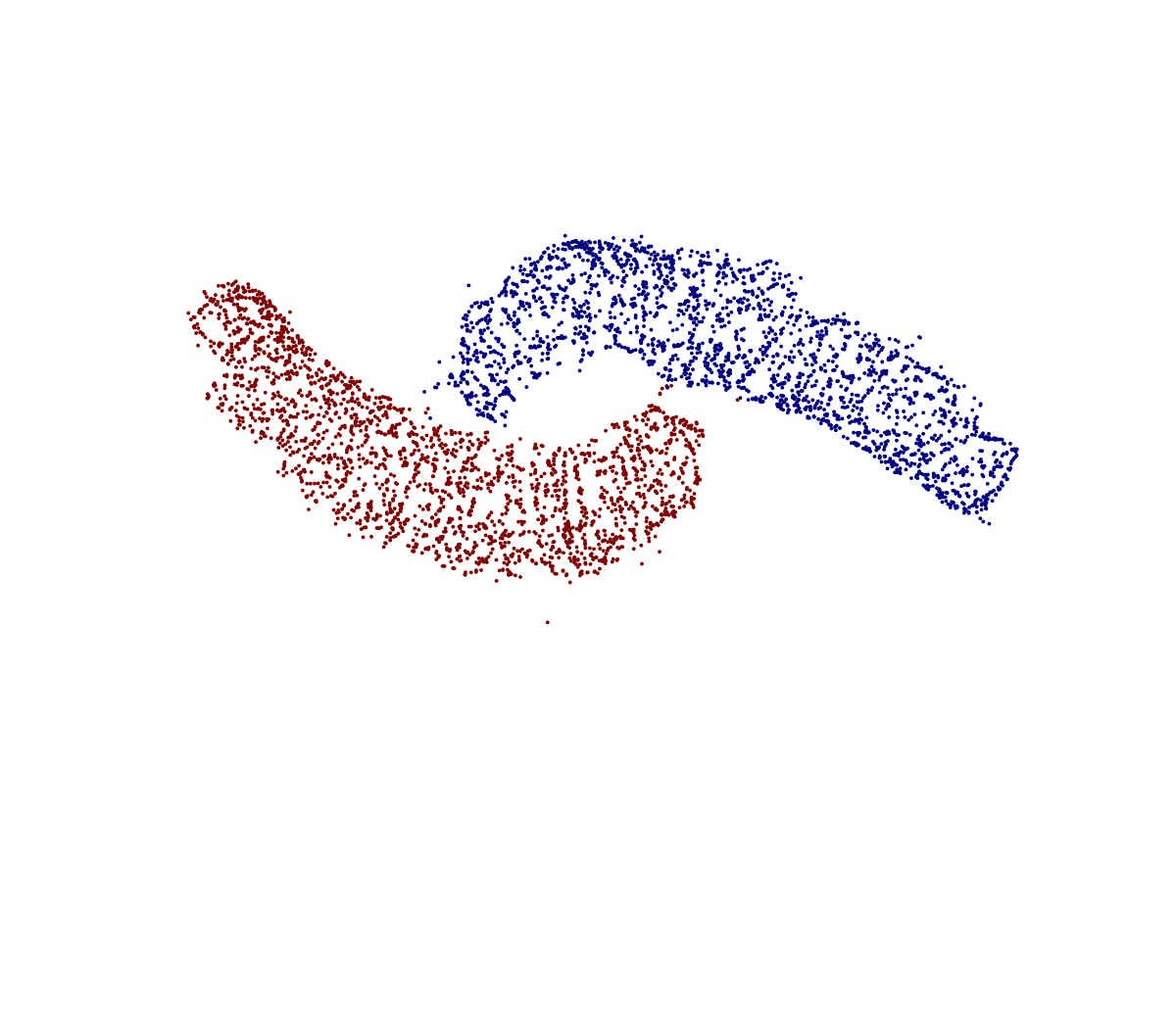} & \includegraphics[width=0.15\textwidth]{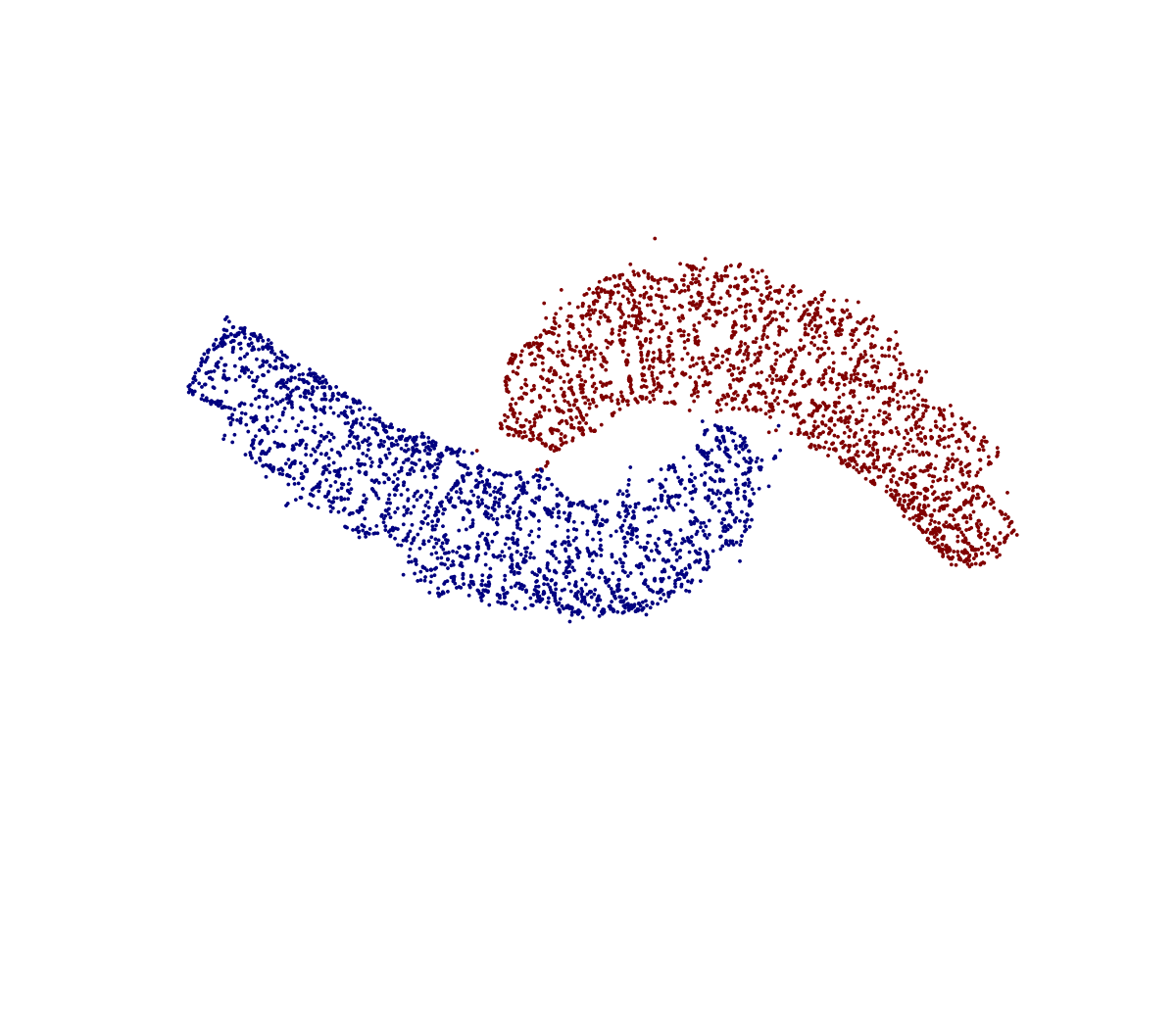} 
		& \includegraphics[width=0.15\textwidth]{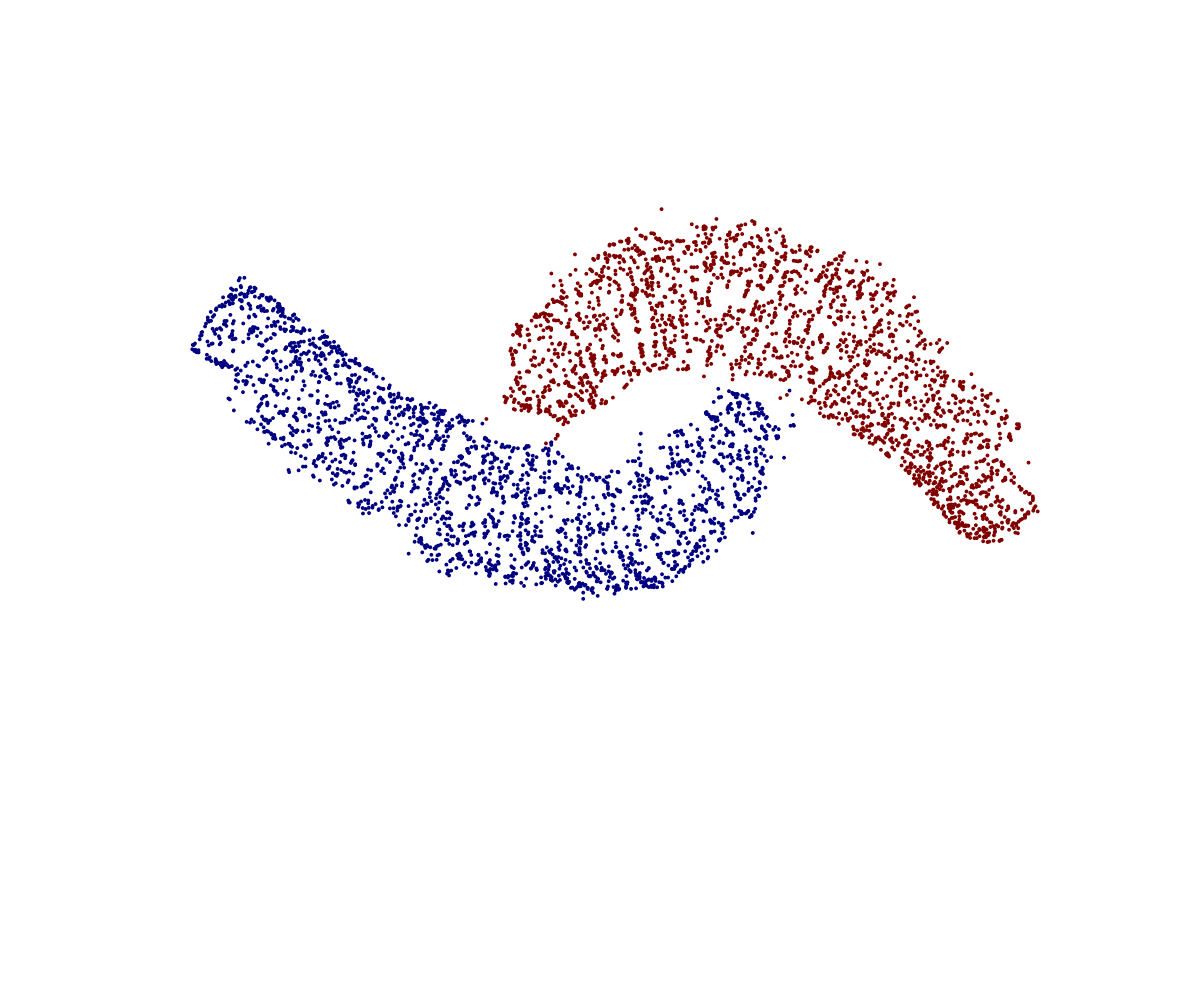} 
		\\
		\textbf{(3) $M_W$} & \includegraphics[width=0.15\textwidth]{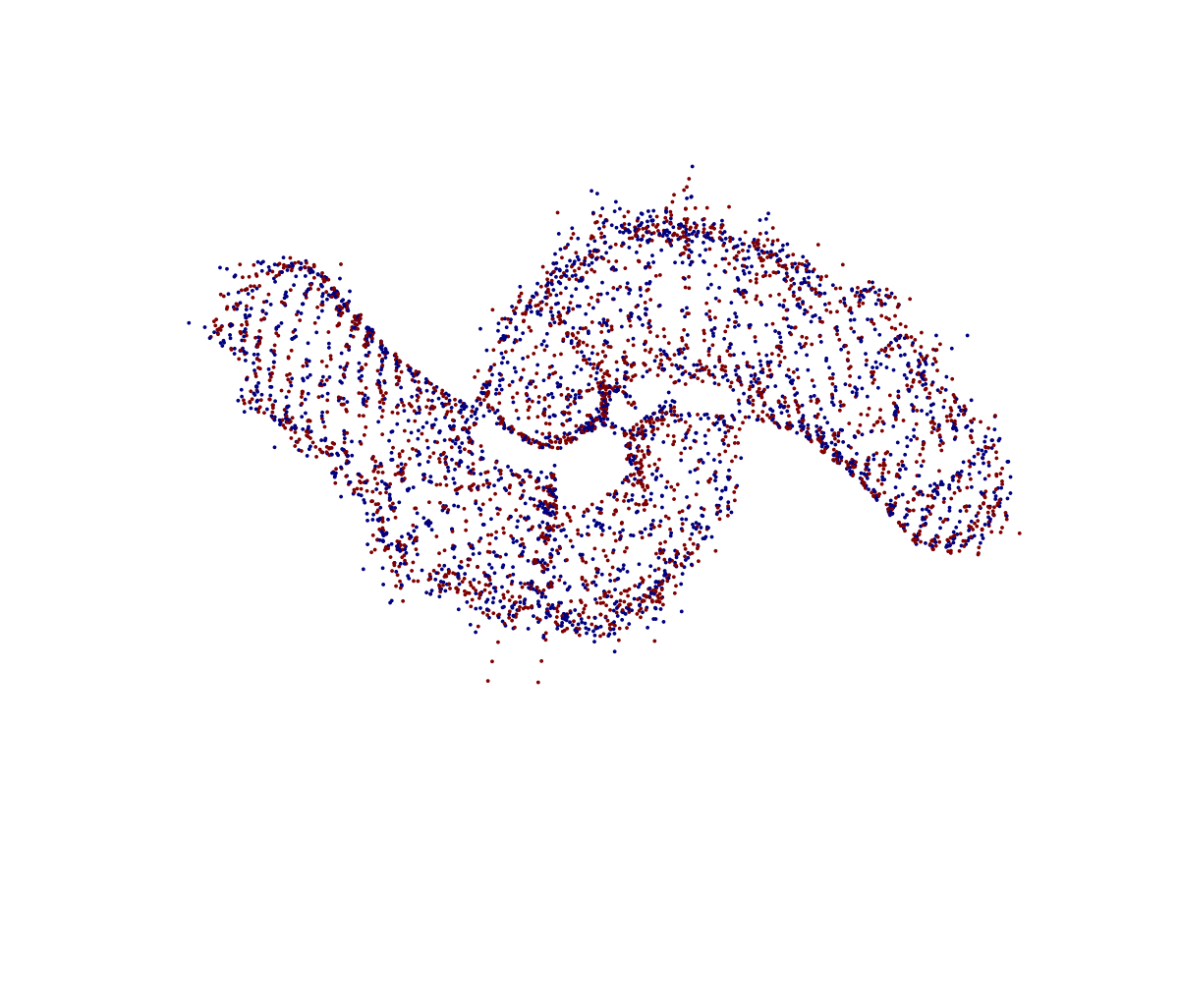} & \includegraphics[width=0.15\textwidth]{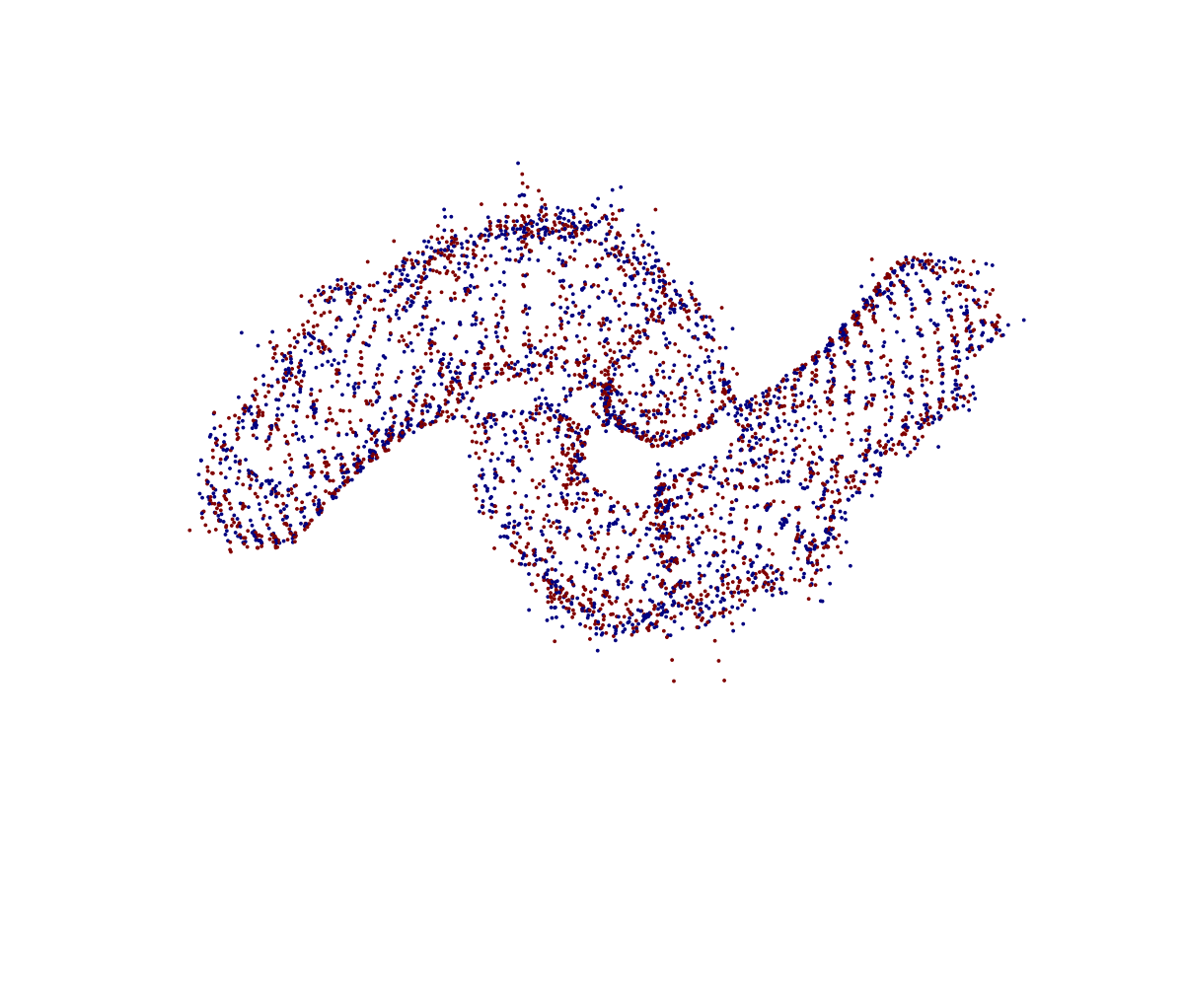} & \includegraphics[width=0.15\textwidth]{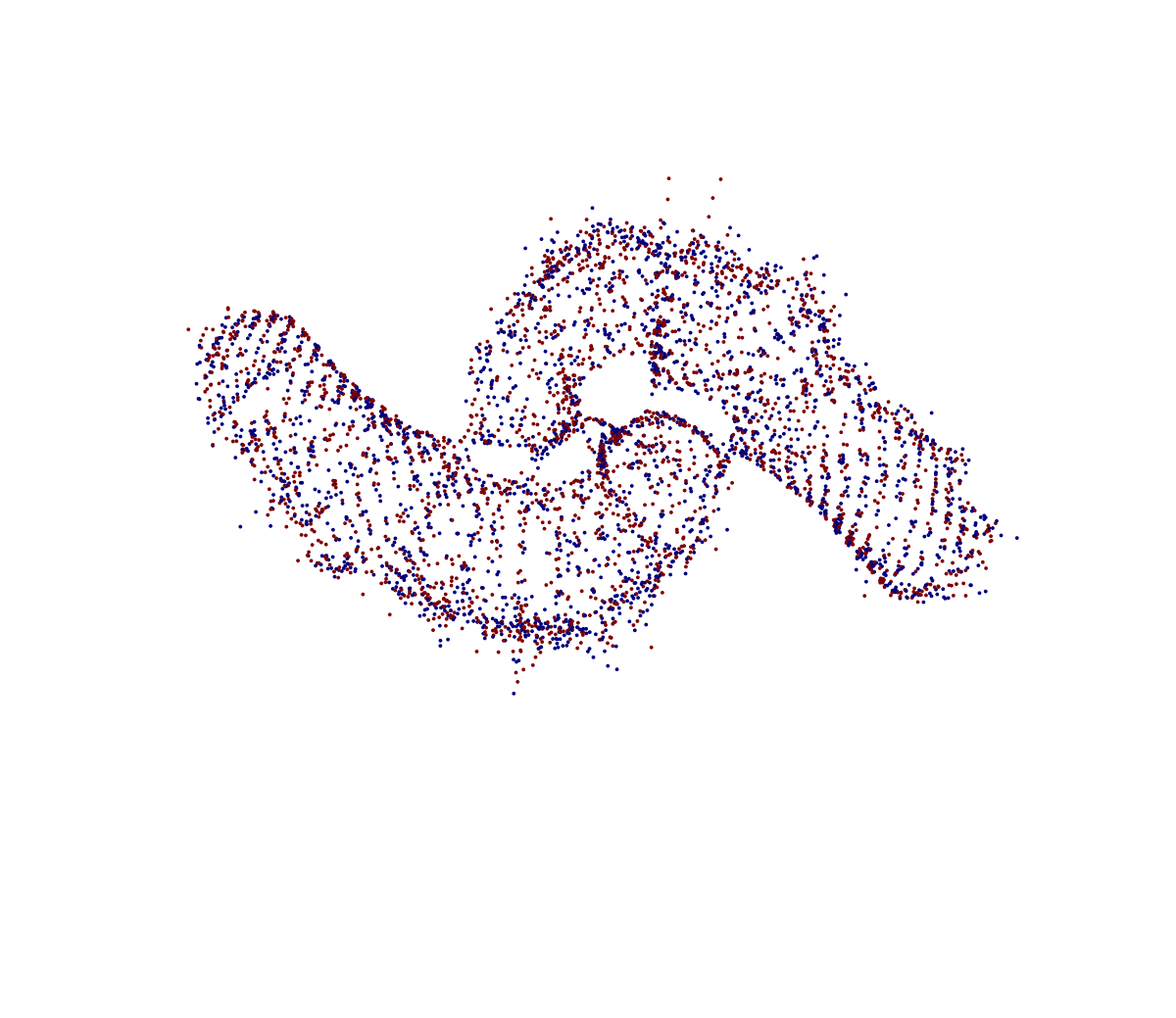} & \includegraphics[width=0.15\textwidth]{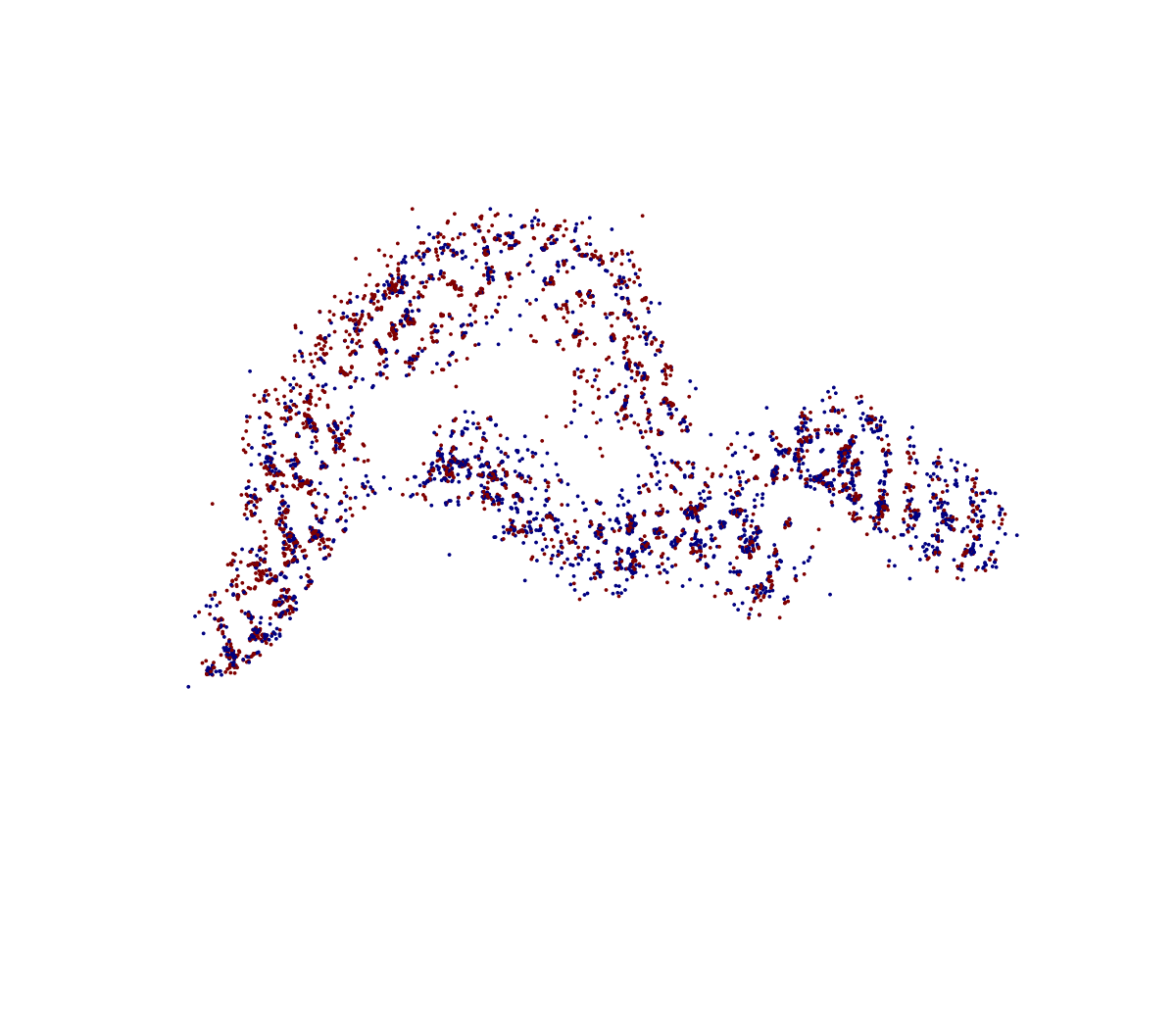} 
		& \includegraphics[width=0.15\textwidth]{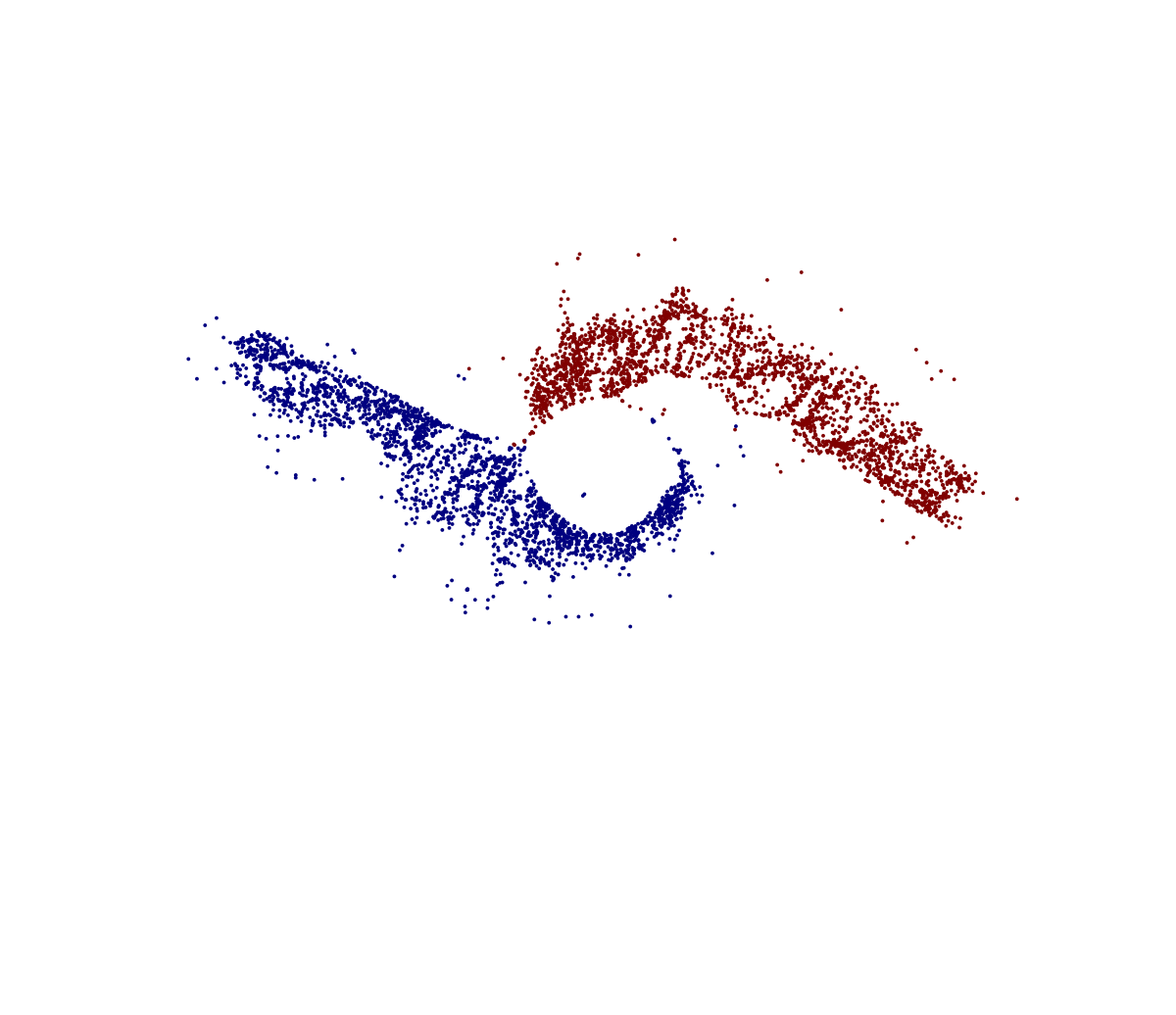}\\
		\textbf{(4) $H$} & & & & &\includegraphics[width=0.15\textwidth]{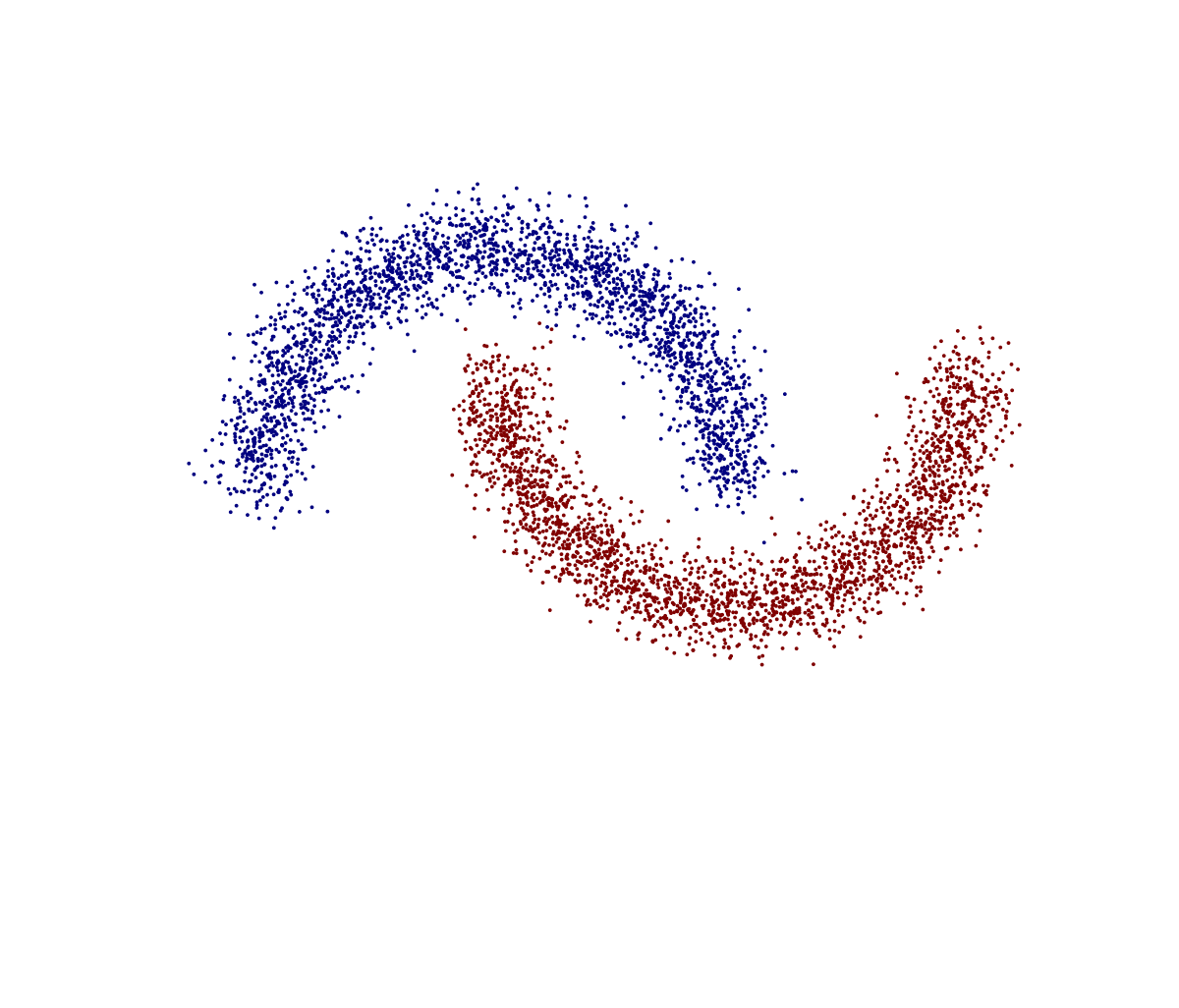}\\
	\end{tabular}
	\caption{Sample 5000 dataset from Two moons, mapped to $2$-d after local normalization of distances and applying $M_V$, $M_{\Pi}$, $M_W$, and $H$ (c.f. \ref{me4}, \ref{productlaw}, \ref{WSlaw}, and \ref{hyplaw} respectively ) with   various corresponding parameters and $k=15$ in neighborhood graph.}
	\label{tab:2moons}
\end{table}
\begin{table}[H]
	\centering
	\tiny
	\begin{tabular}{>{\centering\arraybackslash}m{1cm}|*{4}{>{\centering\arraybackslash}m{3cm}}}
		 & \textbf{$0.25$} & \textbf{$0.5$}  & \textbf{$0.75$} & \textbf{$1.0$} 
		\\
		& (a) & (b) & (c) & (d)  \\
		\hline \\
		\textbf{(1) $M_V$} & \includegraphics[width=0.17\textwidth]{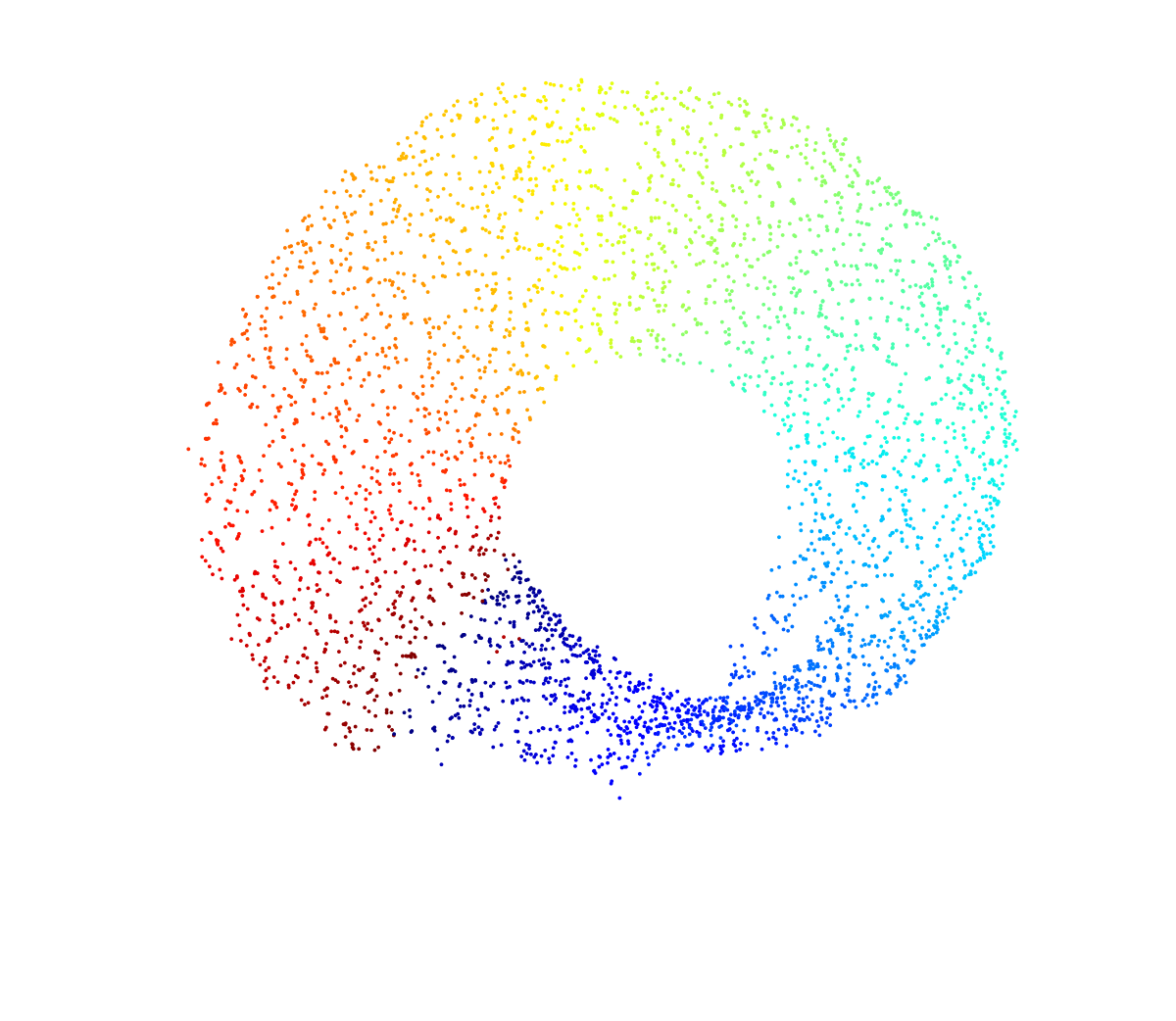} &  \includegraphics[width=0.17\textwidth]{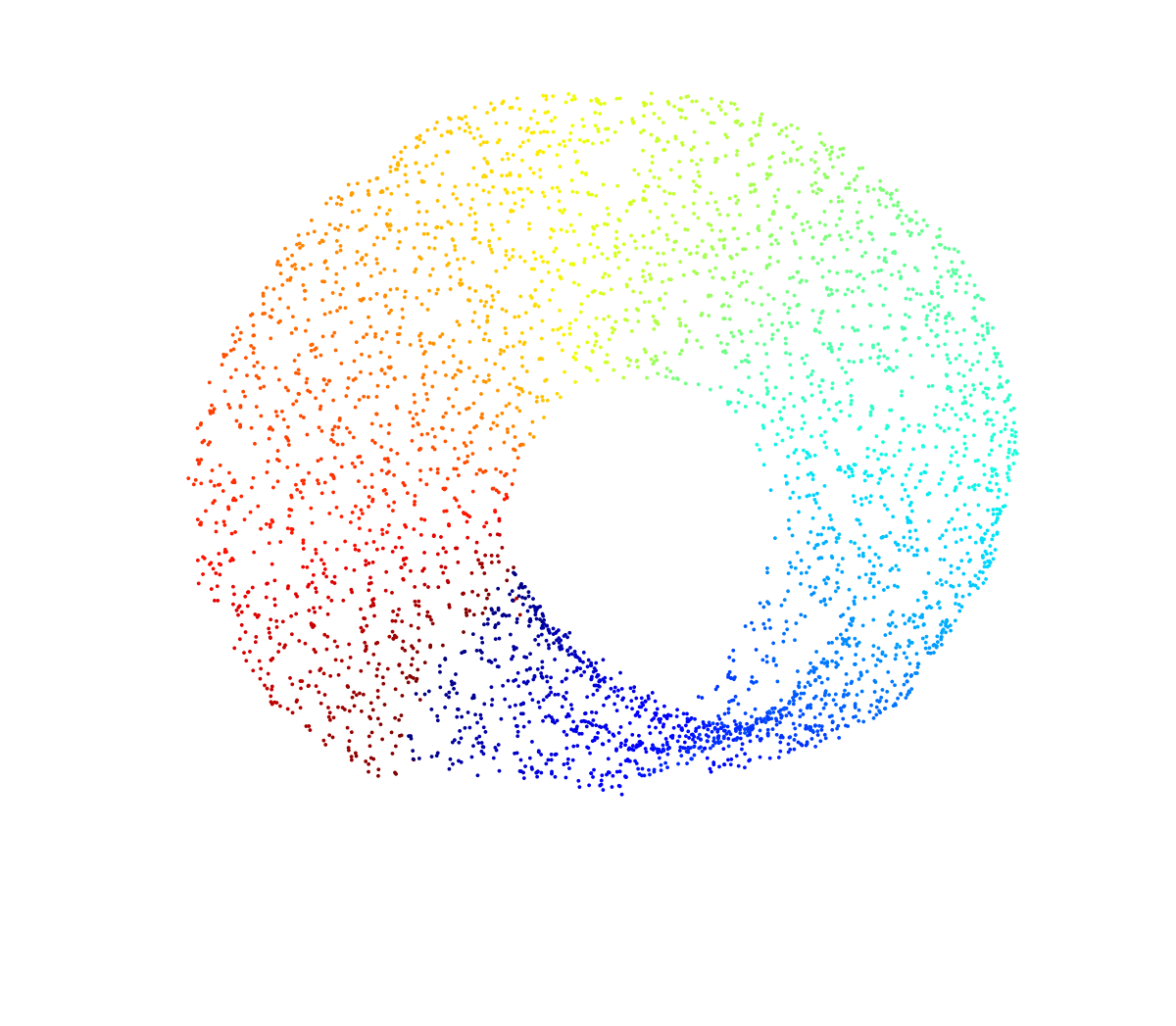} & \includegraphics[width=0.17\textwidth]{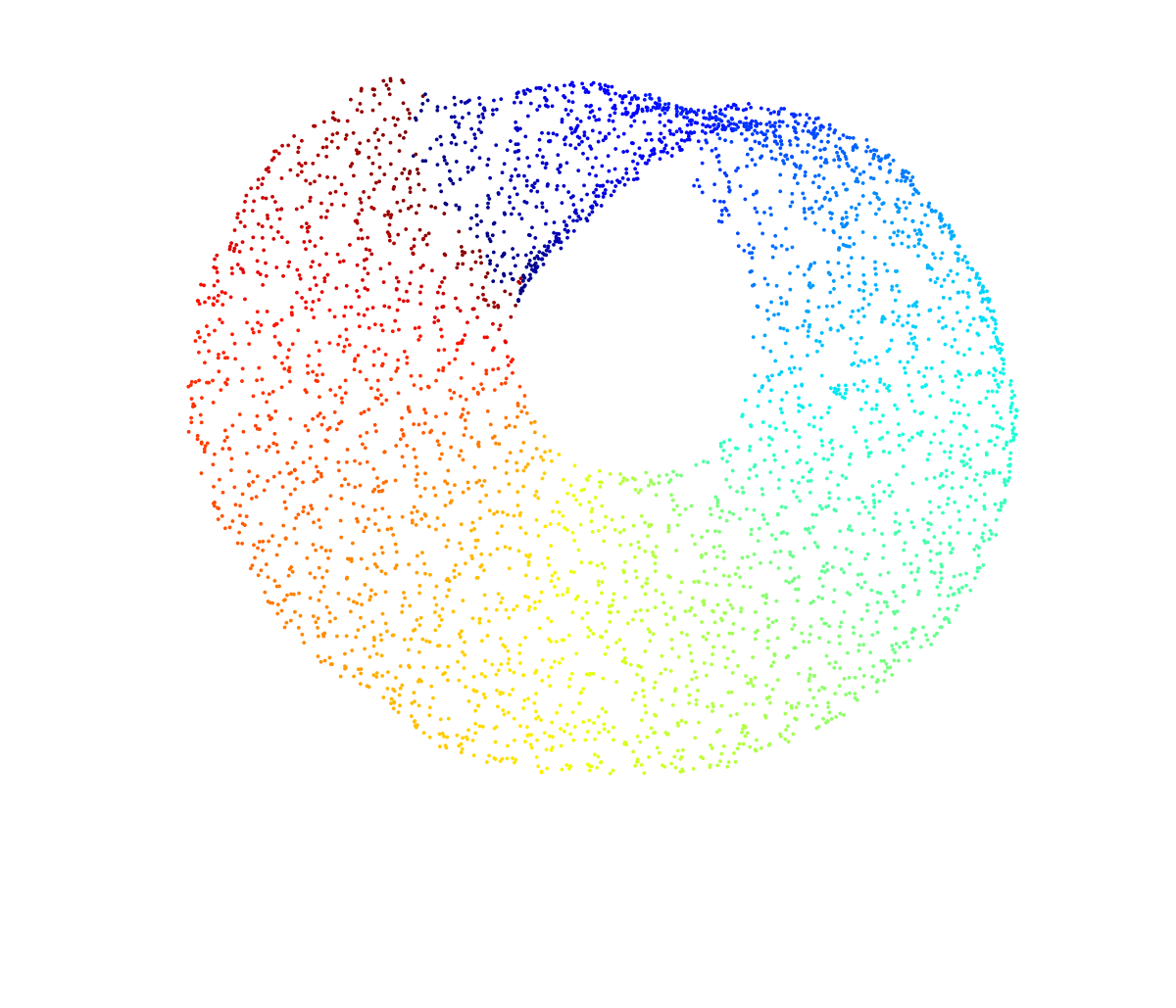}
				&\includegraphics[width=0.17\textwidth]{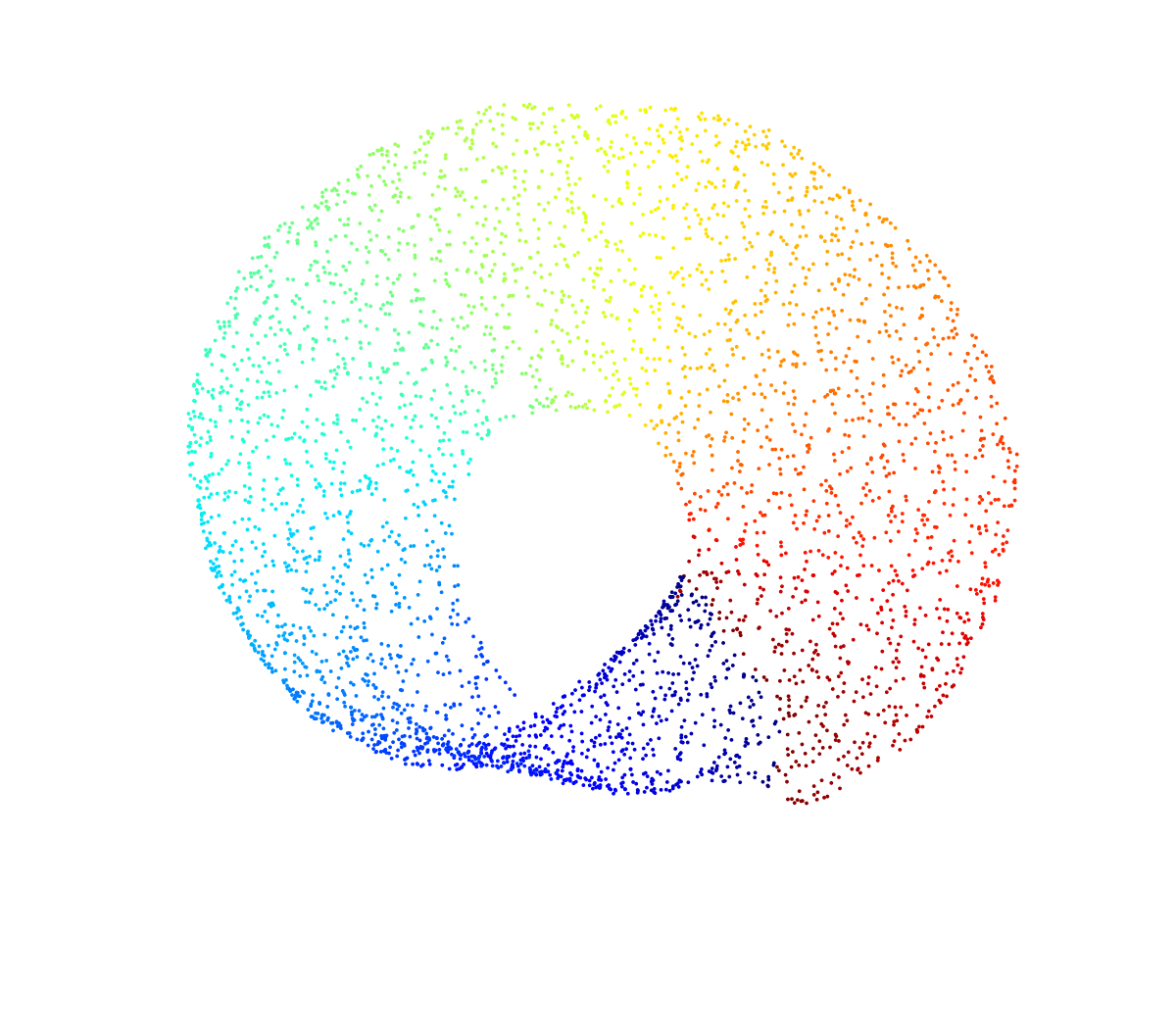} 
		\\
		\textbf{(2) $M_{\Pi}$} & \includegraphics[width=0.17\textwidth]{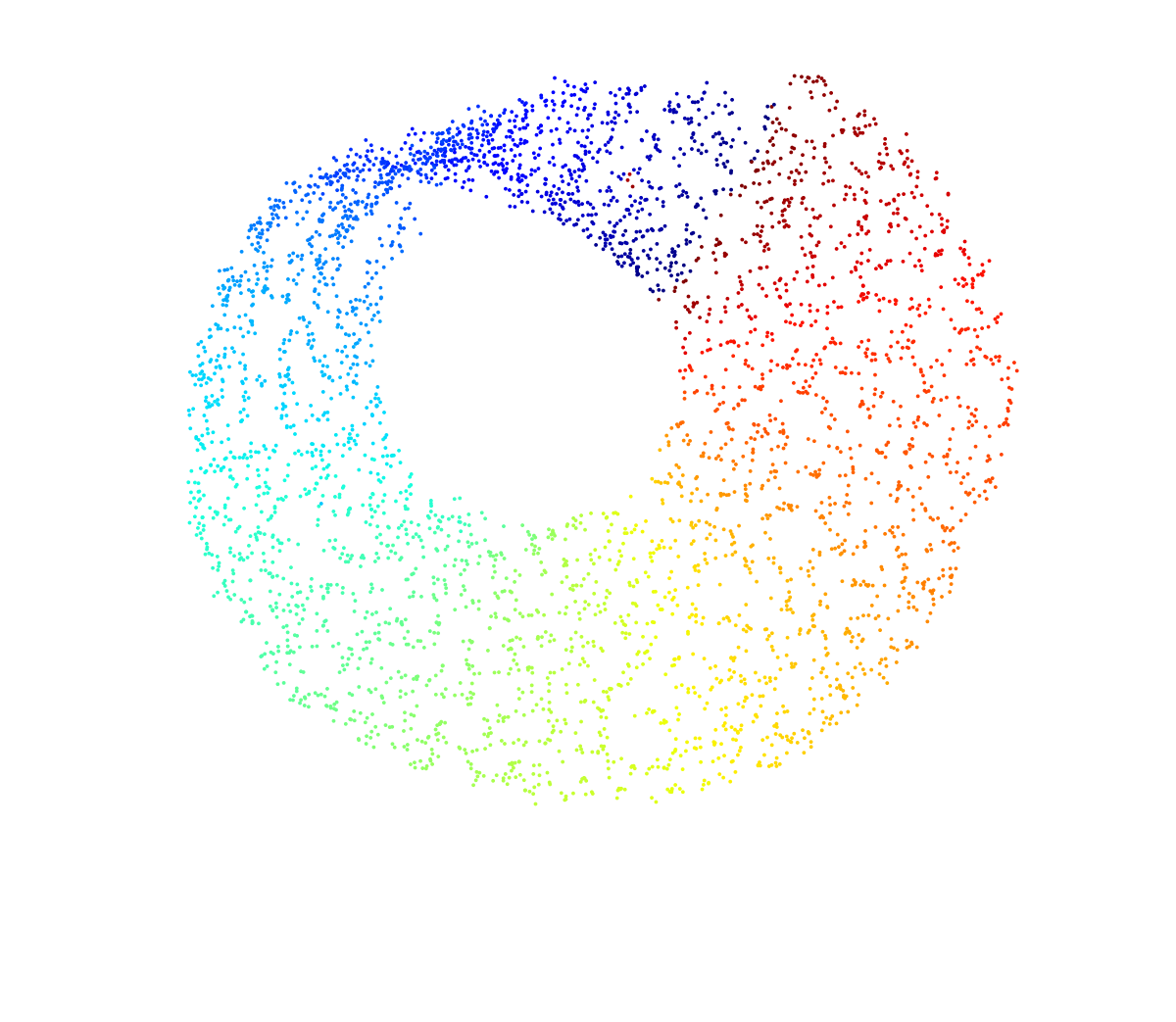} & \includegraphics[width=0.17\textwidth]{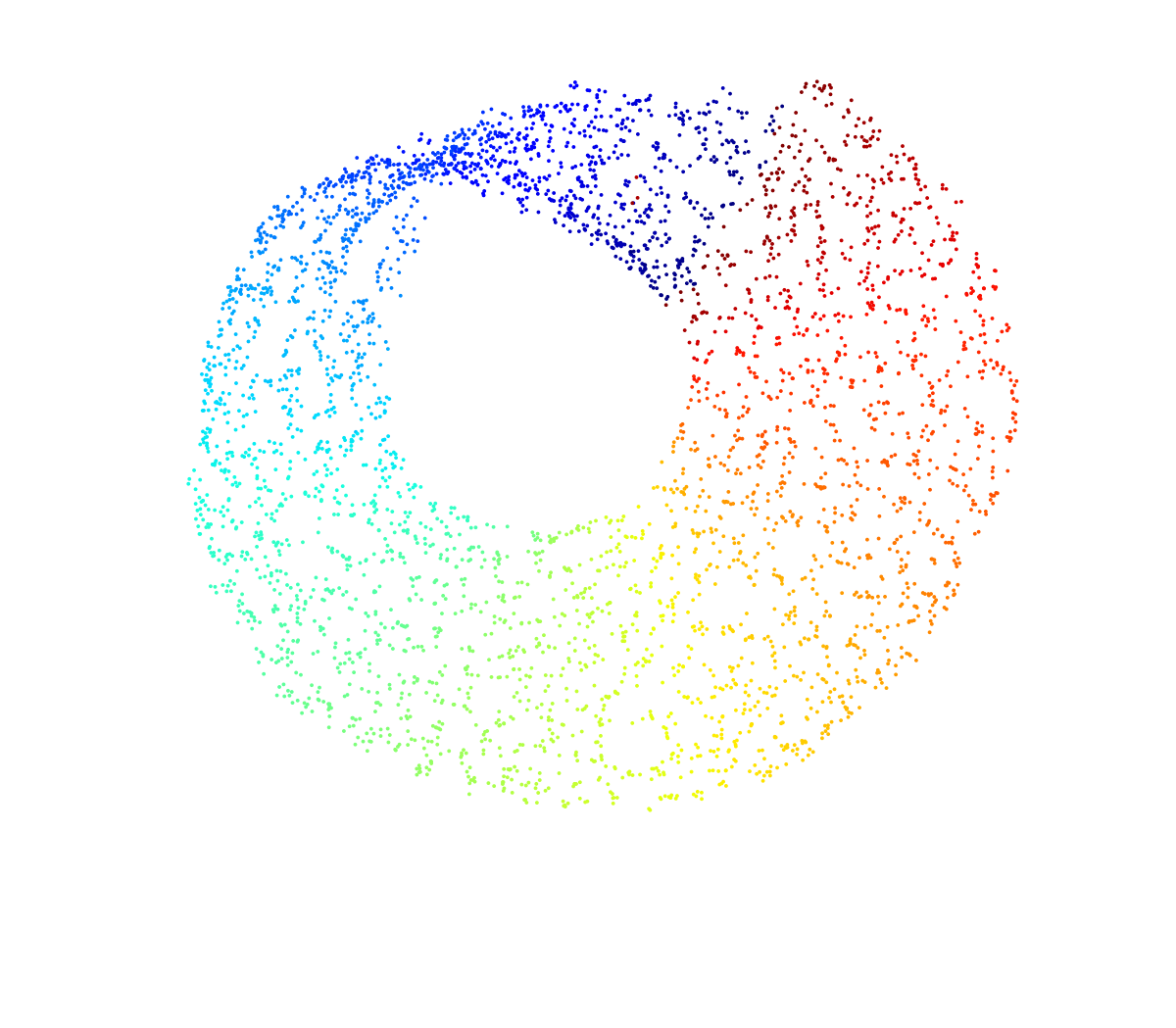} & \includegraphics[width=0.17\textwidth]{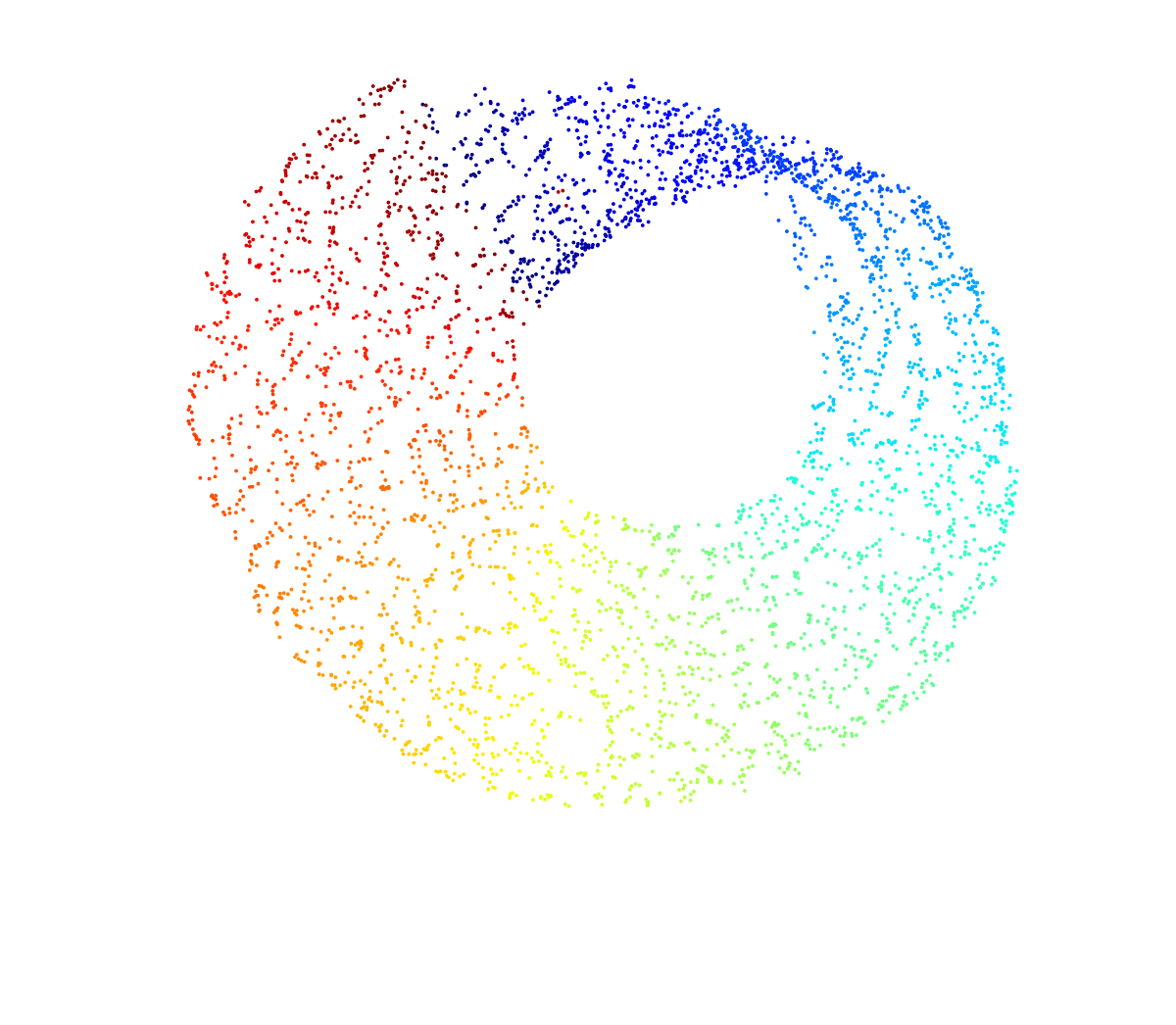} & \includegraphics[width=0.17\textwidth]{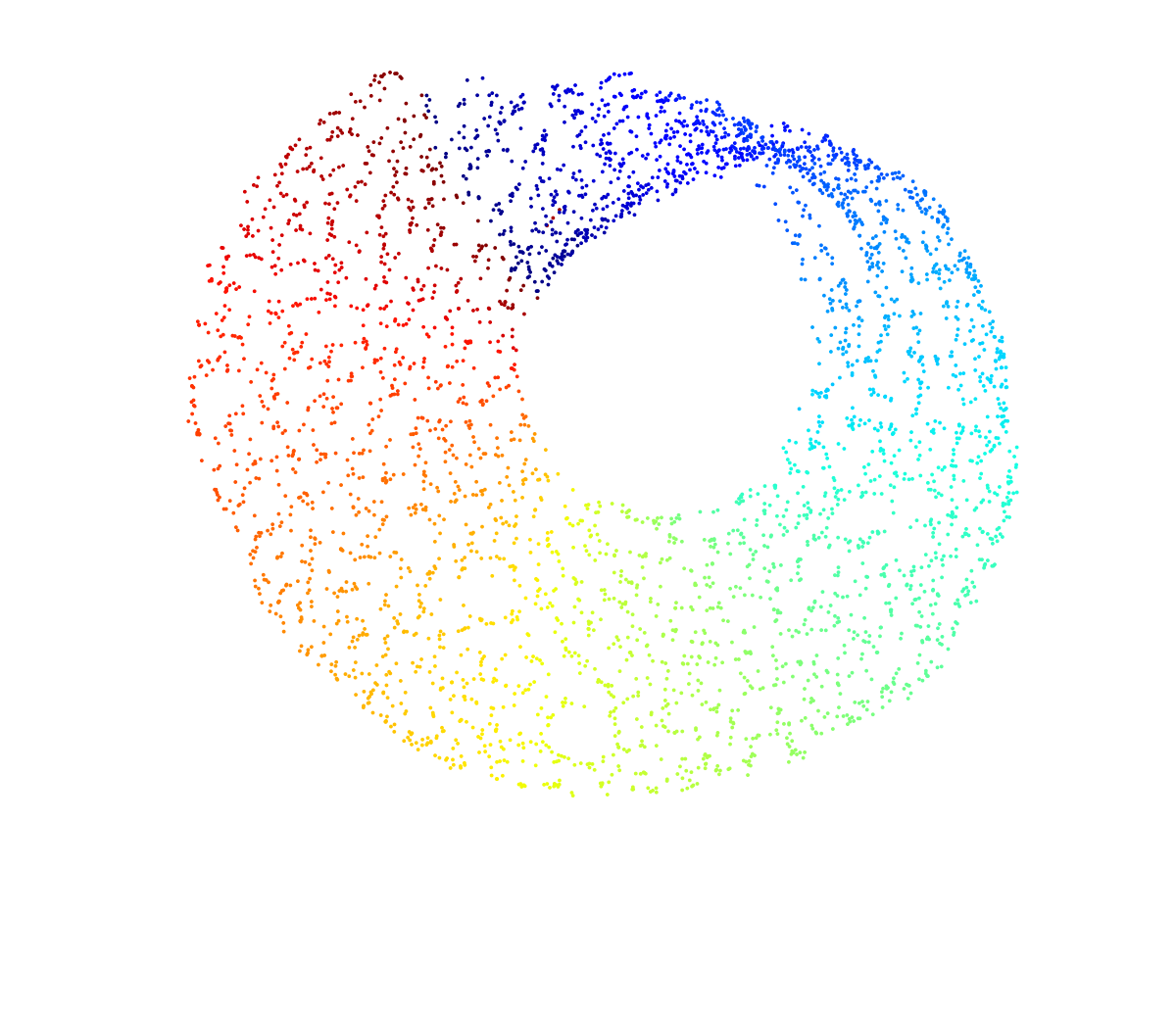} 
				\\
		\textbf{(3) $M_W$} & \includegraphics[width=0.17\textwidth]{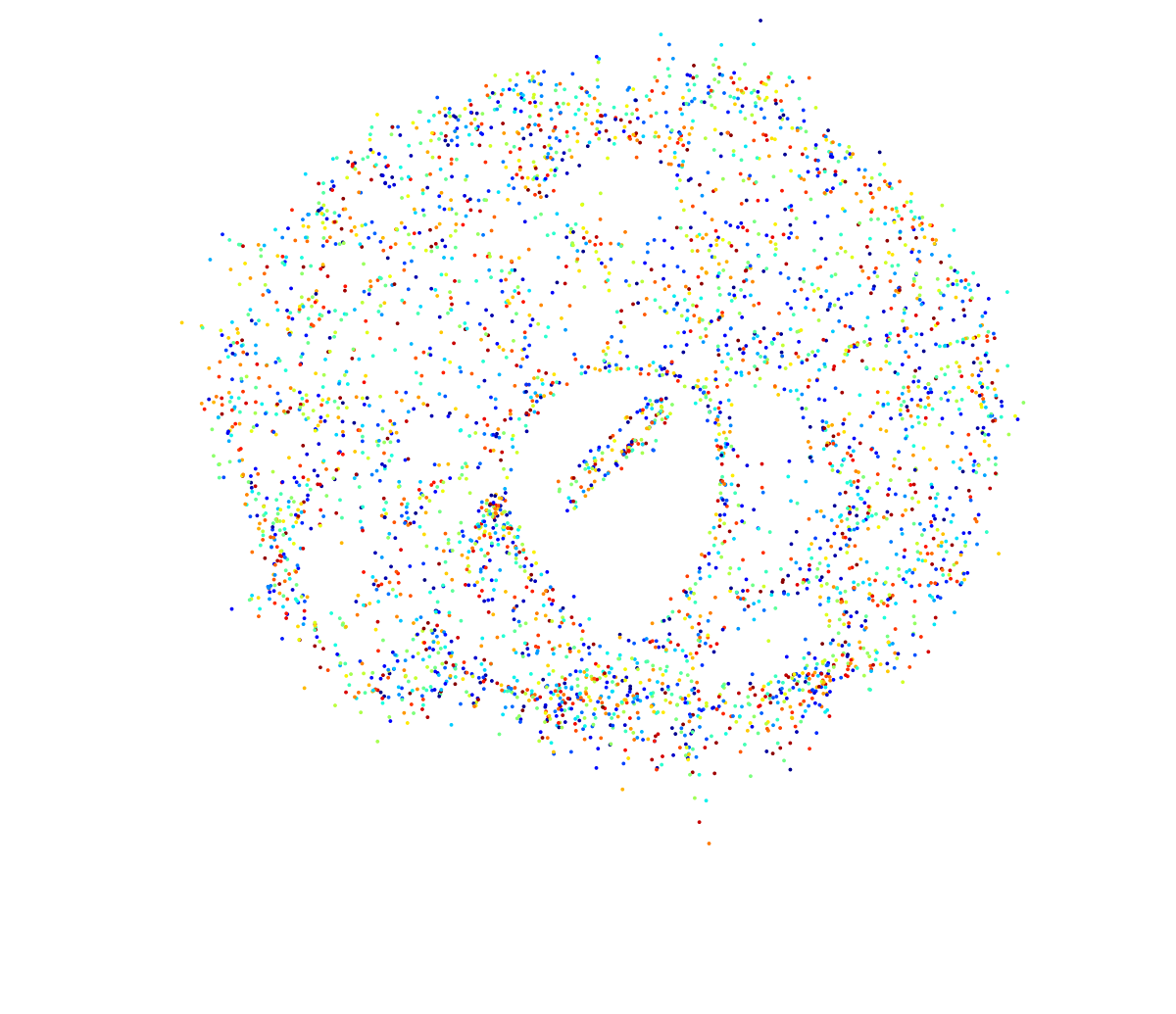} & \includegraphics[width=0.17\textwidth]{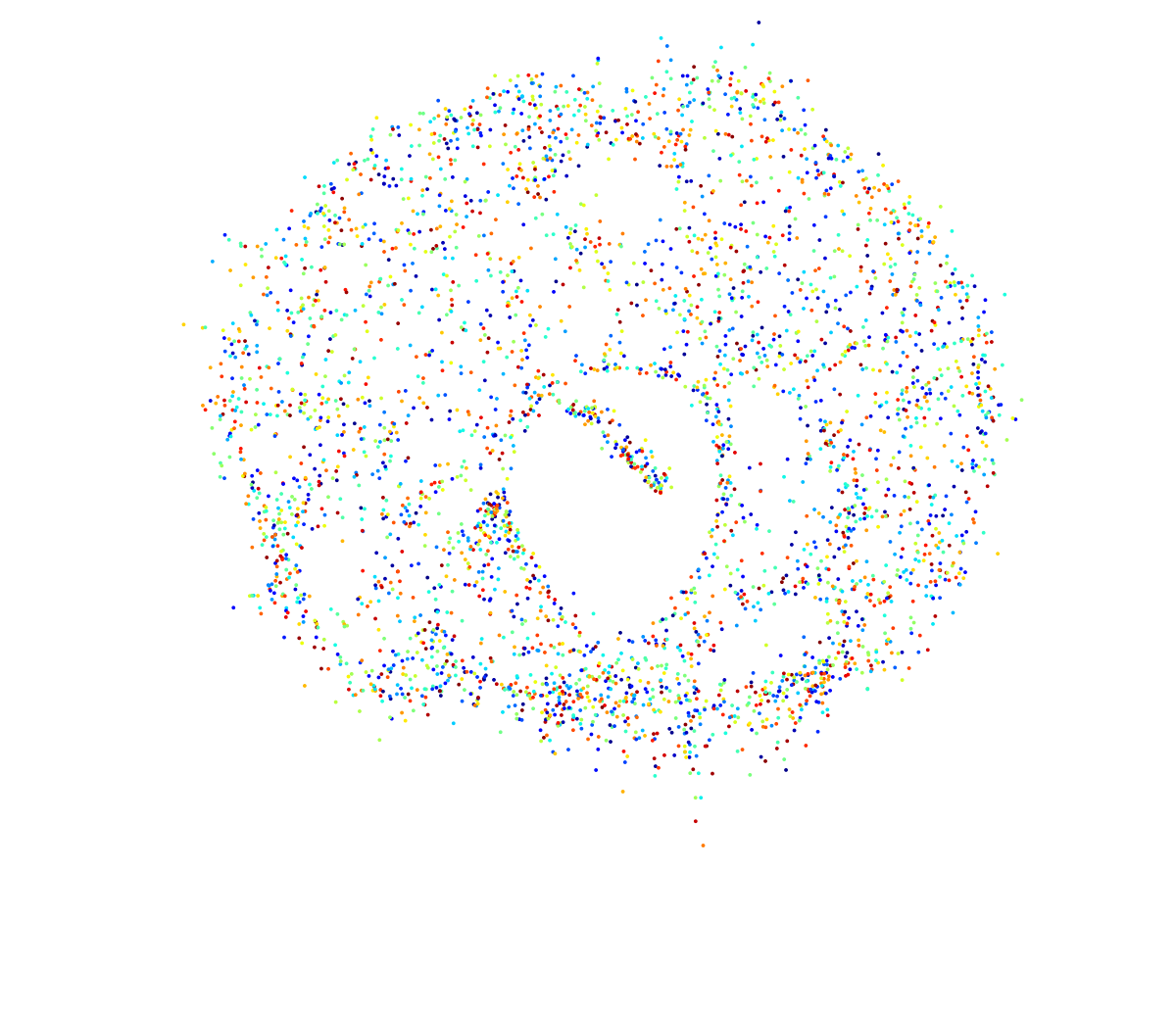} & \includegraphics[width=0.17\textwidth]{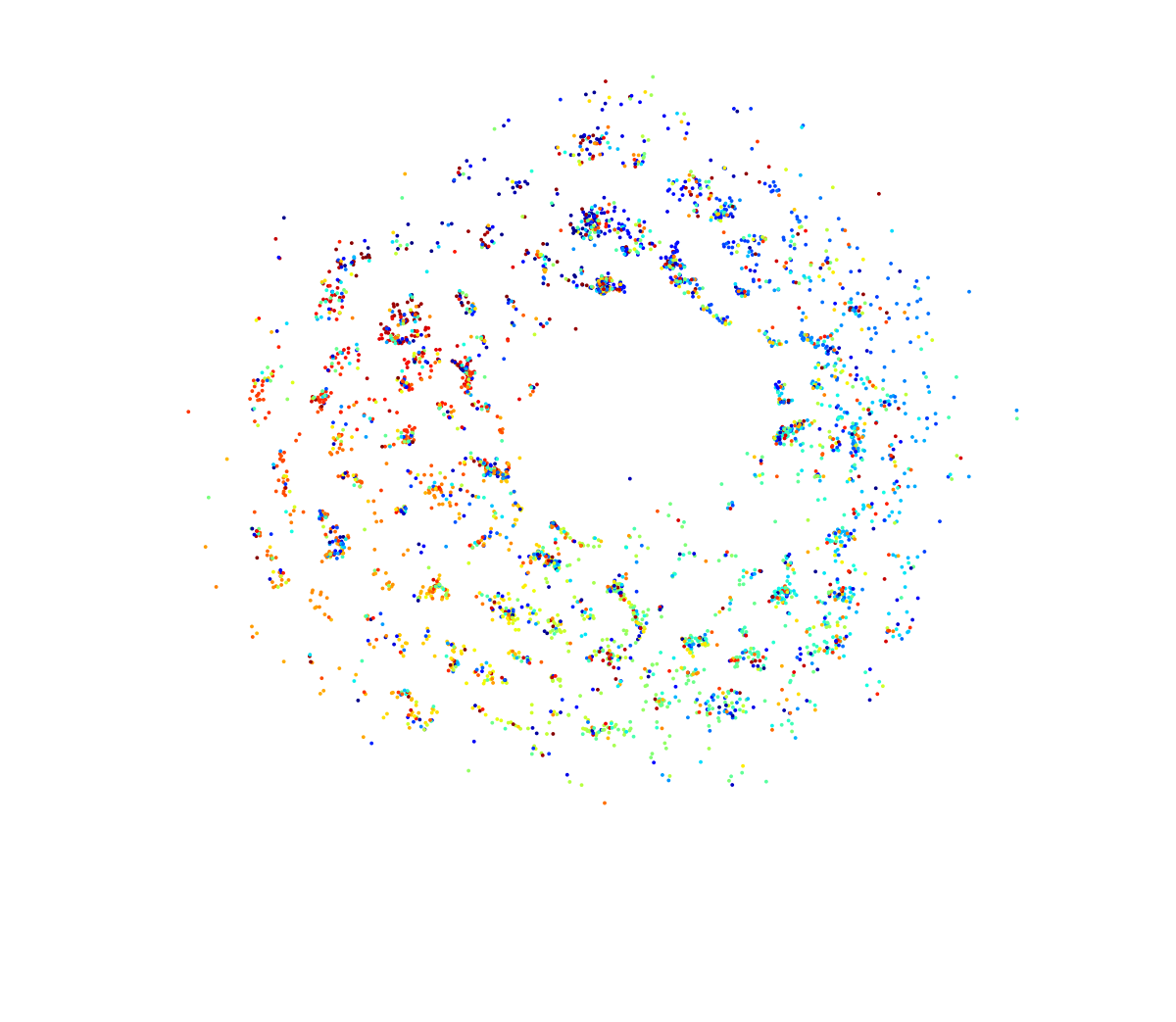} & \includegraphics[width=0.17\textwidth]{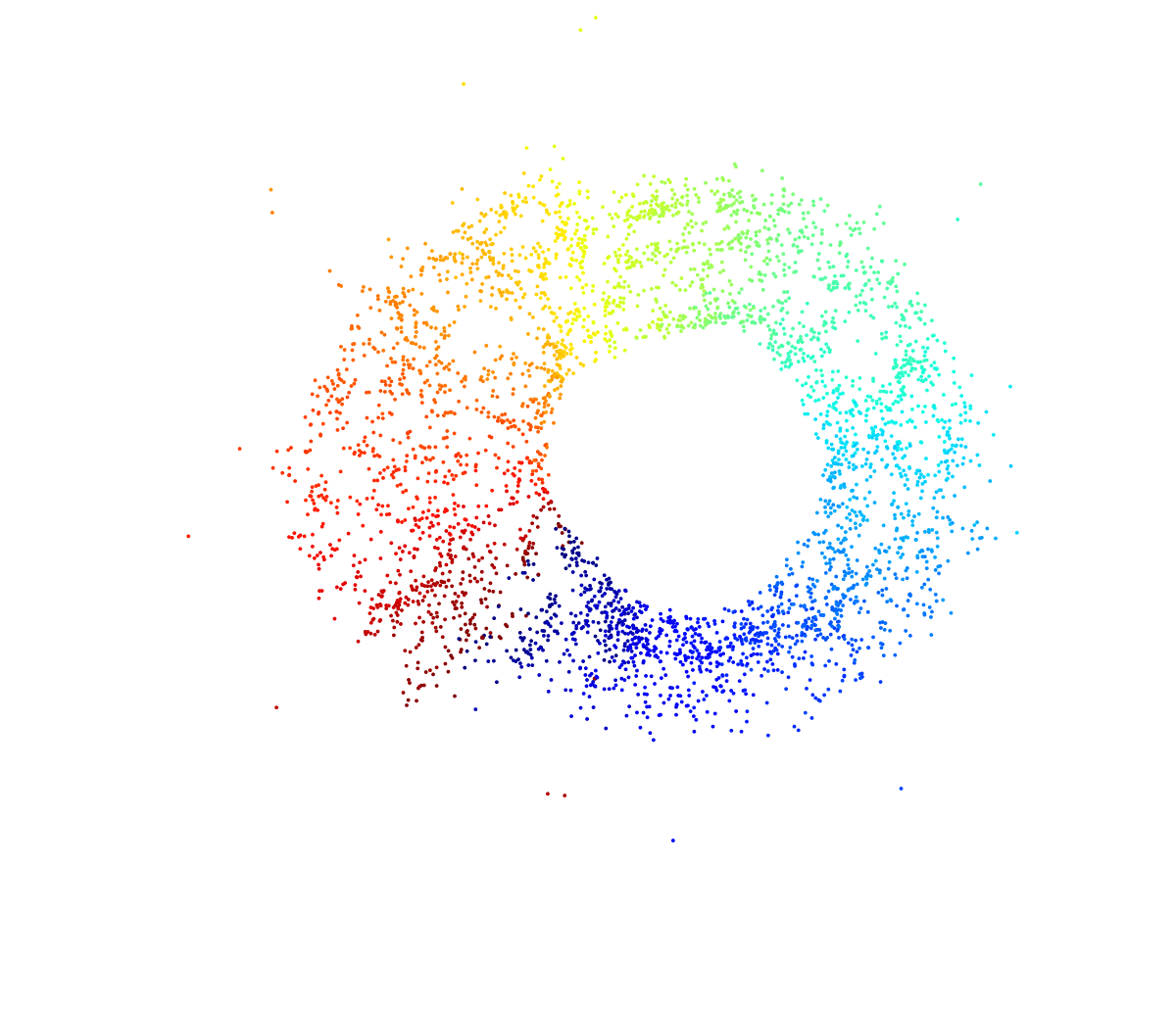}\\ 
		\textbf{(4) $H$} & & & &\includegraphics[width=0.17\textwidth]{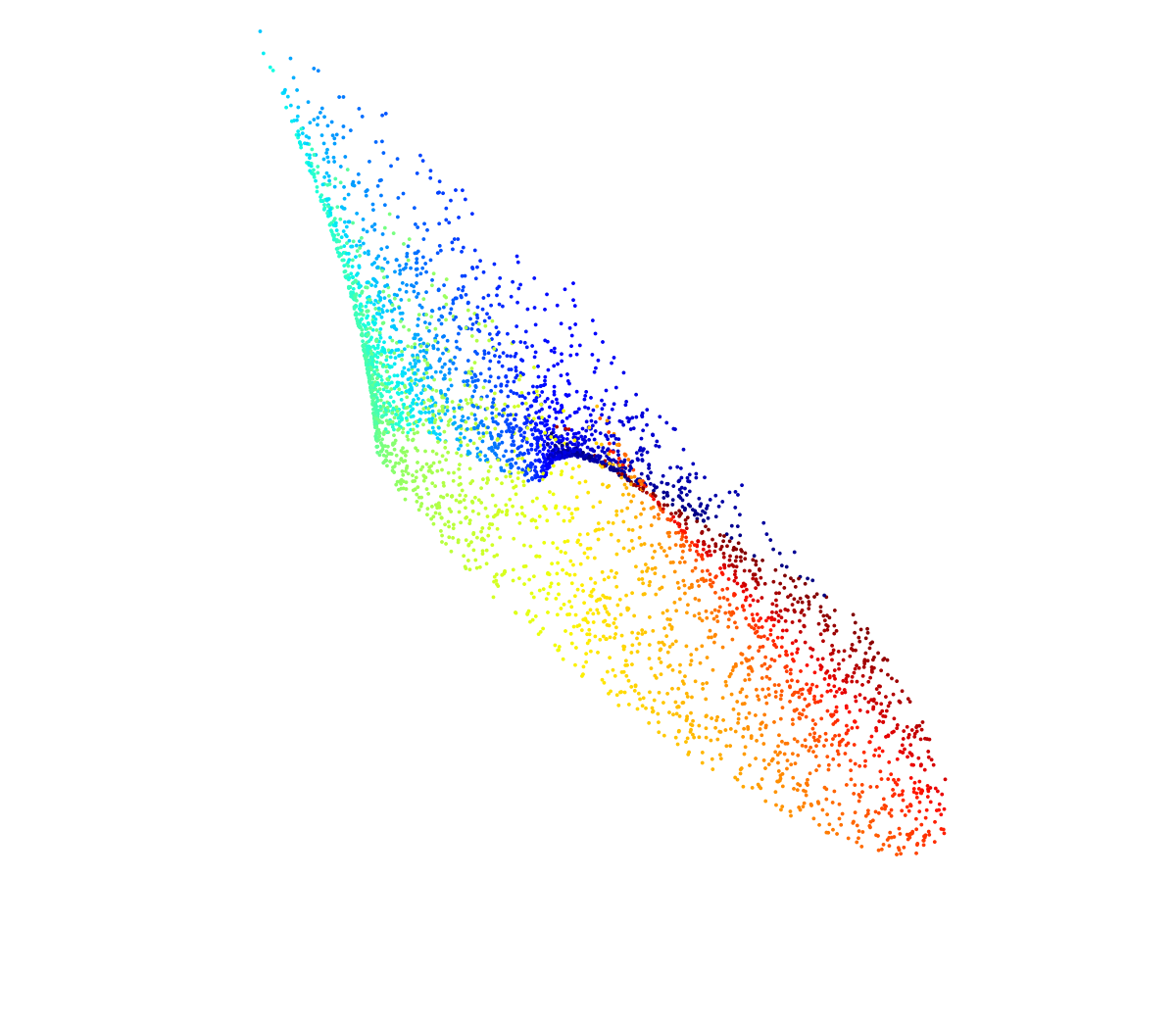}\\
			\end{tabular}
	\caption{Sample 4000 dataset from Mobius strip , mapped to $2$-d after local normalization of distances and applying $M_V$, $M_{\Pi}$, $M_W$, and $H$ (c.f. \ref{me4}, \ref{productlaw}, \ref{WSlaw}, and \ref{hyplaw} respectively ) with   various corresponding parameters and $k=15$ in neighborhood graph.}
	\label{tab:Mobius}
\end{table}
\begin{table}[H]
	\centering
	\tiny
	\begin{tabular}{>{\centering\arraybackslash}m{1cm}|*{3}{>{\centering\arraybackslash}m{4cm}}}
		& \textbf{$2$} & \textbf{$5$} & \textbf{$10$}  
		\\
		& (a) & (b) & (c)  \\
		\hline \\
		\textbf{$M_{\Pi}$ on SwissRoll} & \includegraphics[width=0.25\textwidth]{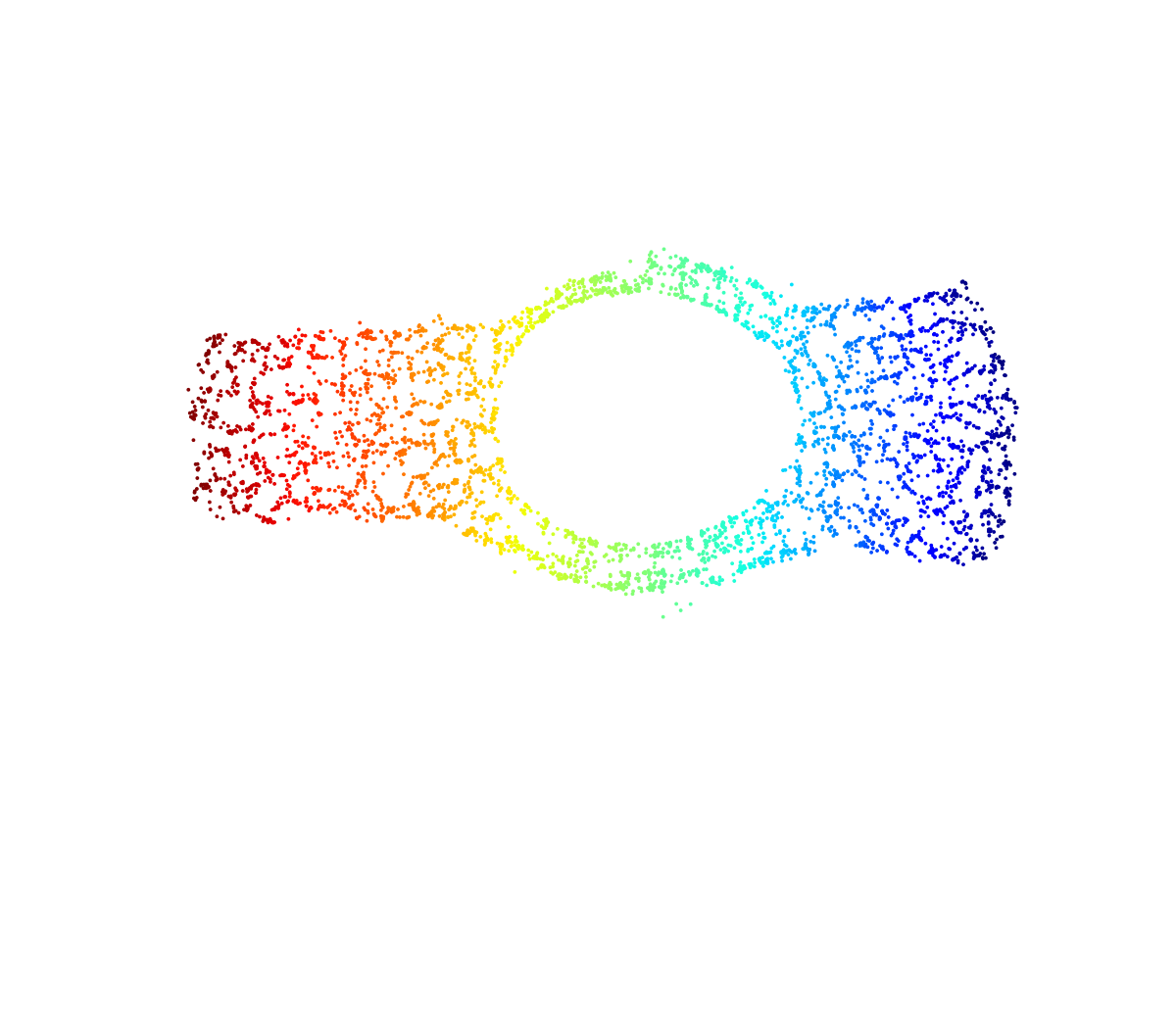} &  \includegraphics[width=0.25\textwidth]{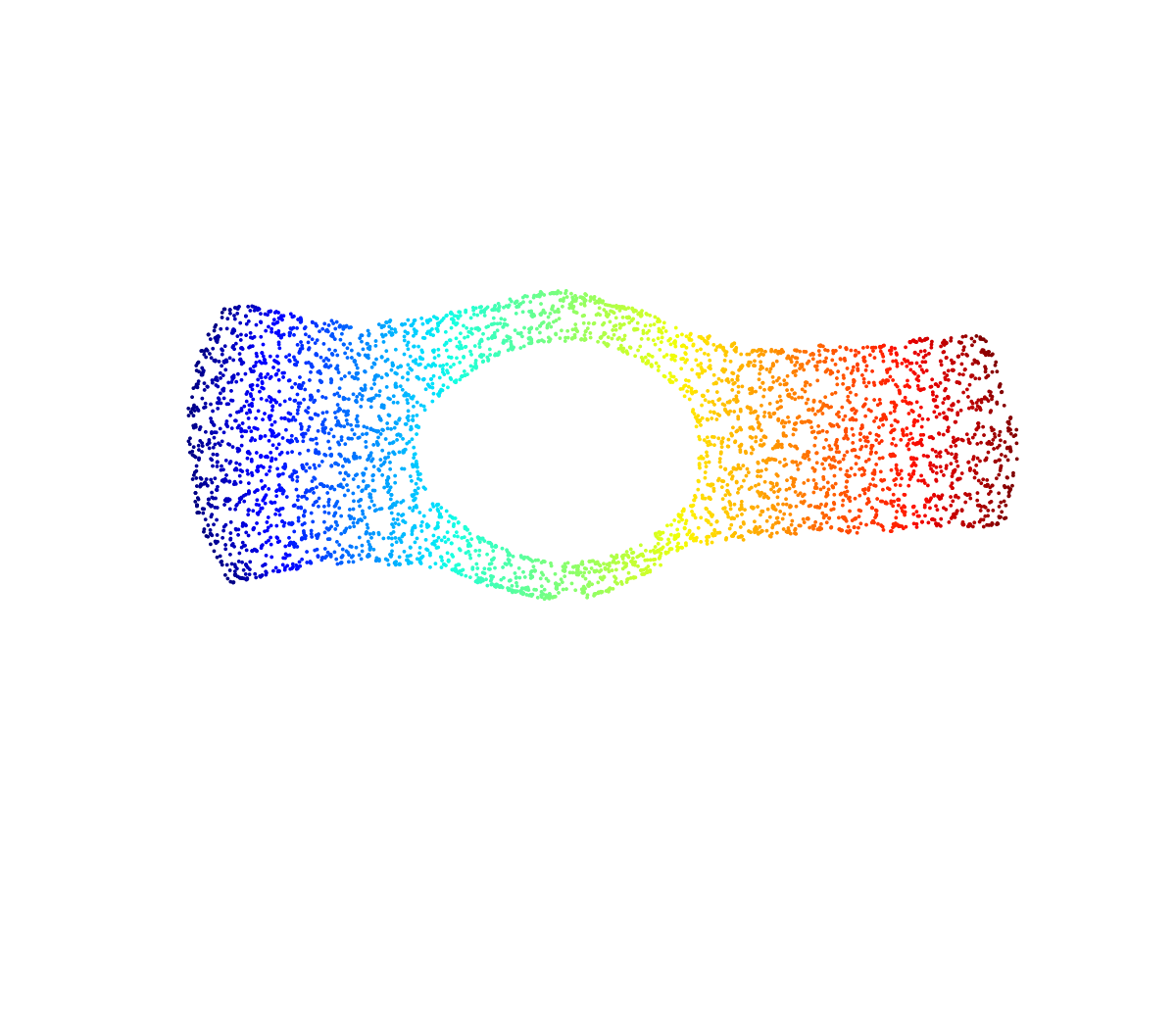} & \includegraphics[width=0.25\textwidth]{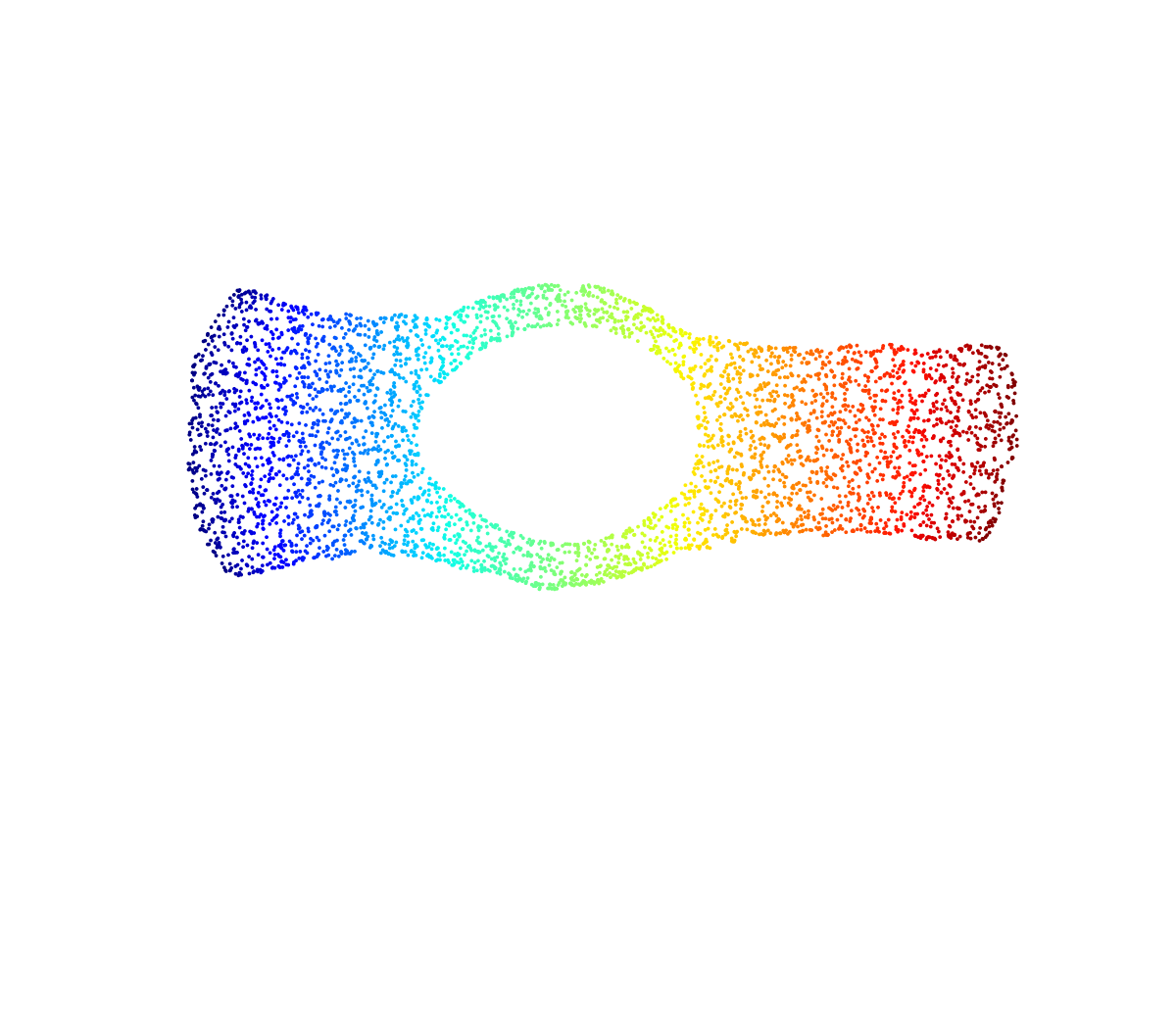}		\\
		\textbf{$M_{W}$ on SwissRoll} & \includegraphics[width=0.25\textwidth]{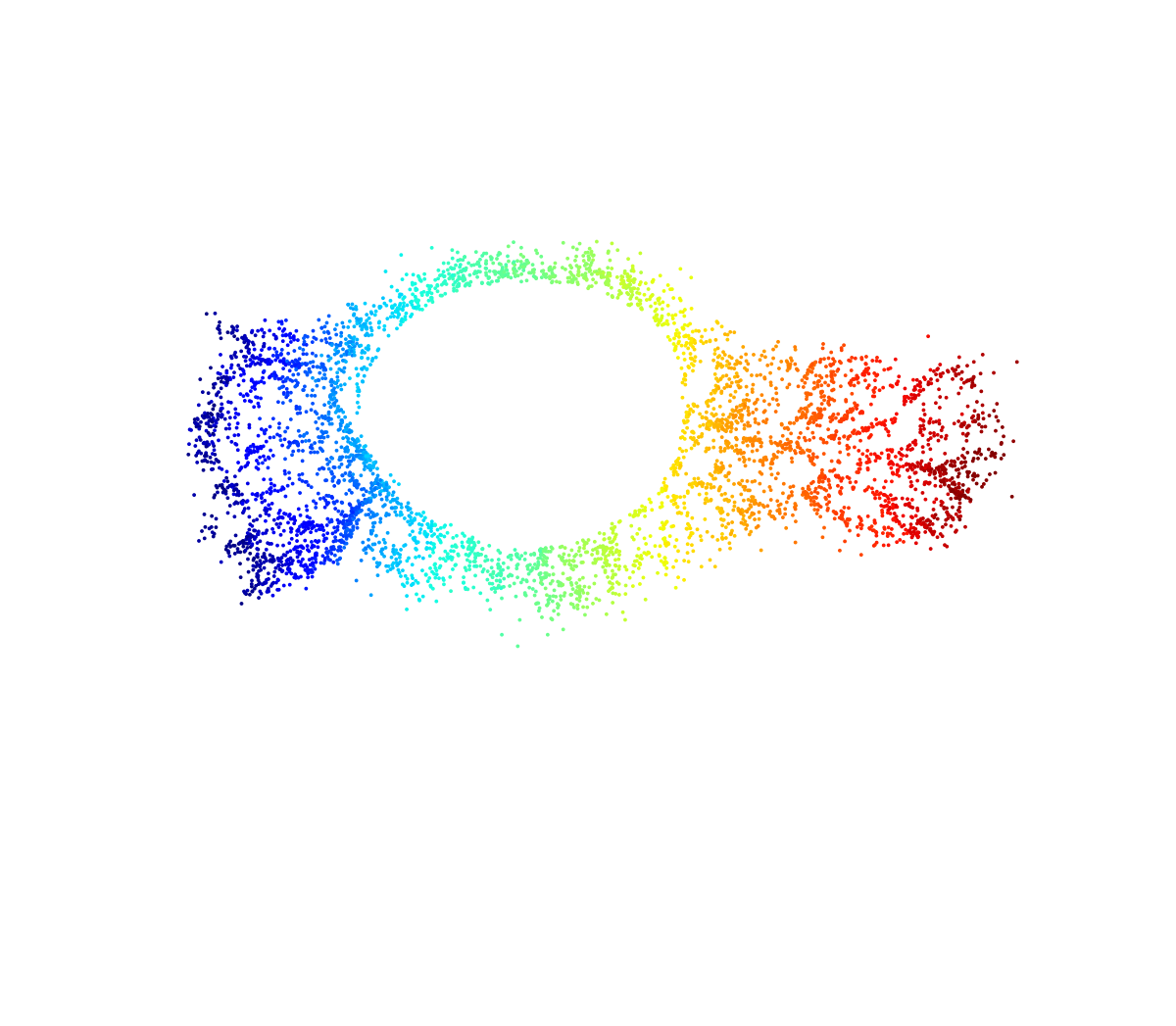} & \includegraphics[width=0.25\textwidth]{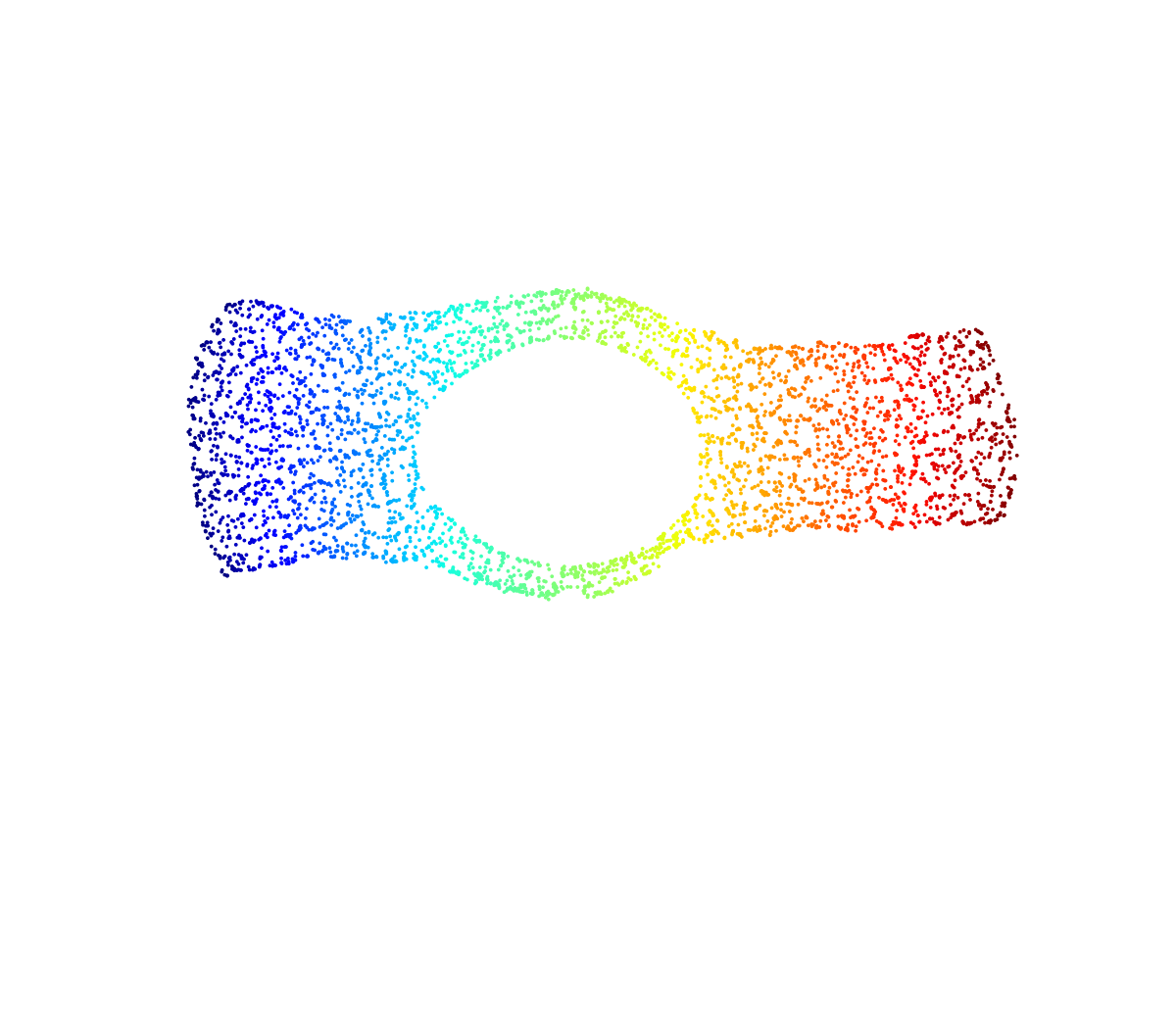} & \includegraphics[width=0.25\textwidth]{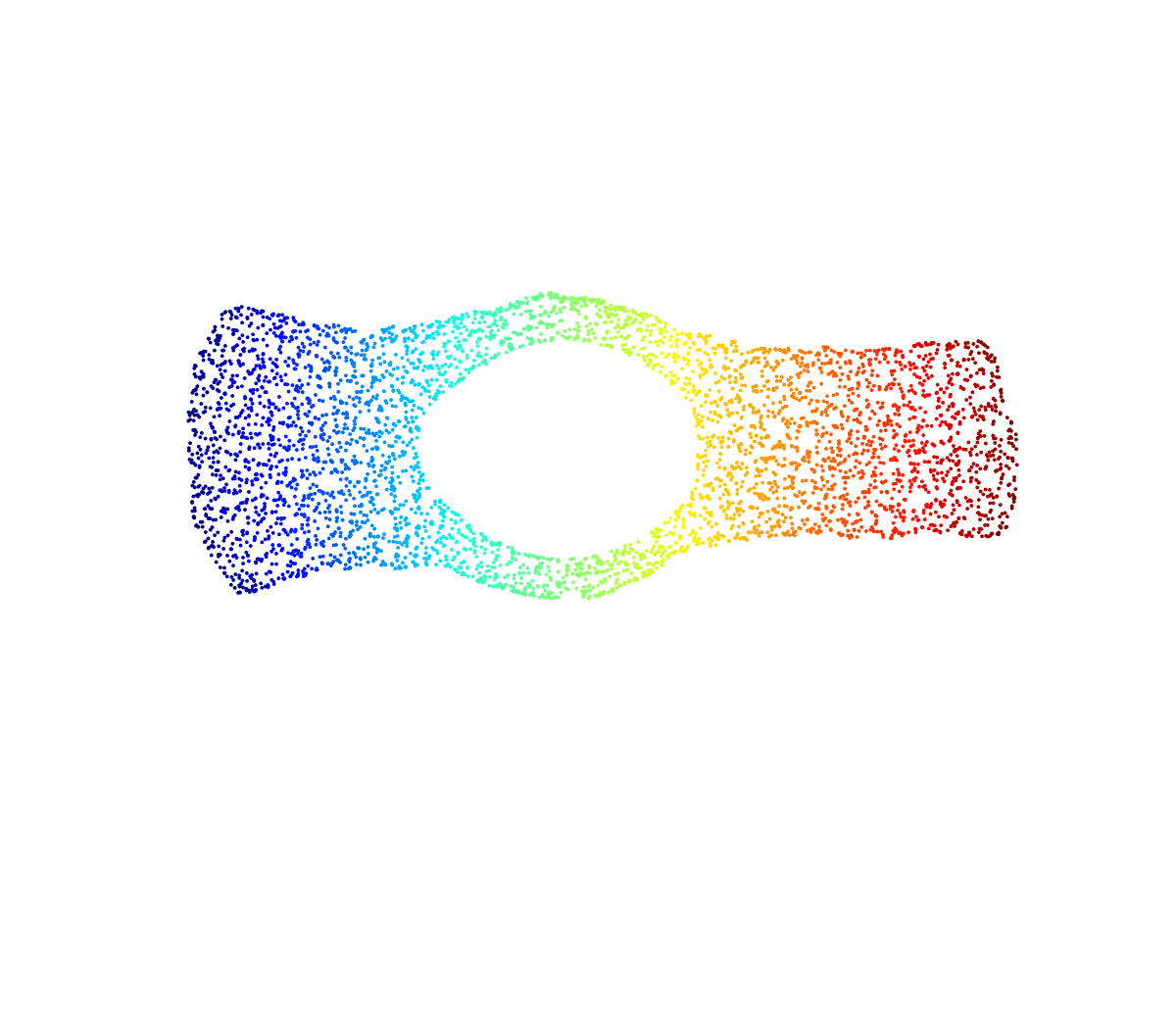}
		\\
		\textbf{$M_{\Pi}$ on Twoomoons} & \includegraphics[width=0.25\textwidth]{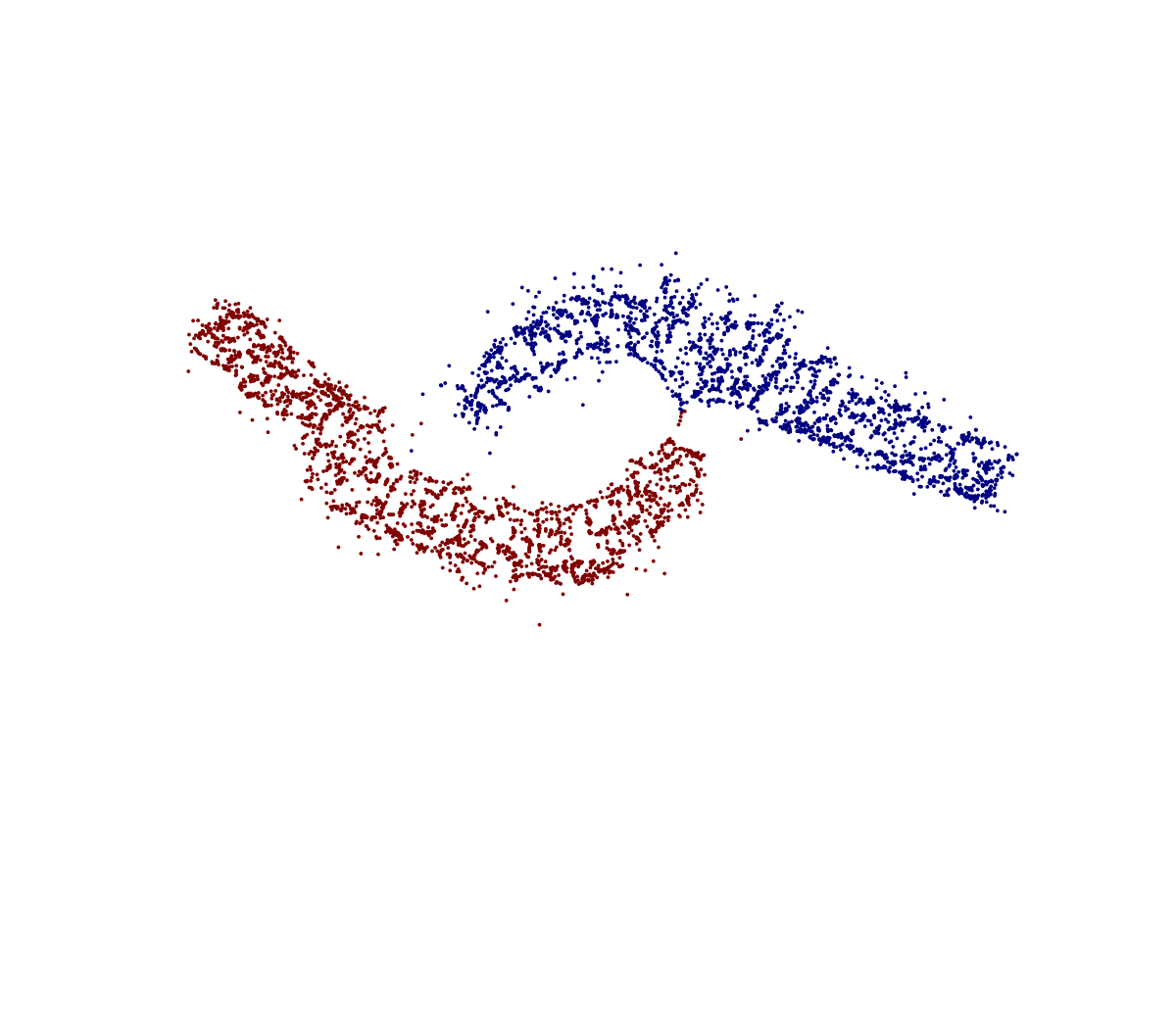} & \includegraphics[width=0.25\textwidth]{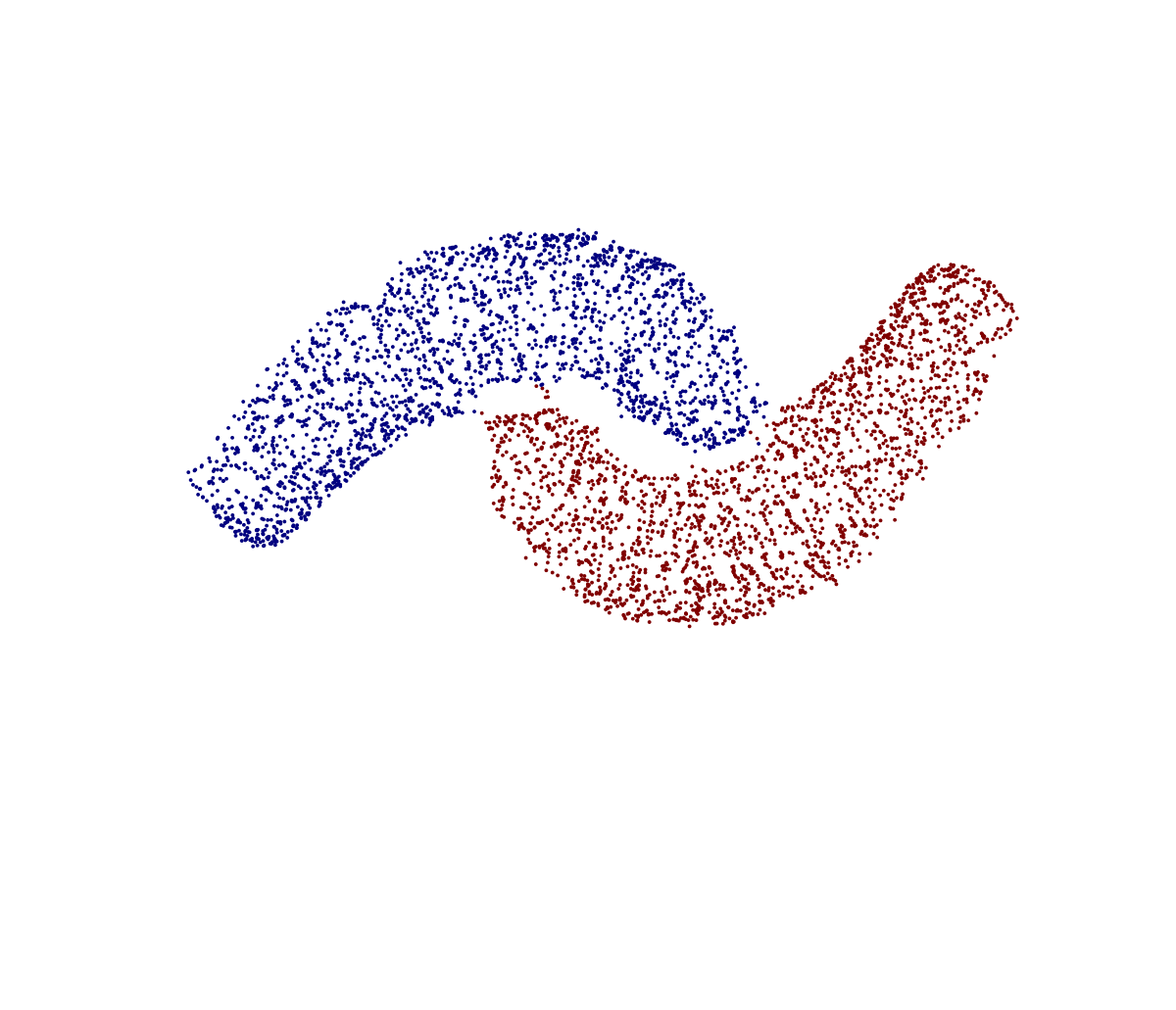} & \includegraphics[width=0.25\textwidth]{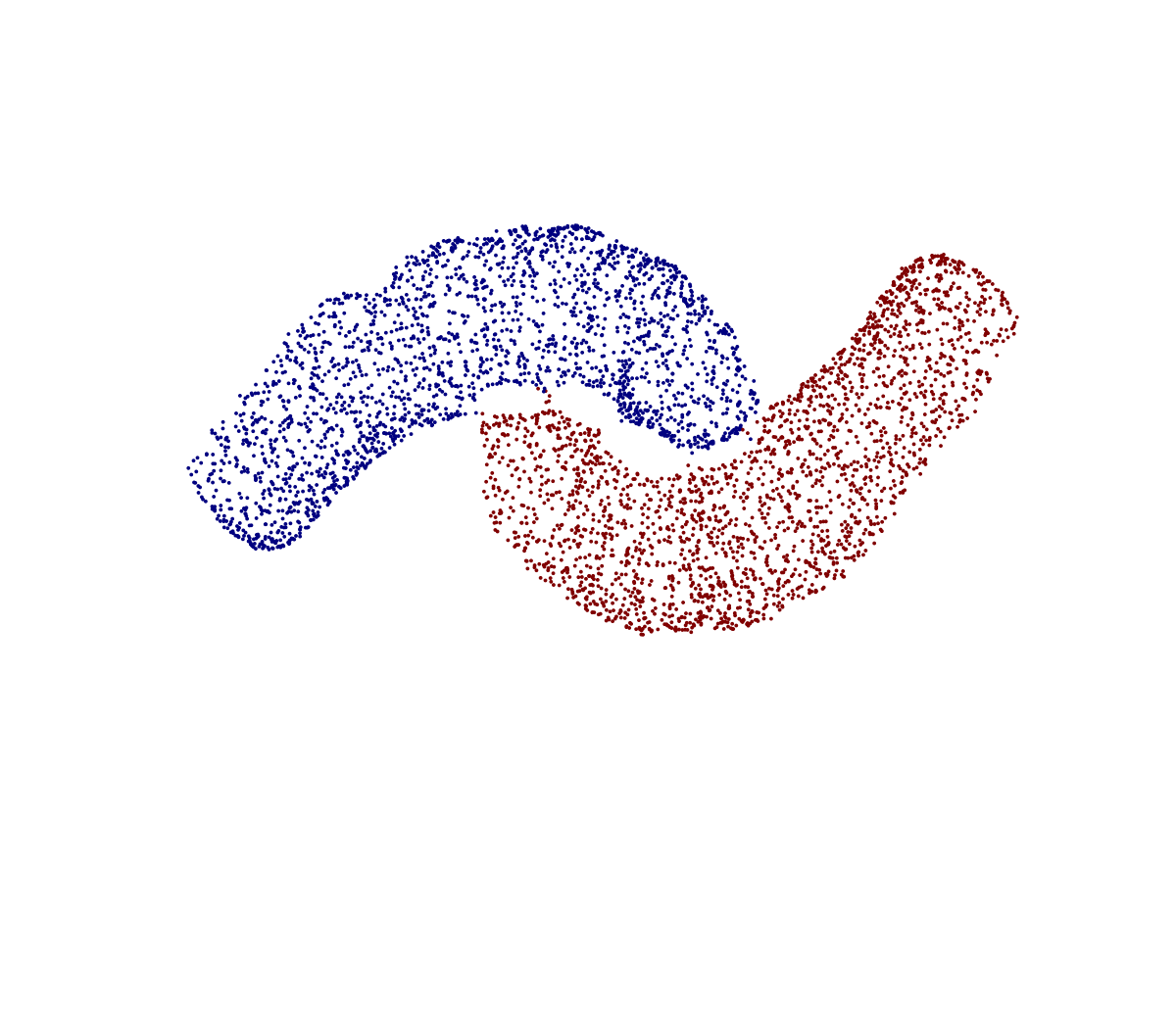} \\
\textbf{$M_W$ on Twomoons} &\includegraphics[width=0.25\textwidth]{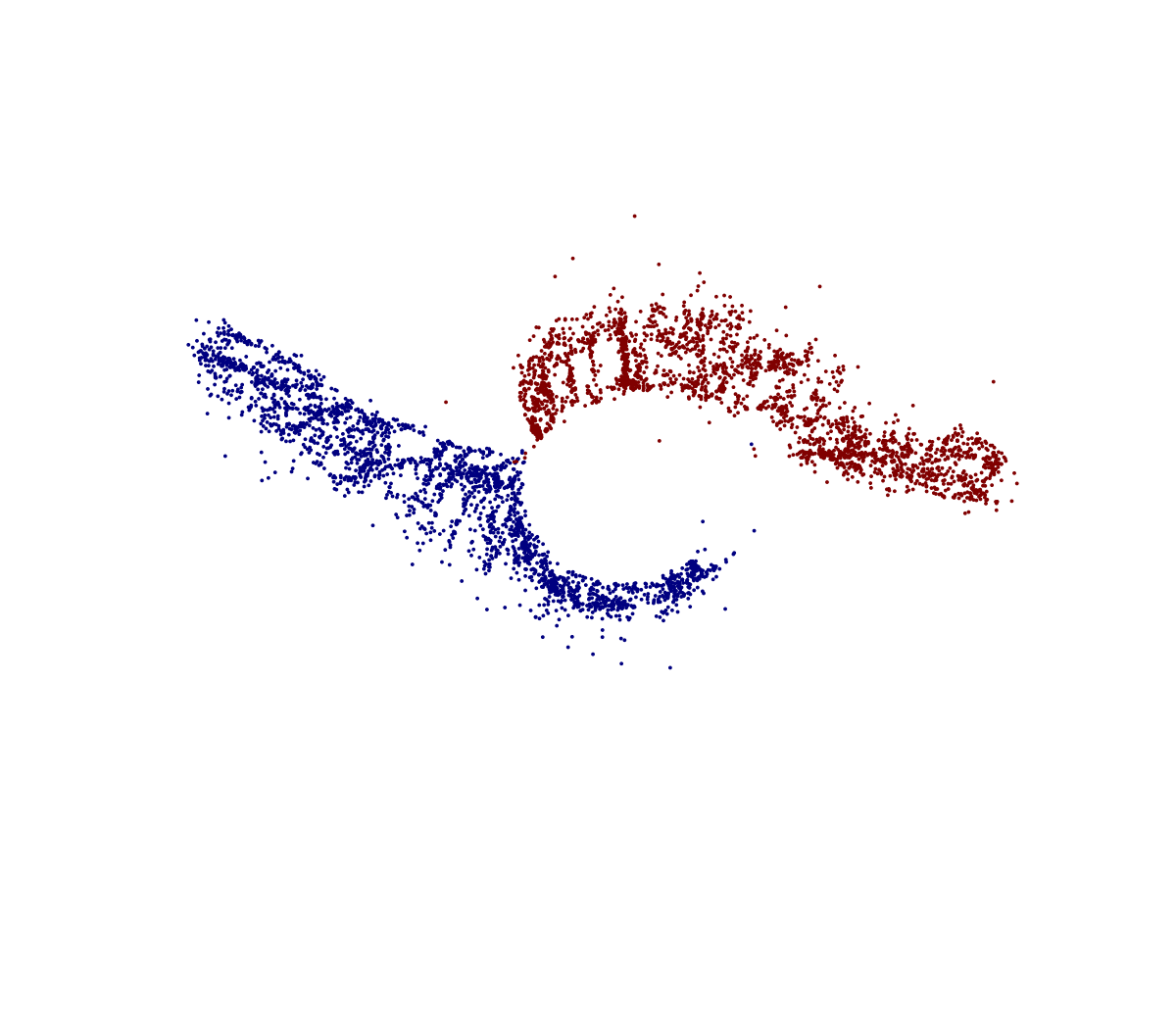} 	
&\includegraphics[width=0.25\textwidth]{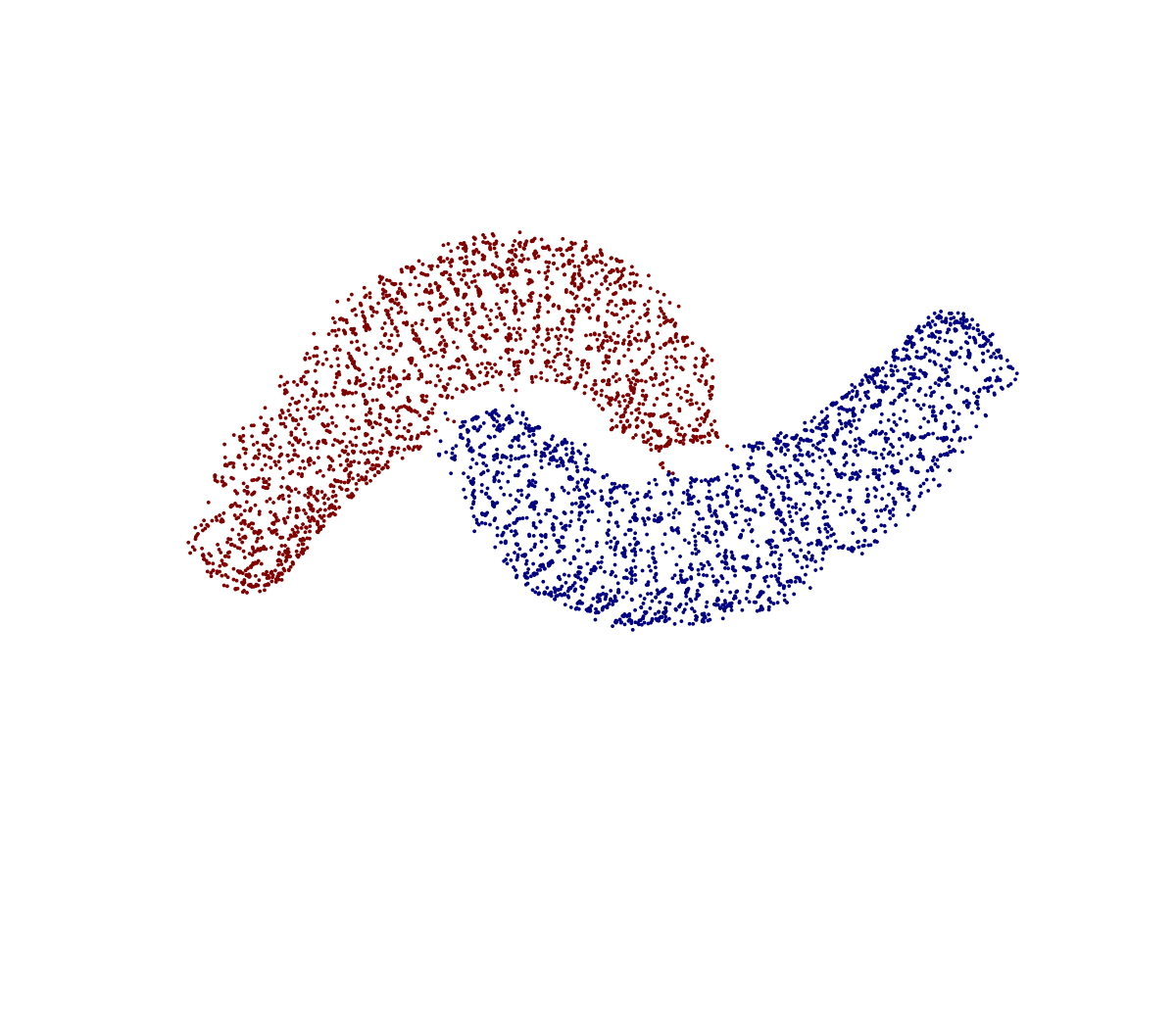} &\includegraphics[width=0.25\textwidth]{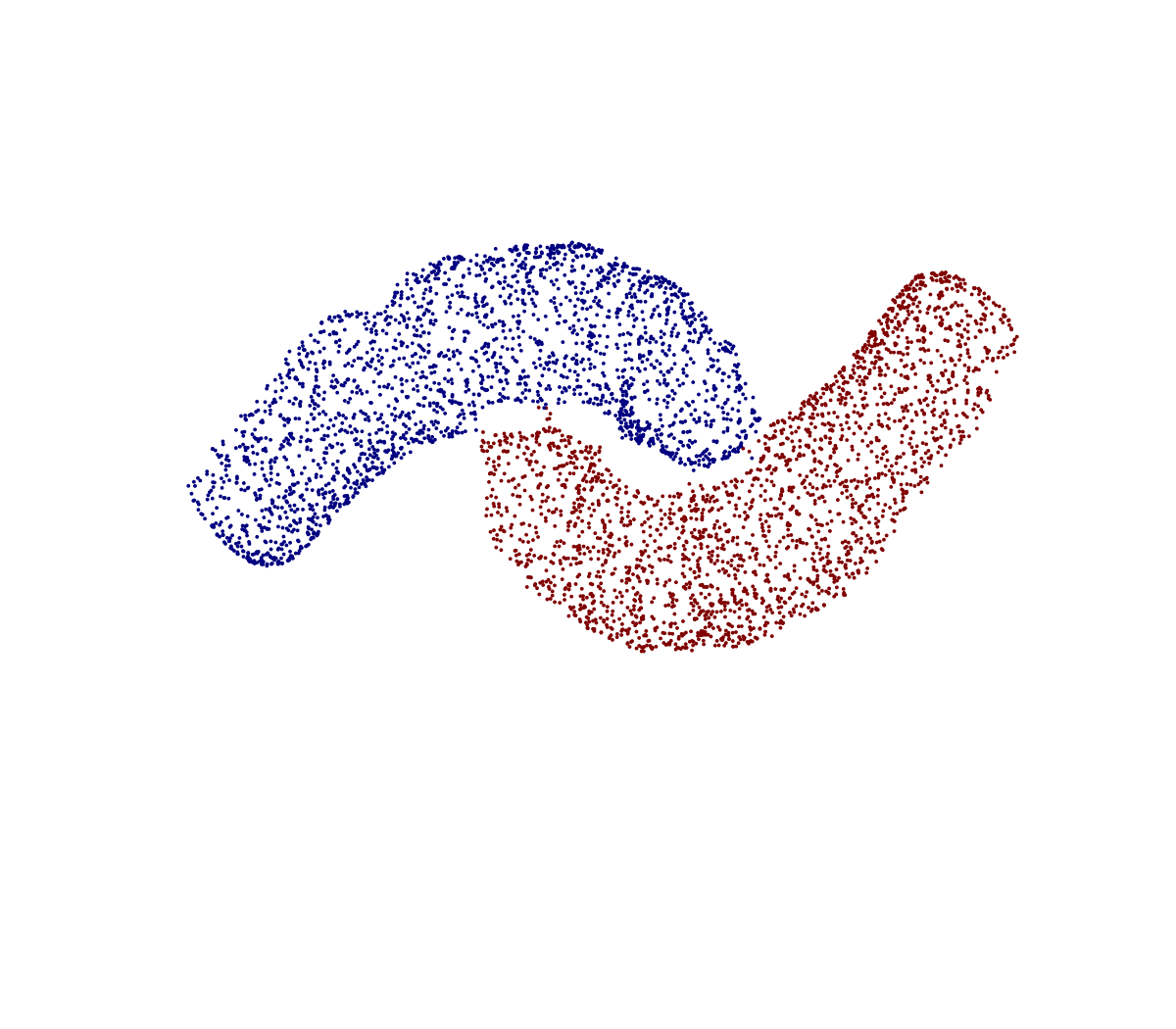} 
	\end{tabular}
	\caption{Sample 5000 dataset from Swiss Roll and Two moons, mapped to $2$-d   applying $M_{\Pi}$ and $M_W$ with  parameters $c=2,5,10$ and $k=15$ in neighborhood graph.}
	\label{tab:higher c}
\end{table}

\bibliography{Hazy_Sets_with_mSchemes}
\bibliographystyle{plain}
\end{document}